%% file: main.tex
\documentclass[10pt,twocolumn,letterpaper]{article}
\usepackage{cvpr}
\input{preamble}

\definecolor{cvprblue}{rgb}{0.21,0.49,0.74}
\usepackage[pagebackref,breaklinks,colorlinks,allcolors=cvprblue]{hyperref}

\def\paperID{40439}
\def\confName{CVPR}
\def\confYear{2026}
\title{Counterfactual Segmentation Reasoning: Diagnosing and Mitigating Pixel-Grounding Hallucination}
\author{Xinzhuo Li\thanks{Equal Contribution}, Adheesh Juvekar\footnotemark[1], Jiaxun Zhang\thanks{Equal Contribution}, Xingyou Liu\footnotemark[2], Muntasir Wahed, \\
Kiet A. Nguyen, Yifan Shen, Tianjiao Yu, Ismini Lourentzou \\
University of Illinois Urbana-Champaign \\
{\tt\small \{xinzhuo4, adheesh2, lourent2\}@illinois.edu}
}

\definecolor{IllinoisOrange}{HTML}{FF5F05}
\definecolor{IllinoisBlue}{HTML}{13294B}
\newcommand{\logoicon}{%
  \raisebox{-0.20\height}{\includegraphics[height=1.2em]{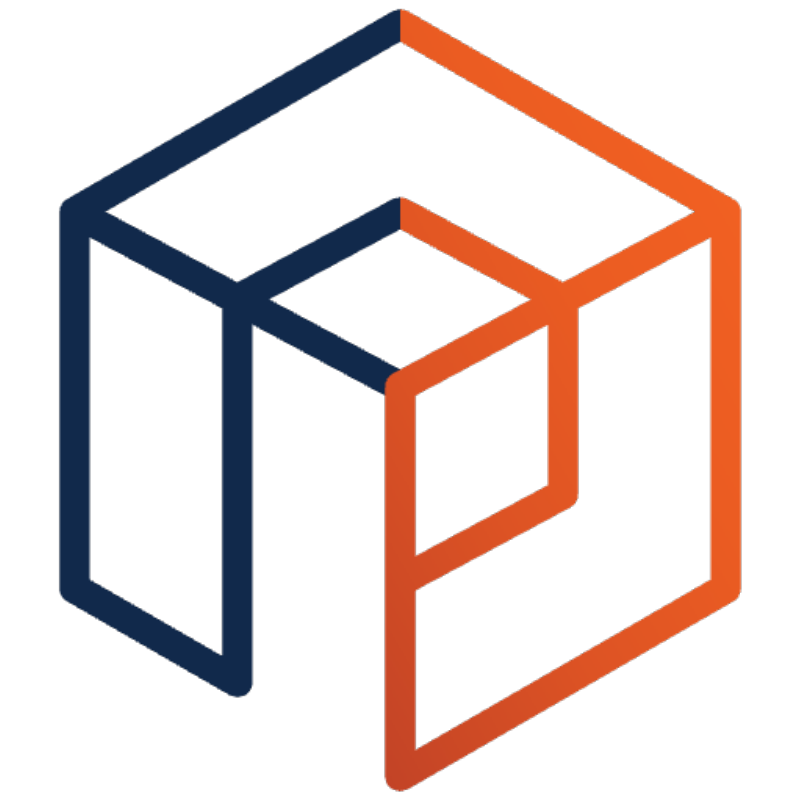}}%
}

\begin{document}
\twocolumn[{%
\renewcommand\twocolumn[1][]{#1}%
\maketitle

\begin{center}
    \centering
    \captionsetup{type=figure}
    \vspace{-0.8cm}
    \includegraphics[width=0.99\linewidth]{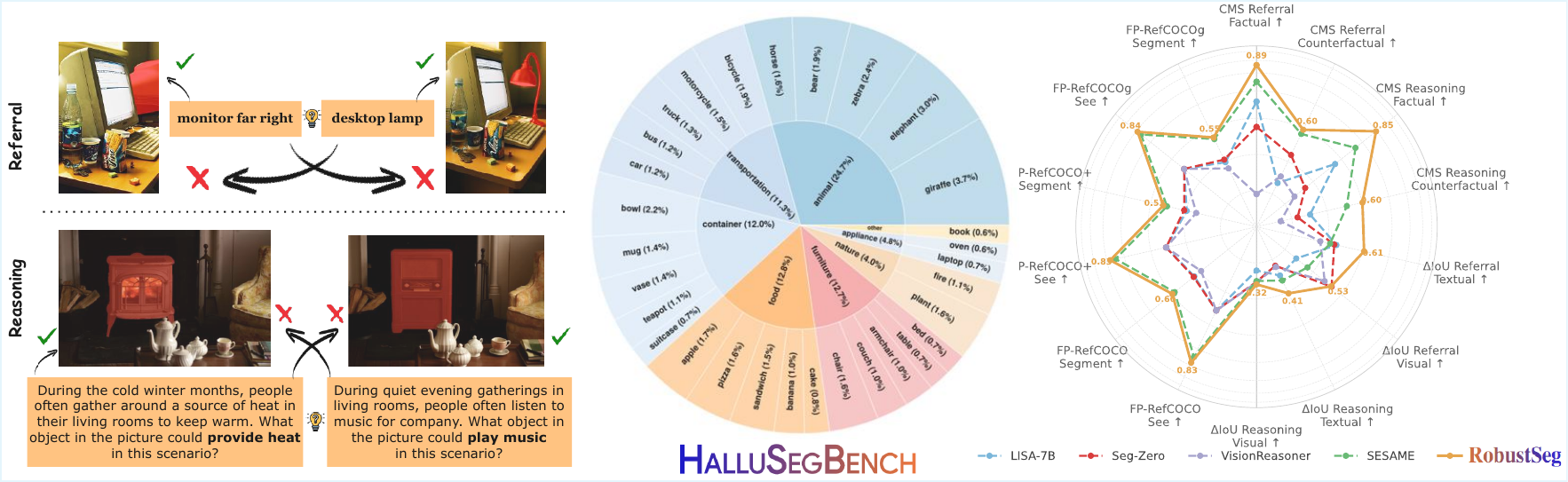}
    \vspace{-0.3cm}
    \caption{\textbf{(left) Counterfactual Segmentation Reasoning.} We introduce the \ul{new task} of counterfactual segmentation, where given \emph{Referring} (top) or \emph{Reasoning} (bottom) instructions, a model must segment the referenced object in the factual image (\textcolor{DarkGreen}{\ding{51}}) and \emph{abstain} in its counterfactual counterpart (\textcolor{red}{\ding{55}}). \textbf{(middle) \benchmarknamegradient{}.} Our \ul{new benchmark} supports both referring and reasoning queries and enables fine-grained analysis of hallucination severity. The chart shows the most frequent objects in each category. \textbf{(right) \modelnamegradient{} Segmentation VLM.} Our \ul{new vision-language segmentation reasoning model}, \modelname{}, trained with a counterfactual finetuning objective, suppresses hallucinations while achieving strong performance across \benchmarkname{} and FP-RefCOCO(+/g). \(\cms\) values are inverted.}
\vspace{0.2cm}
    \label{fig:teaser}
\end{center}%
}]

\maketitle
{\renewcommand{\thefootnote}{\fnsymbol{footnote}}
\footnotetext[1]{Equal contribution \quad
\footnotemark[2] Equal contribution}}
\input{sections/0_abstract}    
\input{sections/1_intro}
\input{sections/2_related_work}

\input{sections/3_method}
\input{sections/4_model}
\input{sections/5_experiments}
\input{sections/6_conclusion}

{
    \small
    \bibliographystyle{ieeenat_fullname}
    \bibliography{main}
}
\appendix
\input{sections/X_suppl}
\end{document}

%% file: preamble.tex
\definecolor{orangy}{RGB}{255,94,0}

\newcommand{\benchmarkname}{{\textsc{HalluSegBench}}\xspace}
\newcommand{\benchmarknamegradient}{\textbf{\textsc{\gradientRGB{HalluSegBench}{0,0,205}{255, 94, 0}}\xspace}}

\newcommand{\modelname}{{{RobustSeg}}\xspace}
\newcommand{\modelnamegradient}{\textbf{{\gradientRGB{RobustSeg}{255, 94, 0}{0,0,205}}\xspace}}

\newcommand\scalemath[2]{\scalebox{#1}{\mbox{\ensuremath{\displaystyle #2}}}}

\newcommand{\csr}{\textsc{CSR}}
\newcommand{\cft}{\textsc{CFT}}
\newcommand{\iou}[2]{\frac{| #1 \cap #2 |}{| #1 \cup #2 |}}
\newcommand{\cms}{\mathrm{CMS}}
\newcommand{\cmsfact}{\mathrm{CMS}_{\text{fact}}}
\newcommand{\cmscf}{\mathrm{CMS}_{\text{counterfact}}}
\newcommand{\diou}{\Delta\mathrm{IoU}}
\newcommand{\dioutext}{\Delta\mathrm{IoU}_{\text{textual}}}
\newcommand{\diouvis}{\Delta\mathrm{IoU}_{\text{visual}}}

\newcommand{\cmark}{\textcolor{ForestGreen}{\ding{51}}}  
\newcommand{\xmark}{\textcolor{red}{\ding{55}}}
\PassOptionsToPackage{pagebackref,breaklinks,colorlinks,allcolors=cvprblue}{hyperref}

\usepackage[T1]{fontenc}
\usepackage{microtype}           
\usepackage{nicefrac}            
\usepackage{xspace}              
\usepackage{amsmath, amssymb, amsfonts, amsthm, mathtools, mathrsfs}
\usepackage{dsfont}              
\usepackage{fdsymbol}            

\usepackage{booktabs}            
\usepackage{nicematrix}         
\usepackage{multirow}
\usepackage{makecell}
\usepackage{bigstrut}
\usepackage{enumitem}            
\setitemize{label=\textbullet, leftmargin=*, nolistsep}
\usepackage{graphicx}
\usepackage{adjustbox}           
\usepackage{float}               
\usepackage{wrapfig}             
\usepackage{subcaption}          
\expandafter\def\csname ver@subfig.sty\endcsname{}  
\usepackage{color}
\usepackage{fontawesome5}        
\usepackage{pifont}              
\usepackage{hyperref}            
\hypersetup{colorlinks=true, citecolor=LightBlue}
\usepackage[all]{hypcap}         
\usepackage{url}                 
\usepackage{algorithm}
\usepackage{algorithmicx}
\usepackage{algpseudocode}
\usepackage[normalem]{ulem}      
\usepackage{lipsum}              
\usepackage{rotating}            
\usepackage{stackengine}         
\usepackage{caption}             
\usepackage{listings}            
\usepackage{soul}                
\usepackage{gradient-text}       
\usepackage{pgfplots}            
\pgfplotsset{compat=newest}
\usepackage{pgfplotstable}       
\usepackage{tikz}                
\usepackage{svg}                 

\usepackage{tcolorbox}
\tcbuselibrary{breakable}

\definecolor{LightBlue}{rgb}{0.68, 0.85, 0.9}
\definecolor{amberyellow}{rgb}{1.0, 0.75, 0.0}
\definecolor{DarkGreen}{rgb}{0.0, 0.5, 0.0}
\definecolor{DarkBlue}{rgb}{0,0,205}

%% file: sections/0_abstract.tex
\begin{abstract}
Segmentation Vision-Language Models (VLMs) have significantly advanced grounded visual understanding, yet they remain prone to pixel-grounding hallucinations, producing masks for incorrect objects or for objects that are absent. Existing evaluations rely almost entirely on text- or label-based perturbations, which check only whether the predicted mask matches the queried label. Such evaluations overlook the spatial footprint and severity of hallucination and therefore fail to reveal vision-driven hallucinations, which are more challenging and more prevalent.
To address this gap, we formalize the task of \textbf{Counterfactual Segmentation Reasoning (\csr)}, where a model must segment the referenced object in the factual image and abstain in its counterfactual counterpart. 
To support this task, we curate \benchmarknamegradient{}, the first large-scale benchmark to diagnose referring and reasoning expression segmentation hallucinations using controlled visual counterfactuals, alongside new evaluation metrics that measure hallucination severity and disentangle vision- and language-driven failure modes.
We further introduce \modelnamegradient{},
a segmentation VLM trained with counterfactual finetuning (\cft) to learn when to segment and when to abstain. Experimental results confirm \modelname{} reduces hallucinations by 30\%, while improving segmentation performance on FP-RefCOCO(+/g).

\noindent \logoicon~\href{https://plan-lab.github.io/hallusegbench}{\textcolor{IllinoisBlue}{PLAN Lab}~\textcolor{IllinoisOrange}{https://plan-lab.github.io/hallusegbench}}
\end{abstract}

%% file: sections/1_intro.tex
\section{Introduction}
Vision-Language Models (VLMs) have rapidly advanced multimodal understanding by integrating visual and textual information at scale~\cite{alayrac2022flamingo,liu2024visual}. VLMs now achieve state-of-the-art performance across diverse tasks, such as visual question answering~\cite{liu2024visual,sinha2025guiding}, image captioning~\cite{li2023blip,wang2022ofa}, and object detection~\cite{ma2024groma}. Recent progress extends these capabilities to reasoning-based segmentation~\cite{lai2024lisa,rasheed2024glamm,ren2024pixellm, liu2025medsam3, nguyen2025calico} and spatial reasoning tasks~\cite{chen2020uniter,chen2024spatialvlm}, where models must localize fine-grained visual entities conditioned on natural language descriptions. This fine-grained alignment of linguistic and visual representations has introduced new possibilities for segmenting objects through contextual understanding and semantic cues, bridging the gap between textual queries and precise pixel-level predictions.

Despite these advances, hallucinations remain a fundamental bottleneck~\cite{zhang2023siren,huang2024visual}. 
Hallucinations can manifest in several ways, such as pixel-grounding hallucinations~\cite{wu2024see,xia2024gsva,liu2023gres}, where a model may generate spatially plausible but semantically incorrect masks, or object-level hallucinations~\cite{li2023evaluating,dai2023plausible,rohrbach2018object,geigle2024does}, when a model incorrectly mentions or labels objects that do not exist in a scene.
While object hallucinations are relatively straightforward to diagnose by comparing predicted labels against global visual content, pixel-grounding hallucinations are far more insidious, as a model may draw a coherent mask tightly aligned with the scene, yet for the wrong object. 
Unlike object hallucinations, which can be identified without dense annotations, detecting pixel-grounding hallucinations requires fine-grained, pixel-level supervision. 
Such failures are especially problematic in dense prediction tasks and cluttered scenes, where visual similarity and local context make it easy for models to substitute true grounding with learned priors.

Yet, existing benchmarks provide limited evaluation of pixel-grounding hallucinations. Most evaluation protocols attempt to elicit hallucinations by introducing text queries and synthetic labels for objects absent from the 
image~\cite{wu2024towards,wu2024see}. While useful for probing robustness, these synthetic negatives are often semantically implausible, visually ungrounded, and insufficient to reveal whether a model genuinely attends to visual evidence, making it relatively straightforward for models to reject them without truly demonstrating grounding capabilities. 
Furthermore, they fail to expose models to \textit{counterfactual scenarios}, \ie, perceptually coherent but semantically modified images that require the model to distinguish fine-grained visual differences grounded in referential language, a challenging setting where hallucination is both more subtle and more harmful.

To bridge this gap, we introduce \textbf{Counterfactual Segmentation Reasoning}, a new paired grounding task in which, given a referring or reasoning query, a model must segment the referenced object in a factual scene and abstain in its counterfactual counterpart.
We further present \benchmarknamegradient{}, a large-scale benchmark for diagnosing and mitigating pixel-grounding hallucinations under controlled, visually coherent counterfactual edits. \benchmarkname{} spans both segmentation reasoning and referring expression segmentation settings, and consists of 
factual-counterfactual image pairs where a target object is systematically replaced with a similar alternative, while the surrounding context remains unchanged. Each image pair is accompanied by ground truth masks for both object classes, enabling fine-grained analysis of hallucination severity. 
Our dataset comprises 3,673 factual–counterfactual image pairs across 816 categories. To the best of our knowledge, this is the first benchmark enabling pixel-grounding hallucination evaluation in both referring and reasoning settings.

Beyond the benchmark, we introduce \textbf{a suite of new evaluation metrics} that capture hallucination robustness along complementary dimensions: (i) sensitivity to contextual shifts, (ii) semantic consistency under label or object replacements, and (iii) structured hallucination severity via distractor-aware analysis. These metrics reveal failure modes that conventional accuracy or false-premise evaluations fail to expose. Across state-of-the-art segmentation VLMs, our analysis uncovers that vision-driven hallucinations are significantly more prevalent than label-driven ones, indicating reliance on contextual priors rather than visual evidence.

Finally, we introduce \modelnamegradient{}, a hallucination-mitigating segmentation VLM trained with a counterfactual finetuning objective that leverages factual/counterfactual supervision. \modelname{} learns to reliably distinguish visual evidence from contextual bias, achieving large and consistent reductions in hallucination while preserving segmentation accuracy. This positions \benchmarkname{} as a rigorous evaluation benchmark but also as a practical training resource for building more reliable segmentation models.

\noindent The contributions of this work are:

\begin{itemize}[itemsep=0.5ex, parsep=0pt, topsep=-2.3pt, leftmargin=0.4cm]
\item \textbf{New task:} We introduce the novel task of Counterfactual Segmentation Reasoning (CSR) that requires models to segment objects when present and abstain under controlled counterfactual removals, directly probing whether grounding is driven by visual evidence.

\item \textbf{New benchmark:} We curate \benchmarkname{}, the first benchmark for pixel-grounding hallucination via counterfactual visual reasoning, and propose novel evaluation metrics that quantify counterfactual consistency and distractor-aware hallucination severity.

\item \textbf{New model:} We propose \modelname{}, a segmentation VLM, trained with Counterfactual Finetuning (CFT) to acquire abstention-aware grounding capabilities. \modelname{} substantially reduces hallucinations while preserving segmentation fidelity, achieving strong gains on both \benchmarkname{} and FP-RefCOCO(+/g) benchmarks.

\end{itemize}

%% file: sections/2_related_work.tex
\section{Related Work}
\textbf{Hallucination Evaluation and Mitigation.}
Advances in segmentation have significantly enhanced the ability of VLMs to understand and reason about fine-grained visual concepts~\cite{ren2024grounded,kirillov2023segment}. Seminal work by LISA~\cite{lai2024lisa} pioneered the integration of LLMs with pixel-level reasoning, demonstrating fine-grained alignment between natural language descriptions and precise pixel regions, and inspiring a series of follow-up efforts~\cite{zhang2024omg, wahed2024prima, liu2025seg, wei2024hyperseg, wei2024instructseg,rasheed2024glamm,ren2024pixellm}. Despite these advances, hallucination remains a persistent challenge. Existing studies have sought to quantify and mitigate object hallucination in VLMs~\cite{ben2024mitigating,li2023evaluating, lovenia2024negative, liu2025phd}. However, these evaluations are often limited to textual mismatches without grounding or precisely locating objects. Recent approaches that introduce mask-based evaluation and mitigation for segmentation hallucination ~\cite{wu2024towards,wu2024see} primarily rely on synthetic text-based perturbations or randomly sampled false premises that are disconnected from the actual visual context and train with them for mitigation. Furthermore, the absence of altered visual segmentation accompanying hallucinated labels renders these datasets insufficiently challenging for comprehensive segmentation hallucination evaluation at the pixel level. 
Recently, CCEval~\cite{zhai2023halle} proposed a GPT-4-assisted evaluation method for {caption-based hallucination detection} in LLMs, but focuses on whether hallucinated objects appear in textual descriptions, rather than in pixel-level grounding or visual segmentation hallucinations. In contrast, \benchmarkname{} directly evaluates segmentation hallucination at the pixel level through controlled {counterfactual interventions}, enabling the discovery of {hallucination-induced segmentation failures} in VLMs that are undetectable through caption-based methods alone.

\noindent
\textbf{Counterfactual Reasoning.}
Counterfactual reasoning has emerged as a powerful tool for evaluating model reliability under hypothetical visual alterations. In Visual Question Answering (VQA), counterfactual methods reveal hidden biases by perturbing image-question pairs~\cite{niu2021counterfactual,liang2020learning}. Similarly, Vision-and-Language Navigation (VLN) works use counterfactual path generation~\cite{parvaneh2020counterfactual,wang2022counterfactual} to improve navigational robustness through hypothetical scenarios. In Referring Expression Comprehension (REC), counterfactual is also applied for hallucination mitigation~\cite{yu2024revisiting, liu2023gres}. Recent work~\cite{bhattad2025visual} introduces a training-free counterfactual inpainting approach to discover object dependencies by measuring the semantic impact of object removals on scene plausibility. While these approaches underscore the value of counterfactual visual reasoning, they primarily focus on high-level tasks or structural coherence rather than fine-grained visual grounding. Our proposed \benchmarkname{} and \modelname{} aim to bridge this gap by introducing a framework with visually grounded counterfactual interventions. 

\noindent
\textbf{Controlled Image Editing.}
Generative models have enabled controlled image modifications while preserving semantic coherence~\cite{karras2019style,zhu2017unpaired}, with diffusion models gaining attention for image editing tasks~\cite{zhang2023magicbrush, brooks2023instructpix2pix, huang2024smartedit, sobieski2024rethinking}. In mask-guided image editing frameworks, user-defined masks enable precise region-specific modifications, ensuring coherent adjustments that align with textual or semantic prompts~\cite{avrahami2023blended,nichol2022glide}. Instruction-guided variants~\cite{wang2023instructedit, li2024zone} allow grounded edits with mask-generation, while counterfactual diffusion~\cite{song2024doubly} introduces structured scene alterations. In the counterfactual setting, Countercurate~\cite{zhang2024countercurate} and FineCops-Ref~\cite{liu2024finecops} apply image editing to generate counterfactual pairs for referring comprehension. However, their methods are limited to bounding-box-level grounding and do not provide finer-grained pixel-level supervision. While prior image editing work~\cite{yin2024benchmarking} benchmarks segmentation models by altering object attributes with a mask-preserved method, it remains confined to intra-object or global variations, limiting its applicability to scenarios involving the same object. These approaches fall short in pixel-level counterfactual scenarios where object replacements are performed within their original context, settings that require models to maintain visual grounding without introducing spurious hallucinations. Our work preserves contextual integrity across factual and counterfactual scenarios, enabling a more granular analysis and mitigation framework of segmentation reliability and hallucination robustness under counterfactual interventions.
\input{assets/figures/examples_pred}

%% file: assets/figures/examples_pred.tex
\begin{figure}[t!]
    \centering
    \resizebox{0.99\columnwidth}{!}{
        \begin{subfigure}[t]{0.14\textwidth}
        \centering
        \includegraphics[width=\linewidth]{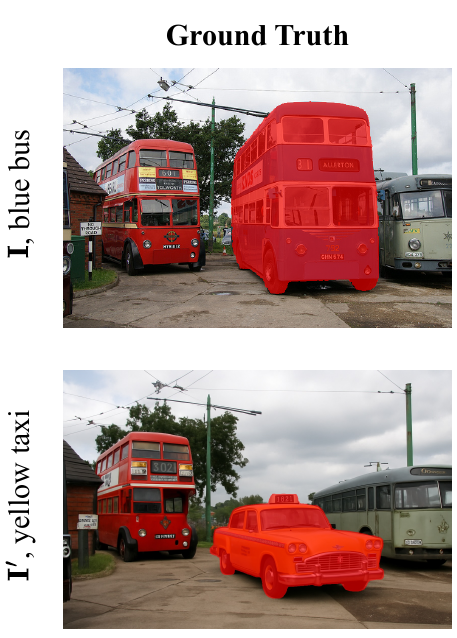}
        \caption{Ground-truth}
        \label{fig:pred_vs_gt_a}
    \end{subfigure}
    \begin{subfigure}[t]{0.28\textwidth}
        \centering
        \includegraphics[width=\linewidth]{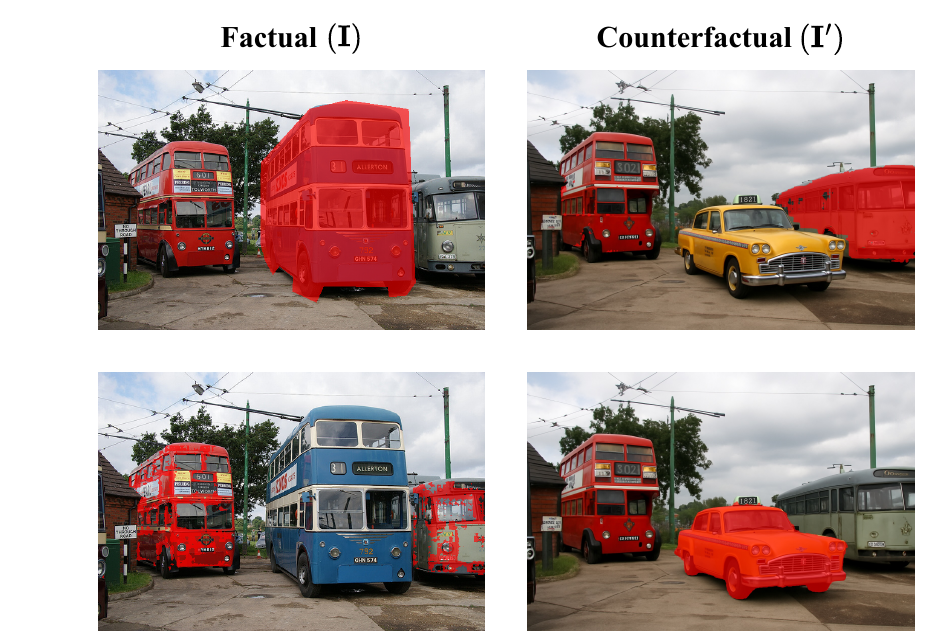}
        \caption{Predicted Masks}
        \label{fig:pred_vs_gt_b}
    \end{subfigure}
    }
    \vspace{-0.2cm}
    \caption{\textbf{Segmentation from LISA~\cite{lai2024lisa} Segmentation VLM Under Factual and Counterfactual Settings.} The rows represent the factual image \(\mathbf{I}\) and its counterfactual variant \(\mathbf{I}'\), where the original object of class \textbf{\textcolor{blue}{``blue bus''}} is replaced with a visually similar object of class \textbf{\textcolor{amberyellow}{``yellow taxi''}}.
    (a) shows ground-truth masks, while (b) shows the corresponding model predictions. Hallucinations can be observed when there's an image-label mismatch.}  
    \label{fig:pred_vs_gt}
    \vspace{-0.2cm}
\end{figure}

%% file: sections/3_method.tex
\input{assets/figures/dataset_statistics}

\section{Counterfactual Segmentation Reasoning}\label{sec:cvr}
Grounded segmentation aims to localize objects specified by natural-language queries, such as referring expressions (\eg, ``red cup on the left'')~\cite{yu2016modeling,wu2024towards,ren2024grounded,liu2023gres} or reasoning-based descriptions (\eg, ``item used for watering plants'')~\cite{lai2024lisa,rasheed2024glamm,ren2024pixellm,wahed2024prima}. While hallucinations are extensively studied in image-level vision-language understanding tasks like image captioning and VQA~\cite{niu2021counterfactual,liang2020learning}, segmentation hallucination evaluation remains underexplored.

To address this gap, we introduce the Counterfactual Segmentation Reasoning (CSR) task to test whether segmentation VLMs adapt their predictions under controlled visual interventions. CSR is motivated by a characteristic failure in reasoning queries, where models often segment a contextually plausible object, such as a bottle for ``item used for watering plants'',  even when the true object is absent. Such outputs are visually coherent and semantically reasonable, and cannot be exposed through text-only perturbations, false-premise queries, or object-level correctness checks. CSR directly exposes this behavior through controlled counterfactual edits: the target object in a factual scene is replaced with a semantically similar alternative, while all remaining pixels are held fixed. This factual-counterfactual pairing enables fine-grained measurement of whether a model correctly abstains when the object is removed and, if not, how severe the hallucination is, measured by the size and structure of incorrect mask regions overlapping distractor objects.

Formally, given a factual image \(\mathbf{I}\) containing a target object of class \(c\), and a text prompt that references the same object of class \(c\) to be segmented (\eg ``the red apple on the table''), a vision-language segmentation model \(f\)  predicts a segmentation mask \(\hat{\mathbf{M}}_c\!=\!f(\mathbf{I}, c)\)\footnote{We use the notation \(f(\mathbf{I}, c)\) instead of explicitly referencing the text prompt for brevity and clarity.}. 

\noindent CSR evaluates four predictions:
\begin{itemize}[itemsep=0ex, parsep=0pt, topsep=0pt, leftmargin=0.3cm]
    \item \(\hat{\mathbf{M}}_c\!=\!f(\mathbf{I}, c)\) where object \(c\) in factual image \(\mathbf{I}\).
    \item \(\hat{\mathbf{M}}_{c'}\!=\!f(\mathbf{I}, c')\) where counterfactual object \(c'\) in image \(\mathbf{I}\).
    \item \(\hat{\mathbf{M}}_c'\!=\!f(\mathbf{I}', c)\) with object  \(c\) in counterfactual image \(\mathbf{I}'\).
    \item \(\hat{\mathbf{M}}_{c'}'\!=\!f(\mathbf{I}', c')\) for counterfactual object and image \(c'\),\(\mathbf{I}'\).
\end{itemize}
Here, \(\mathbf{I}'\) is an edited version of \(\mathbf{I}\), in which the object of class \(c\) is replaced with a semantically similar object of another class \(c'\), and all other pixels in the scene are held constant. This controlled intervention isolates the visual evidence associated with object \(c\), enabling us to test whether the model's predictions are grounded in true visual cues or driven by semantic priors.
For the factual image \(\mathbf{I}\), we expect the model to segment \(c\) accurately while suppressing predictions for \(c'\). Conversely, for the counterfactual image \(\mathbf{I}'\), the model should suppress \(c\) while segmenting \(c'\) reliably. 

\Cref{fig:pred_vs_gt} illustrates an example of the predicted masks (\Cref{fig:pred_vs_gt_b}) across these four scenarios, and their corresponding ground truth masks (\Cref{fig:pred_vs_gt_a}). Counterfactual pairs evaluate whether the model correctly segments the target object when it is present (\ie \(\hat{\mathbf{M}}_c\) and \(\hat{\mathbf{M}}_{c'}'\)), and if it hallucinates visually absent objects (\ie \(\hat{\mathbf{M}}_{c'}\) and \(\hat{\mathbf{M}}_c'\)). Predictions are compared against the corresponding ground truth masks \(\mathbf{M}_c\) and \(\mathbf{M}_{c'}'\). 
This structured setting serves as the foundation of our hallucination analysis.

\input{assets/figures/dataset_donuts}
\section{\benchmarknamegradient{}}

We introduce \benchmarkname{}, a benchmark designed to evaluate segmentation hallucination through controlled visual counterfactuals. Section \S\ref{sec:dataset} presents the dataset while Section \S\ref{sec:metrics} describes our novel metrics used to assess hallucination severity and grounding fidelity.

\subsection{Benchmark}\label{sec:dataset}
\benchmarkname{} is constructed using images from the RefCOCO~\cite{yu2016modeling} referring expression and ReasonSeg~\cite{lai2024lisa} segmentation reasoning datasets. Images from the validation and test splits are used solely for evaluation, while the training splits are used for model finetuning.
To enable precise counterfactual analysis, we systematically identify object instances suitable for reliable replacement and generate deterministic class-level edit instructions for each target instance. These instructions follow a controlled format, \eg \emph{``Change \textless object A\textgreater{} to \textless object B\textgreater{}''}, which drives an automated image editing module to produce visually coherent counterfactual images while preserving global scene structure.
For each factual image \(\mathbf{I}\), we retain the ground-truth mask \(\mathbf{M}_c\) provided by the original datasets and generate the corresponding mask \(\mathbf{M}_{c'}'\) based on the replacement class. For the reasoning split, we additionally construct counterfactual reasoning queries by replacing the referenced object in the original question with the corresponding counterfactual class while preserving the query structure. All generated pairs are manually filtered to ensure visual fidelity and annotation correctness. Additional details on dataset construction and quality control are provided in the Appendix.

\noindent \textbf{Dataset Statistics.}
\benchmarkname{} comprises \(3,673\) factual-counterfactual pairs, yielding \(7,346\) object masks over \(2,958\) images and spanning across \(816\) unique object classes. 
The referring split has \(1{,}472\) train and \(1{,}340\) test pairs (\(2{,}944\) and \(2{,}680\) masks; \(1{,}373\) and \(1{,}002\) images), while the reasoning split includes \(462\) train and \(399\) test pairs (\(924\) and \(798\) masks; \(208\) and \(375\) images). 
\Cref{fig:dataset_a} (left) shows mask-area distributions for factual, counterfactual, and all masks (combined). Most objects occupy 5–15\% of the image area, consistent with natural scenes where objects are embedded within larger visual contexts. Factual and counterfactual masks exhibit nearly identical size distributions, confirming that replacements preserve relative object scale.
\Cref{fig:dataset_b} (right) lists the top {factual} \(\rightarrow\) {counterfactual} replacements (\eg \emph{elephant}\(\rightarrow\)\emph{giraffe}, \emph{bowl}\(\rightarrow\)\emph{teapot}), highlighting the spatial and semantic diversity of \benchmarkname{}. 
Notably, many of the top replacements involve visually and semantically similar objects, providing challenging settings for assessing fine-grained grounding fidelity.
\Cref{fig:teaser} (b) depicts the overall object-category distribution of \benchmarkname{}, which spans diverse indoor and outdoor categories (animals, vehicles, furniture, \etc).

\subsection{Evaluation Metrics}\label{sec:metrics}
To assess hallucination in segmentation VLMs under counterfactual conditions, we introduce two new sets of counterfactual hallucination evaluation metrics.

\input{assets/figures/cft}

\noindent \textbf{Consistency-based Metrics.}
This class of metrics assesses the sensitivity of segmentation predictions to controlled counterfactual interventions. They quantify how model outputs change when either the visual input or the text query is systematically manipulated, thereby capturing both segmentation fidelity and hallucination susceptibility. Concretely, we compare prediction quality on the factual image with performance under swapped textual labels or counterfactual visual inputs. A robust model should suppress predictions on mismatched pairs.

Let \(\mathbf{M}_c\) and \(\mathbf{M}_{c'}'\) denote the ground truth masks for query \(c\) in the factual image \(\mathbf{I}\) and query \(c'\) in the counterfactual image \(\mathbf{I}'\), respectively. Correspondingly, \(\hat{\mathbf{M}}_c\), \(\hat{\mathbf{M}}_{c'}\), and \(\hat{\mathbf{M}}_c'\) denote the predicted masks for class \(c\) in \(\mathbf{I}\), class \(c'\) in \(\mathbf{I}\), and class \(c\) in \(\mathbf{I}'\), respectively. 
\begin{equation}
\setlength{\abovedisplayskip}{7pt}   
\setlength{\belowdisplayskip}{7pt}   
\scalemath{0.73}{
\text{IoU}_{\text{fact}}\!=\!\iou{\mathbf{M}_c}{\hat{\mathbf{M}}_c},\;
\text{IoU}_{\text{textual}}\!=\!\iou{\mathbf{M}_c}{\hat{\mathbf{M}}_{c'}},\;
\text{IoU}_{\text{visual}}\!=\!\iou{\mathbf{M}_{c'}'}{\hat{\mathbf{M}}_c'}
}
\label{eq:ious}
\end{equation}
Here, \(\text{IoU}_{\text{fact}}\) measures the segmentation accuracy in the factual setting, \(\text{IoU}_{\text{textual}}\) measures model leakage when querying with a counterfactual label on the same image, and \(\text{IoU}_{\text{visual}}\) evaluates hallucination when the queried object is absent in the counterfactual image but still prompted.

\(\dioutext\) quantifies the degradation in segmentation performance when the textual query is swapped, while the visual content remains unchanged, \ie, \(\dioutext \!=\!\text{IoU}_{\text{fact}} - \text{IoU}_{\text{textual}}\). This metric measures the model's ability to suppress predictions for objects that are not visually present when prompted with an incorrect label. Larger values indicate better grounding, while smaller values suggest over-reliance on language priors, leading to hallucinations. \( \diouvis\) captures the change in segmentation performance when the visual content is altered, but the text query remains fixed, \ie, \(\diouvis \!=\!\text{IoU}_{\text{fact}} - \text{IoU}_{\text{visual}} \). This metric evaluates the model's sensitivity to counterfactual visual edits. Smaller values indicate persistent hallucination despite the object's visual absence, revealing vulnerability to vision-driven errors.
Together, \(\dioutext\) and \(\diouvis\) provide orthogonal lenses to diagnose hallucination sources: the former exposes language-driven hallucinations, while the latter highlights vision-driven hallucinations. 
Our experiments demonstrate that pixel-grounding segmentation reasoning models often exhibit lower \(\diouvis\) values, underscoring the unique capability of \benchmarkname{} to elicit hallucination behaviors that remain overlooked when evaluating only under label replacements.

\noindent \textbf{Counterfactual Metrics.}
While consistency-based metrics measure model performance within object-aligned regions, they do not fully capture the spatial extent and structure of hallucinated predictions beyond the boundaries of the ground-truth mask. To address this limitation, we introduce metrics that directly penalize spurious mask predictions in scenarios where the queried object class is intentionally absent. These quantify both the presence of hallucinations and their spatial alignment with other objects in the scene, providing a more granular view of segmentation failures.

A natural starting point for assessing segmentation quality is the Tversky index~\citep{tversky1977features}, a generalization of the IoU metric that balances the trade-off between false positives (FP) and false negatives (FN). Tversky measures the overlap between a predicted mask \(\hat{\mathbf{M}}\) and a ground truth mask \(\mathbf{M}\) as follows:
\begin{equation}
    \scalemath{0.8}{\text{Tversky}(\hat{\mathbf{M}}, \mathbf{M}; \alpha, \beta) \!=\!
    \frac{\lvert \hat{\mathbf{M}} \cap \mathbf{M} \rvert}{\lvert \hat{\mathbf{M}} \cap \mathbf{M} \rvert + \alpha\, \lvert \hat{\mathbf{M}} \setminus \mathbf{M} \rvert + \beta\, \lvert \mathbf{M} \setminus \hat{\mathbf{M}} \rvert}},
\label{eq:tversky}
\end{equation}
where \(|\hat{\mathbf{M}} \cap \mathbf{M}|\) represents True Positives (TP), \(|\hat{\mathbf{M}} \setminus \mathbf{M}|\) False Positives (FP), \(|\mathbf{M} \setminus \hat{\mathbf{M}}|\) False Negatives (FN), and parameters \(\alpha\) and \(\beta\) control the relative penalty for FP and FN. However, in our counterfactual setting, the queried object class is explicitly removed from the image, resulting in an empty ground truth mask (\(\mathbf{M}\!=\!\varnothing\)). Under these conditions, both TP and FN are zero, causing the Tversky index to simplify to \(\text{Tversky}(\hat{\mathbf{M}}, \varnothing; \alpha, \beta)\!=\!0/\alpha\, |\hat{\mathbf{M}}|\). This collapses to zero for any non-empty prediction \(\hat{\mathbf{M}}\), failing to distinguish between structured hallucinations, where the model predicts plausible objects that overlap with distractors, and unstructured noise, where the model generates arbitrary masks with no semantic alignment. This limitation makes the Tversky index unsuitable for diagnosing hallucinations in counterfactual segmentation, where the absence of the queried object is guaranteed by design. 
Thus, we introduce the \textbf{Confusion Mask Score (\(\cms)\)}, which, unlike Tversky, \(\cms\) is designed to distinguish between hallucinated predictions that overlap with semantically confounding regions and those that do not. Given a factual image \(\mathbf{I}\) and a predicted mask \(\hat{\mathbf{M}}_{c'}\) for a query \(c'\) that is not present, we measure its overlap with the ground truth mask \(\mathbf{M}_c\) of the present query \(c\). We decompose the hallucinated prediction into two components: \(\mathbf{C}\!=\!\hat{\mathbf{M}}_{c'} \cap \mathbf{M}_c \) (confusing region) and \(\mathbf{N}\!=\!\hat{\mathbf{M}}_{c'} \setminus \mathbf{M}_c \)  (non-overlapping region). The Confusion Mask Score (\(\cms\)) is then computed as follows:
\begin{equation}
\setlength{\abovedisplayskip}{5pt}   
\setlength{\belowdisplayskip}{4pt}   
\cms\!=\!\frac{\alpha\,|\mathbf{C}| + |\mathbf{N}|}{\alpha\,|\mathbf{M}_c|},
\label{eq:cms}
\end{equation}
Here, \(\alpha > 1\) penalizes overlap with confusing regions more heavily than with unrelated areas. 
We evaluate \(\cms\) in two complementary settings: the \emph{factual} case, where the model is queried with \(c'\) on the factual image \(\mathbf{I}\), and the \emph{counterfactual} case, where it is queried with the original class \(c\) on the counterfactual image \(\mathbf{I}'\). We refer to these as \(\cmsfact\) and \(\cmscf\), respectively.

%% file: assets/figures/dataset_statistics.tex
\setcounter{figure}{2}
\begin{figure*}[t!]
    \centering
        \begin{subfigure}[t]{0.4\textwidth}
        \centering
        \includegraphics[width=\linewidth]{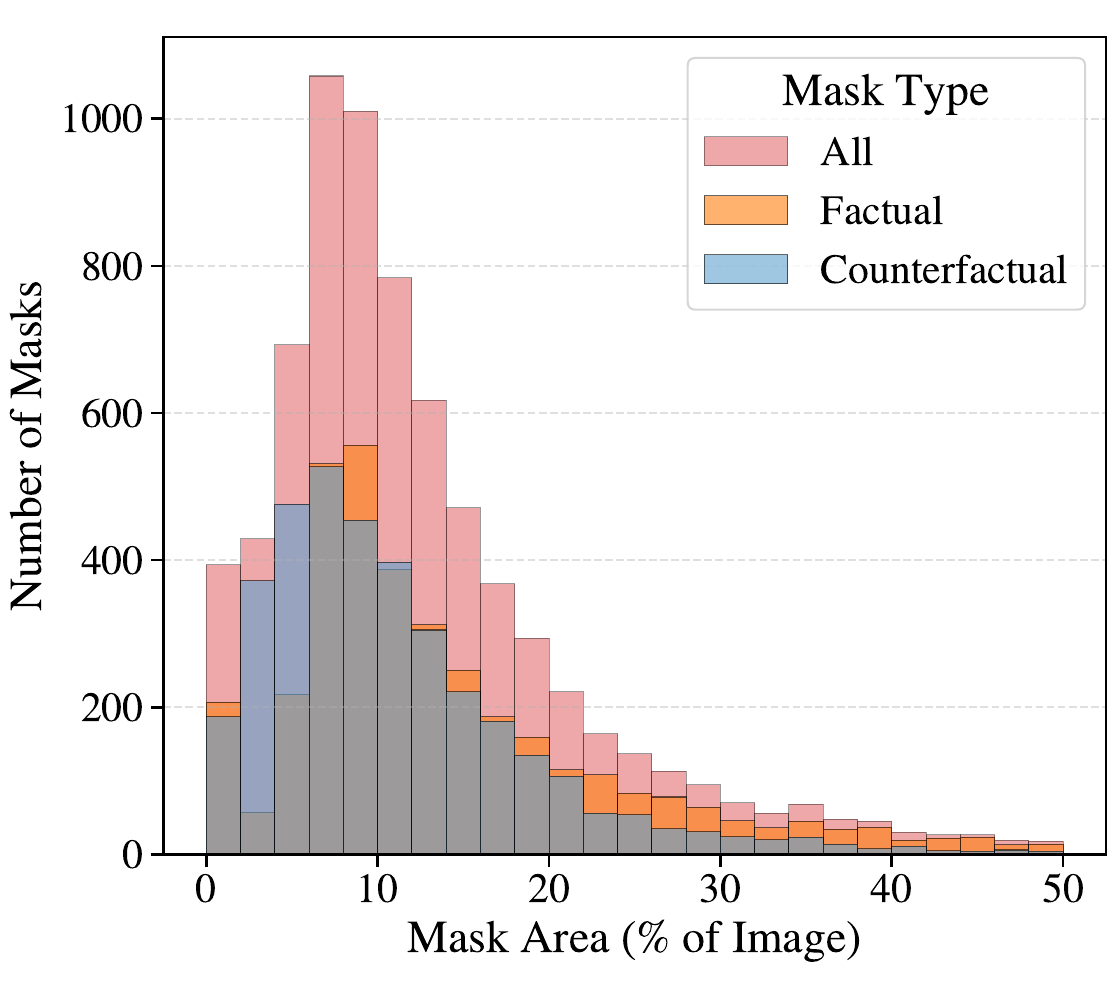}
            \vspace{-0.5cm}
        \caption{Mask Area Distribution}
        \label{fig:dataset_a}
    \end{subfigure}
    \hfill
    \begin{subfigure}[t]{0.52\textwidth}
        \centering
        \includegraphics[width=\linewidth]{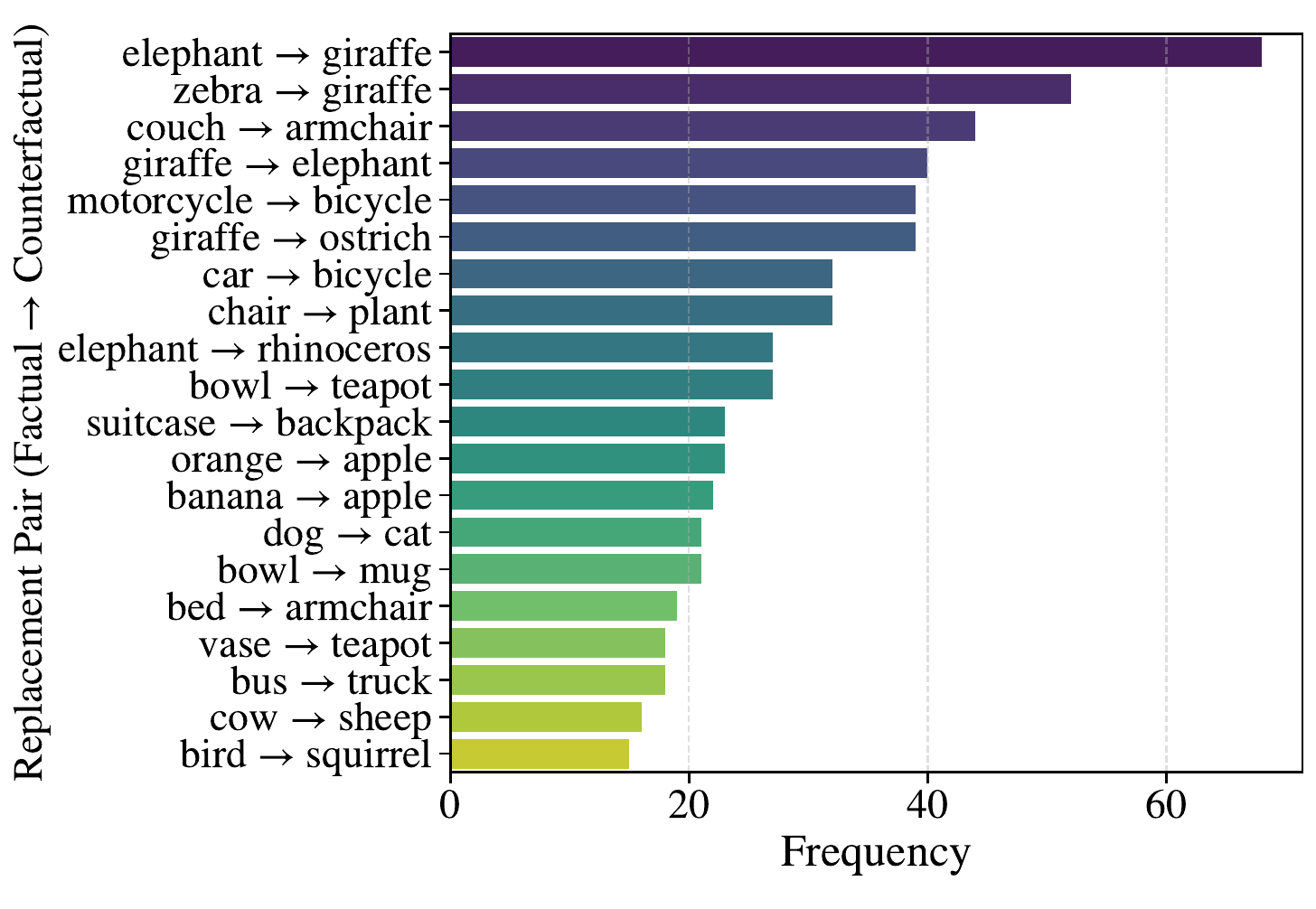}
            \vspace{-0.5cm}
        \caption{Factual-Counterfactual Replacement Pairs}
        \label{fig:dataset_b}
    \end{subfigure}
    \vspace{-0.2cm}
    \caption{\textbf{\benchmarkname{} Dataset Overview.} (a) Distribution of mask sizes as a percentage of the total image area, categorized by mask type. (b) Top-20 most frequent factual-counterfactual object replacement pairs, illustrating common substitution patterns in the dataset.}
    \label{fig:dataset-overview}
\end{figure*}

%% file: assets/figures/dataset_donuts.tex
\begin{figure*}[t!]
    \centering
    \begin{subfigure}[t]{0.24\textwidth}
      \centering
      \includegraphics[width=\linewidth,page=1,trim=2pt 2pt 2pt 2pt,clip]{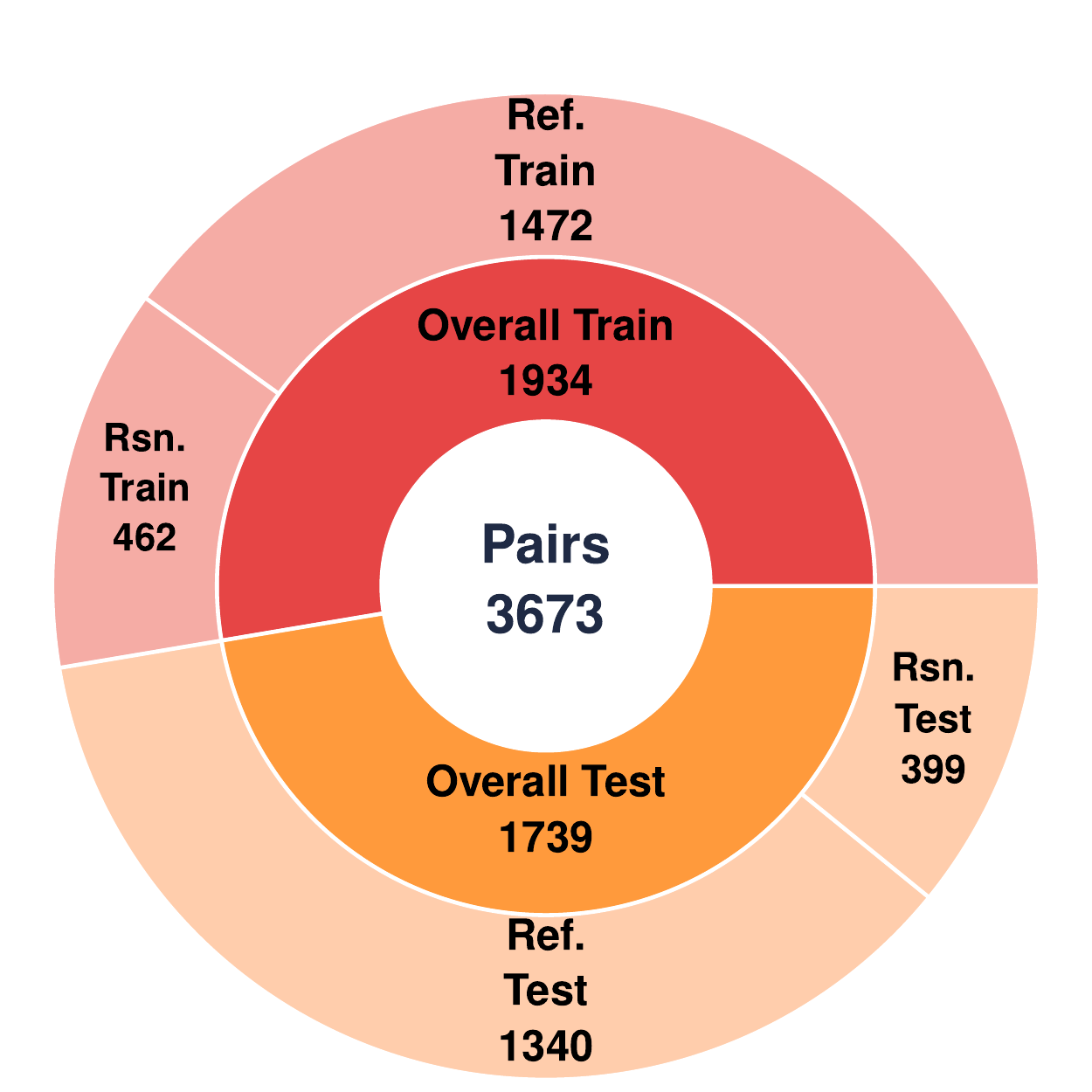}
      \caption{\# instance pairs}
      \label{fig:pairs_donut}
    \end{subfigure}\hfill
    \begin{subfigure}[t]{0.24\textwidth}
      \centering
      \includegraphics[width=\linewidth,page=1,trim=2pt 2pt 2pt 2pt,clip]{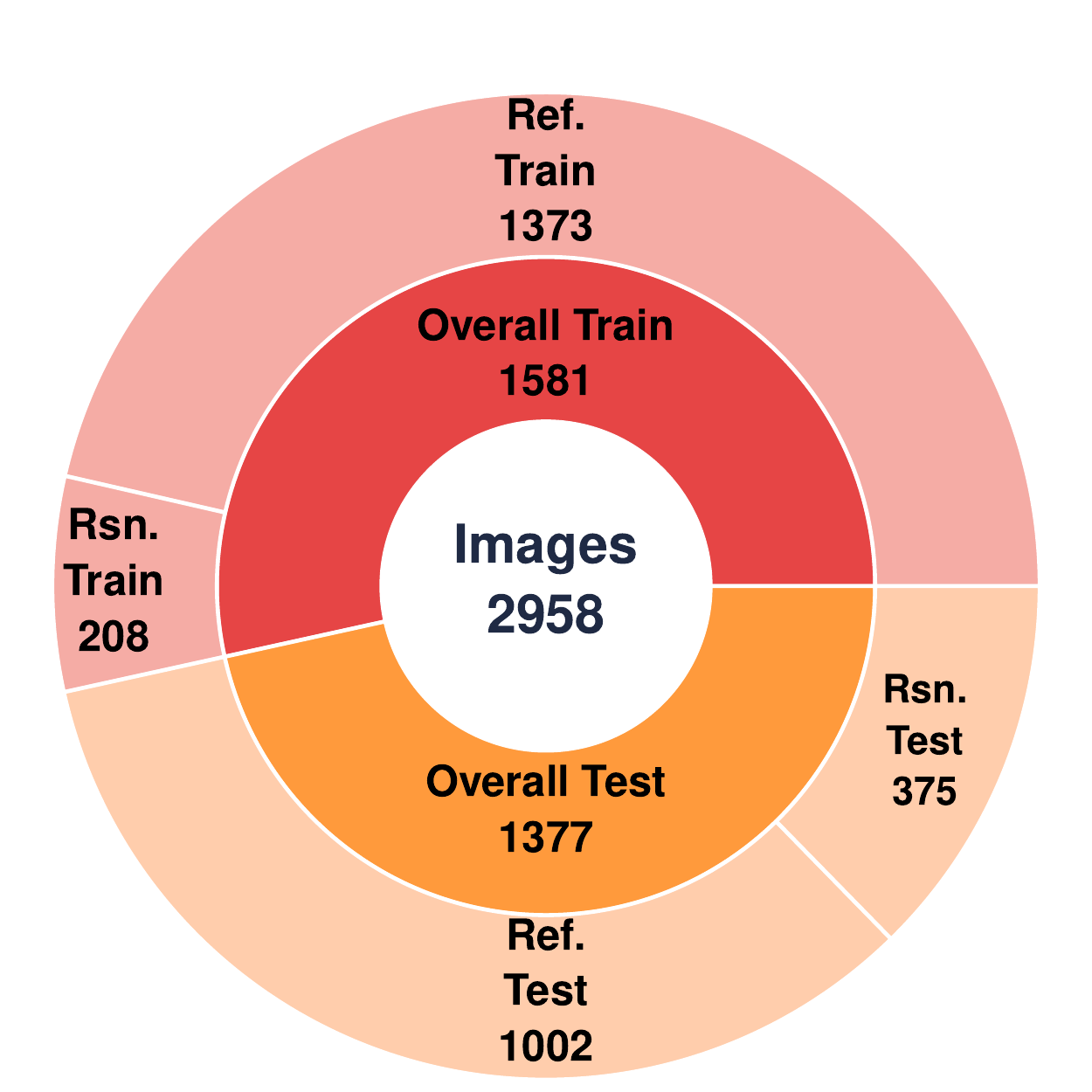}
      \caption{\# images}
      \label{fig:images_donut}
    \end{subfigure}\hfill
    \begin{subfigure}[t]{0.24\textwidth}
      \centering
      \includegraphics[width=\linewidth,page=1,trim=2pt 2pt 2pt 2pt,clip]{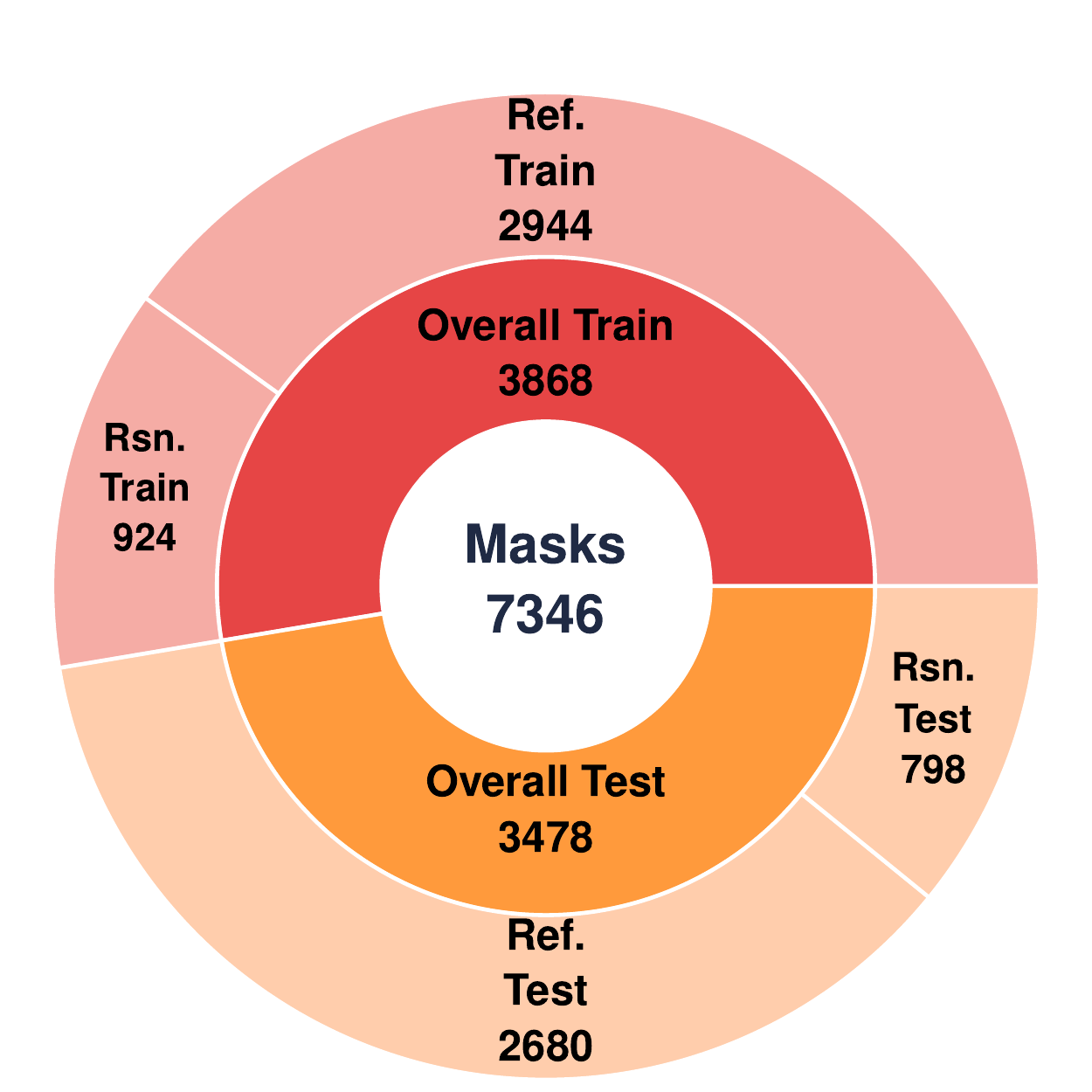}
      \caption{\# segmentation masks}
      \label{fig:masks_donut}
    \end{subfigure}\hfill
    \begin{subfigure}[t]{0.24\textwidth}
      \centering
      \includegraphics[width=\linewidth,page=1,trim=2pt 2pt 2pt 2pt,clip]{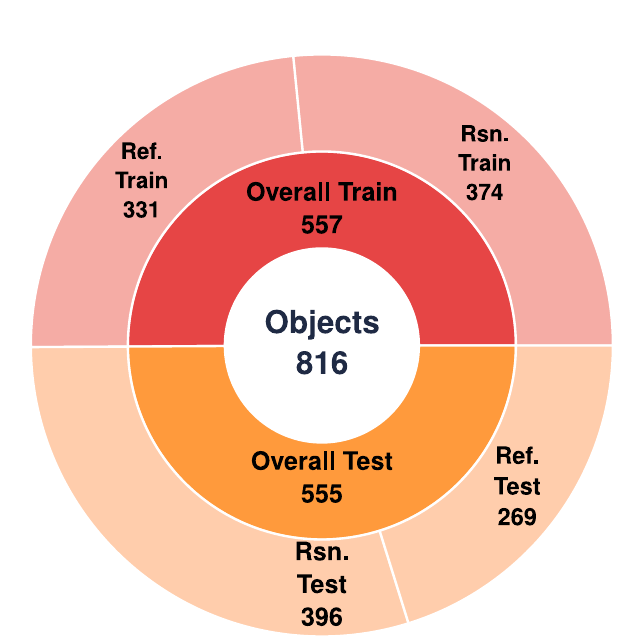}
      \caption{\# unique object categories}
      \label{fig:unique_donut}
    \end{subfigure}
    \vspace{-0.3cm}
    \caption{\textbf{\benchmarkname{} Dataset Statistics.} Sunbursts summarizing the distribution of (a) instance pairs, (b) images, (c) segmentation masks, and (d) unique object categories across referring and reasoning splits. Inner rings represent the overall train/test composition, while outer rings break down contributions from the RefCOCO (referring, shown as “Ref.”) and ReasonSeg (reasoning, shown as “Rsn.”) subsets. 
    }
\end{figure*}

%% file: assets/figures/cft.tex
\begin{figure*}[t!]
  \centering
  \includegraphics[width=0.99\textwidth]{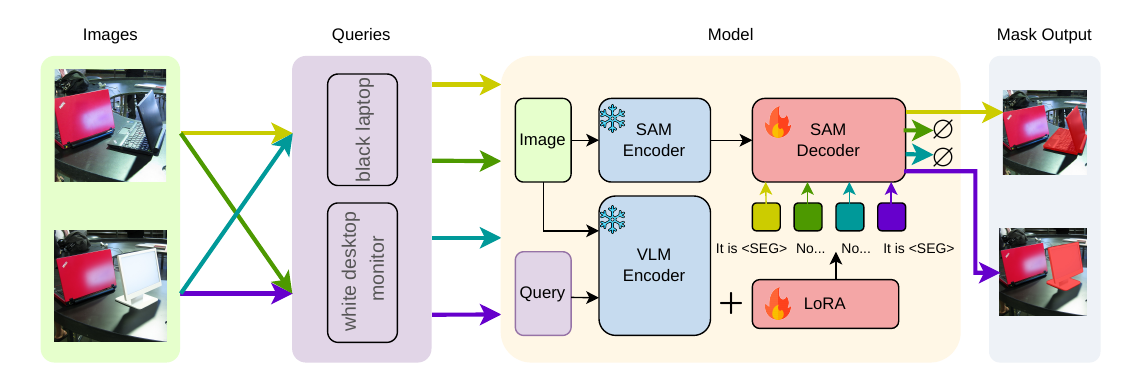}
  \vspace{-0.4cm}
\caption{\textbf{\modelname{} architecture and counterfactual finetuning pipeline.} Given a factual–counterfactual image pair, we construct four image–query combinations. LoRA-based CFT jointly applies positive supervision when the queried object exists and negative supervision when it does not, encouraging the model to segment only when visual evidence is present and suppress hallucination otherwise.}
  \label{fig:model_cft}
      \vspace{-0.2cm}
\end{figure*}

%% file: sections/4_model.tex
\input{assets/tables/hallusegbench}

\section{\modelnamegradient{} VLM}
\modelname{} is a vision–language segmentation model designed to reduce pixel-grounding hallucinations by alignment across factual and counterfactual image–text pairs. The model consists of three main components: a vision encoder, a text encoder, and a segmentation head, followed by our proposed counterfactual finetuning (CFT) stage. The model architecture is illustrated in \Cref{fig:model_cft}.
To mitigate hallucination, we introduce \textbf{Counterfactual Finetuning (\cft)}, which enforces consistent segmentation reasoning across factual–counterfactual vision-text pairs. 

Specifically, we finetune our \modelname{} VLM, denoted as \(f\), using a unified counterfactual loss \(\mathcal{L}_{\text{CF}}\), that jointly optimizes the model across all four configurations. 
Each training instance in \benchmarkname{} provides a factual image \(\mathbf{I}\), its counterfactual variant \(\mathbf{I}'\), along with the queries \(c\) and \(c'\), forming the quartet \((\mathbf{I}, c)\), \((\mathbf{I}, c')\), \((\mathbf{I}', c)\), and \((\mathbf{I}', c')\), as shown in \Cref{fig:model_cft}. 
For any image-query pair \((\mathbf{I}, q)\) with target reasoning tokens \(\mathbf{y}^{\text{text}}\) and target mask \(\mathbf{M}\), the overall counterfactual finetuning objective \(\mathcal{L}_{\text{CF}}\) is expressed as a positive-negative decomposition $\mathcal{L}_{\text{CF}}
= \mathcal{L}_{\text{pos}} + \lambda_{\text{neg}}\,\mathcal{L}_{\text{neg}}$,
where $\lambda_{\text{neg}}$ controls the strength of hallucination suppression.
\noindent
The \emph{positive branch} enforces correct behavior when the object is present in the given scene:
\begin{equation}
\setlength{\abovedisplayskip}{7pt}   
\setlength{\belowdisplayskip}{7pt}   
\mathcal{L}_{\text{pos}}
= \underbrace{\mathcal{L}_{\text{pair}}(\mathbf{I}, c, \mathbf{M}_{c}, \hat{\mathbf{M}_{c}})}_{\text{Factual Positive}}
+ \underbrace{\mathcal{L}_{\text{pair}}(\mathbf{I}', c', \mathbf{M}_{c'}, \hat{\mathbf{M'}_{c'}})}_{\text{Counterfactual Positive}},
\end{equation}
while the \emph{negative branch} suppresses hallucinations under text-only and vision-only counterfactuals:
\begin{equation}
\setlength{\abovedisplayskip}{7pt} 
\setlength{\belowdisplayskip}{7pt}
\mathcal{L}_{\text{neg}}
= \underbrace{\mathcal{L}_{\text{pair}}(\mathbf{I}, c', \emptyset, \hat{\mathbf{M}_{c'}})}_{\text{Text-Negative}}
+ \underbrace{\mathcal{L}_{\text{pair}}(\mathbf{I}', c, \emptyset, \hat{\mathbf{M'}_{c}})}_{\text{Vision-Negative}}.
\end{equation}
Here,  \(\mathcal{L}_{\text{pair}}\) is a composite loss defined as:
\begin{equation}
\setlength{\abovedisplayskip}{7pt} 
\setlength{\belowdisplayskip}{7pt}
 \mathcal{L}_{\text{pair}}
= \lambda_{\text{text}} \mathcal{L}_{\text{text}}(\mathbf{I}, q) 
+ \lambda_{\text{seg}} \mathcal{L}_{\text{seg}}(\mathbf{I}, q, \mathbf{M}, \hat{\mathbf{M}}).
\label{eq:pair_loss}
\end{equation}

\noindent
The text loss \(\mathcal{L}_{\text{text}}\) is a standard Cross Entropy (CE) loss $\mathcal{L}_{\text{text}}
= \mathcal{L}_{\text{CE}}\!\big(\hat{\mathbf{y}}^{\text{text}}_{\mathbf{I},q}, \mathbf{y}^{\text{text}}_{\mathbf{I},q}\big)$,
where \(\hat{\mathbf{y}}^{\text{text}}_{\mathbf{I},q}\) denotes the token distribution of the model for input \((\mathbf{I}, q)\) and \(\mathbf{y}^{\text{text}}_{\mathbf{I},q}\) to the target text.
The segmentation loss combines Binary Cross Entropy (BCE) and Dice loss $\mathcal{L}_{\text{seg}}\!=\!\lambda_{\text{BCE}} \mathcal{L}_{\text{BCE}}\!\big(\hat{\mathbf{M}}, \mathbf{M}\big) + \lambda_{\text{Dice}} \mathcal{L}_{\text{Dice}} \!\big(\hat{\mathbf{M}}, \mathbf{M}\big)$
where \(\hat{\mathbf{M}}\) is the predicted mask and \(\mathbf{M}\) is the ground truth mask. 
For negative pairs, the ground truth mask does not exist, \ie, \(\mathbf{M} = \emptyset\), and the target text
corresponds to the ``no such object’’ answer.

%% file: assets/tables/hallusegbench.tex
\begin{table*}[t!]
    \centering
    \caption{\textbf{Comparison of Reasoning Segmentation Models on Hallucination and Consistency Metrics.} Metrics include factual and counterfactual Confusion Mask Scores (CMS \(\downarrow\)), textual and visual delta IoU (\(\Delta\)IoU\textsubscript{textual/visual} \(\uparrow\)). Hallucination Mitigation (HM) indicates whether the model is explicitly trained to suppress hallucinations. Best performance highlighted with \colorbox{LightBlue}{\phantom{\rule{1ex}{1ex}}}, second best underscored with \_.}
    \label{tab:combined_cms_iou}
    \vspace{-0.3cm}
    \resizebox{0.95\textwidth}{!}{%
    \begin{tabular}{lcccccccccccc}
        \toprule
        \multirow{2}{*}{\textbf{Model}} & \multirow{2}{*}{\textbf{HM}} 
        & \multicolumn{2}{c}{\textbf{CMS\textsubscript{Referring}}} 
        & \multicolumn{2}{c}{\textbf{CMS\textsubscript{Reasoning}}}
        & \multicolumn{2}{c}{\textbf{\(\Delta\)IoU Referring}} 
        & \multicolumn{2}{c}{\textbf{\(\Delta\)IoU Reasoning}} \\
        
        \cmidrule(lr){3-4} \cmidrule(lr){5-6} \cmidrule(lr){7-8} \cmidrule(lr){9-10}
        & & \textbf{Factual} \(\downarrow\) & \textbf{Counterfactual} \(\downarrow\)
        & \textbf{Factual} \(\downarrow\) & \textbf{Counterfactual} \(\downarrow\)
        & \textbf{Textual} \(\uparrow\) & \textbf{Visual} \(\uparrow\)
        & \textbf{Textual} \(\uparrow\) & \textbf{Visual} \(\uparrow\) \\
        \midrule

        \textbf{LISA-7B}~\cite{lai2024lisa} & \xmark & 0.3080 & 0.7317 & 0.4438 & 0.6960 & 0.4534 & 0.2810 & 0.2991 & 0.2418 \\    
        \textbf{PixelLM-7B}~\cite{ren2024pixellm} & \xmark & 0.4748 & 0.7286 & 0.4284 & 0.5824 & 0.3952 & 0.4071 & 0.2792 & 0.2263 \\        
        \textbf{GLaMM-7B}~\cite{rasheed2024glamm} & \xmark & 0.4196 & 0.6052 & 0.3556 & \underline{0.4022} & 0.3273 & 0.3016 & 0.2446 & \underline{0.3164} \\       
        \textbf{LISA-13B}~\cite{lai2024lisa} & \xmark & 0.3194 & 0.6687 & 0.4979 & 0.6647 & \underline{0.4591} & 0.3886 & \underline{0.3430} & \underline{0.3164} \\
        \textbf{PixelLM-13B}~\cite{ren2024pixellm} & \xmark & 0.4306 & 0.7253 & 0.4267 & 0.5424 & 0.4285 & 0.4273 & 0.3106 & 0.3079 \\ 
        \hline
        \textbf{Seg-Zero}~\cite{liu2025seg} & \cmark & 0.4482 & 0.5598 & 0.6554 & 0.7671 & 0.4412 & \colorbox{LightBlue}{\textbf{0.5312}} & 0.2391 & 0.3042 \\ 
        \textbf{VisionReasoner}~\cite{liu2025visionreasoner} & \cmark & 0.8191 & 0.6908 & 0.7315 & 0.8633 & 0.3619 & 0.4822 & 0.2464 & 0.3075 \\
        \textbf{SESAME-7B}~\cite{wu2024see} & \cmark & \underline{0.1983} & \underline{0.4304} & \underline{0.2998} & 0.4873 & 0.4180 & 0.3605 & 0.3296 & 0.3000 \\
        \textbf{\modelnamegradient{} (Ours)} & \cmark & \colorbox{LightBlue}{\textbf{0.1062}} & \colorbox{LightBlue}{\textbf{0.4049}} & \colorbox{LightBlue}{\textbf{0.1541}} & \colorbox{LightBlue}{\textbf{0.3988}} 
        & \colorbox{LightBlue}{\textbf{0.6105}} & \underline{0.5278} & \colorbox{LightBlue}{\textbf{0.4085}} & \colorbox{LightBlue}{\textbf{0.3184}} \\

        \bottomrule
    \end{tabular}
    }
\end{table*}

%% file: sections/5_experiments.tex
\section{Experiments}
\noindent \textbf{Baselines.} We evaluate \modelname{} against a range of pixel-grounding VLMs, including models explicitly designed to mitigate grounded hallucination. The reasoning-based models include \textbf{LISA}~\cite{lai2024lisa}, \textbf{GLaMM}~\cite{rasheed2024glamm}, and \textbf{PixelLM}~\cite{ren2024pixellm}, which leverage large language models for reasoning, and SAM~\cite{kirillov2023segment} or other Transformer-based architectures to perform open-vocabulary segmentation and grounding. In addition, we evaluate \textbf{SESAME}~\cite{wu2024see}, a model that is designed to mitigate segmentation hallucinations by finetuning on data with false-premise text queries. \textbf{Seg-Zero}~\cite{liu2025seg} and \textbf{VisionReasoner}~\cite{liu2025visionreasoner} incorporate Chain-of-Thought (CoT) reasoning to enable the model to generate explicit steps for object identification and localization. While these models demonstrate strong performance on standard benchmarks, they have not been evaluated under counterfactual settings that probe hallucination behavior directly.

\subsection{Quantitative Results on \benchmarkname{}}
We evaluate all baselines along with \modelname{} on \benchmarkname{} with our proposed \(\diou\) and \(\cms\) metrics, for both Referring and Reasoning subsets, across both textual and visual replacements. As shown in \Cref{tab:combined_cms_iou}, on the counterfactual hallucination metrics, \modelname{} consistently attains the lowest \( \cmsfact \) and \( \cmscf \) in both referring and reasoning, while achieving the highest consistency scores, \( \dioutext \) and \( \diouvis \). Relative to the strongest baseline, SESAME~\citep{wu2024see}, \modelname{} reduces \( \cmsfact \) by \({\sim}46\%\) in referring and \({\sim}49\%\) in reasoning while \( \cmscf \) also drops by \({\sim}6\%\) and \({\sim}18\%\), respectively. 

On hallucination metrics,
\modelname{} delivers state-of-the-art performance on three of the four \(\diou\) metrics achieving substantial gains over the strongest baseline \({\sim}19\%\) on Reasoning \(\dioutext\) \({\sim}0.6\%\) on Reasoning \(\diouvis\), and \({\sim}33\%\) on Referring \(\dioutext\). For Referring \(\diouvis\), \modelname{} shows a minimal gap relative to Seg‑Zero (\(0.5312\) vs. \(0.5278\)), indicating that counterfactual finetuning introduces negligible trade‑offs.

Trends across the baseline models also provide key insights into the hallucination behavior of segmentation VLMs.
Scaling up model size (7B→13B LISA~\cite{lai2024lisa}, PixelLM~\cite{ren2024pixellm}) yields moderate improvements in consistency metrics such as \(\diouvis\), yet fails to significantly reduce hallucinations, with \(\cms\) values remaining high. CoT-based models (Seg-Zero~\cite{liu2025seg}, VisionReasoner~\cite{liu2025visionreasoner}) are more prone to hallucination, as reflected in their increased \(\cmsfact\) scores, particularly in the absence of strong visual cues, likely due to multi-step inference mechanisms, which are designed to enhance compositional reasoning and may inadvertently amplify textual biases over visual evidence.

\input{assets/figures/qualitative_main}

\subsection{Qualitative Analysis}
\Cref{fig:qualitative} shows model predictions across the four query-image combinations, along with the corresponding ground truth masks. Consistent with our quantitative findings, all models exhibit grounding hallucinations in the \((\mathbf{I}', c)\) setting, where query \(c\) has been explicitly removed from the scene. Among the baselines, both LISA variants show better hallucination abstention in the \((\mathbf{I}, c')\) case, aligning with their stronger \(\diouvis\) scores. SESAME exhibits generally lower hallucination, as reflected in its low \(\cms\) scores, but still fails under visually ambiguous conditions.
Overall, the qualitative results highlight a persistent inability of current models to distinguish grounded objects from plausible counterfactual distractors.

\input{assets/tables/fprefcoco}
\subsection{Results on FP-RefCOCO (+/g)}
To assess the generalization and robustness of our model, we evaluate \modelname{} on the augmented FP-RefCOCO(+/g), which evaluates model performance on text-based false premise cases. Prior works such as LISA~\cite{lai2024lisa}, Seg-Zero~\cite{liu2025seg}, and VisionReasoner~\cite {liu2025visionreasoner} exhibit limited performance on this task, primarily due to their reliance on large-scale pretraining without access to explicit negative supervision. Cascading~\cite{wu2024see}, a baseline by ~\citet{wu2024see}, and SESAME~\cite{wu2024see} perform comparatively better by incorporating hallucination-aware training signals. In particular, SESAME achieves high grounding accuracy and segmentation quality through targeted label replacement strategies that discourage incorrect mask predictions in ambiguous cases.
In contrast, \modelname{}, trained via counterfactual finetuning using paired factual and counterfactual supervision, achieves state-of-the-art performance across all settings by consistently improving both localization and segmentation while suppressing spurious mask predictions in the absence of supporting visual evidence. Compared to previous state-of-the-art, SESAME~\cite{wu2024see}, \modelname{} improves "See" accuracy by \(3.53\), \(3.00\), \(2.43\) percentage points on FP-RefCOCO, FP-RefCOCO+, FP-RefCOCOg respectively, while also yielding cIoU gains of \(1.64\), \(2.10\), \(0.97\) points. These gains confirm the effectiveness of counterfactual finetuning on \benchmarkname{} in mitigating hallucinations and enhancing grounding fidelity.

%% file: assets/figures/qualitative_main.tex
\begin{figure}[t!]
    \centering
    \setlength{\tabcolsep}{1pt}
    \resizebox{\columnwidth}{!}{
    \begin{tabular}{rcccc}
        & \(\mathbf{I}, c\) & \(\mathbf{I}, c'\) & \(\mathbf{I}', c\) & \(\mathbf{I}', c'\) \\

        \textbf{GT} &
        \includegraphics[height=1.8cm]{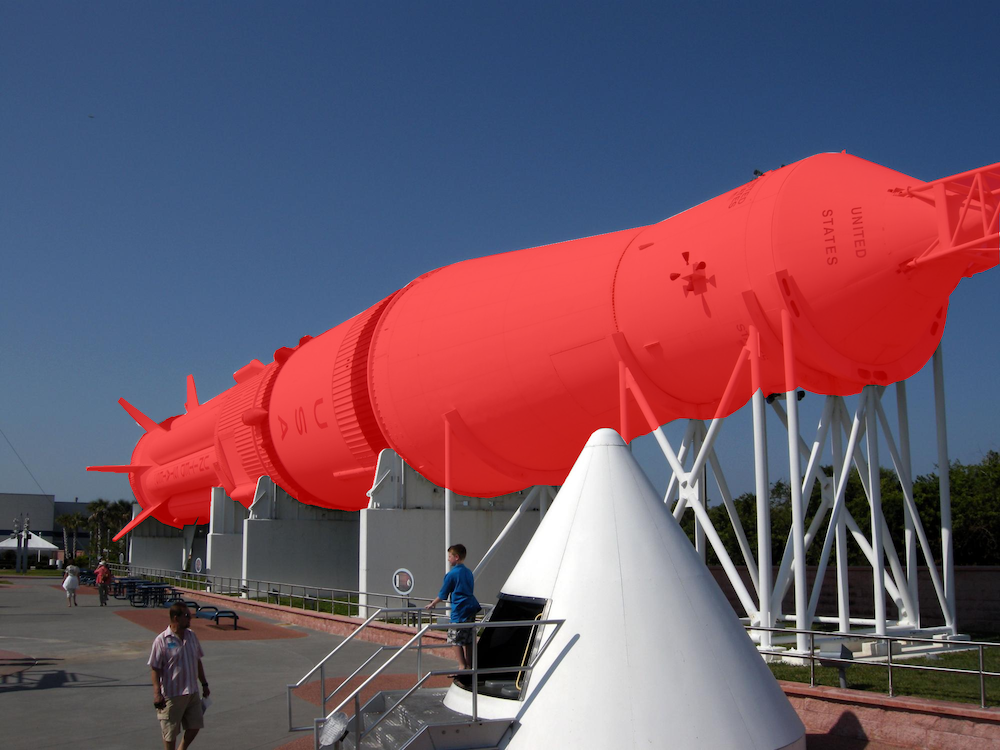} &
        \raisebox{0.6cm}{\(\boldsymbol{\varnothing}\)} &
        \raisebox{0.6cm}{\(\boldsymbol{\varnothing}\)} &
        \includegraphics[height=1.8cm]{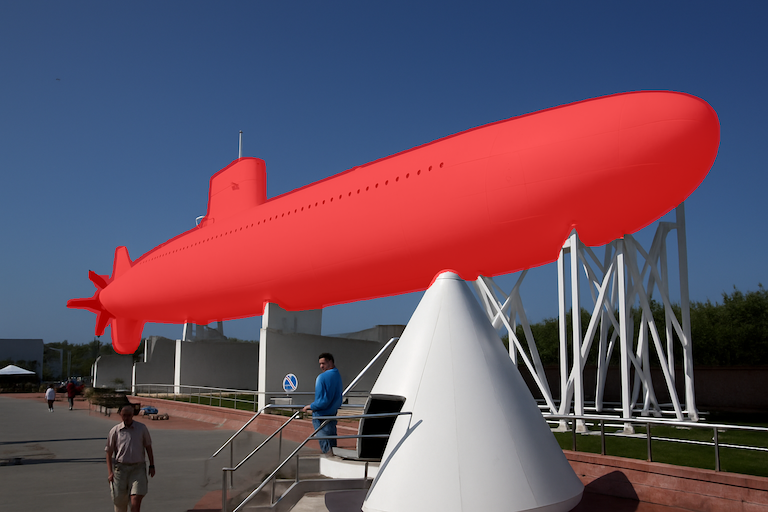} \\

        \textbf{LISA} &
        \includegraphics[height=1.8cm]{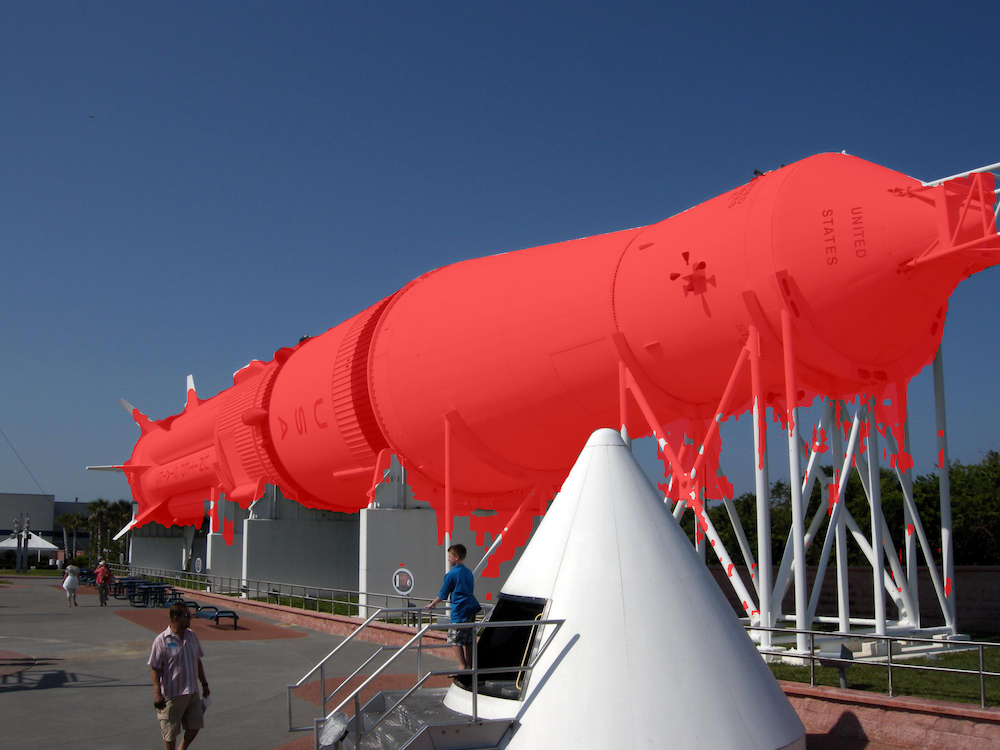} &
        \includegraphics[height=1.8cm]{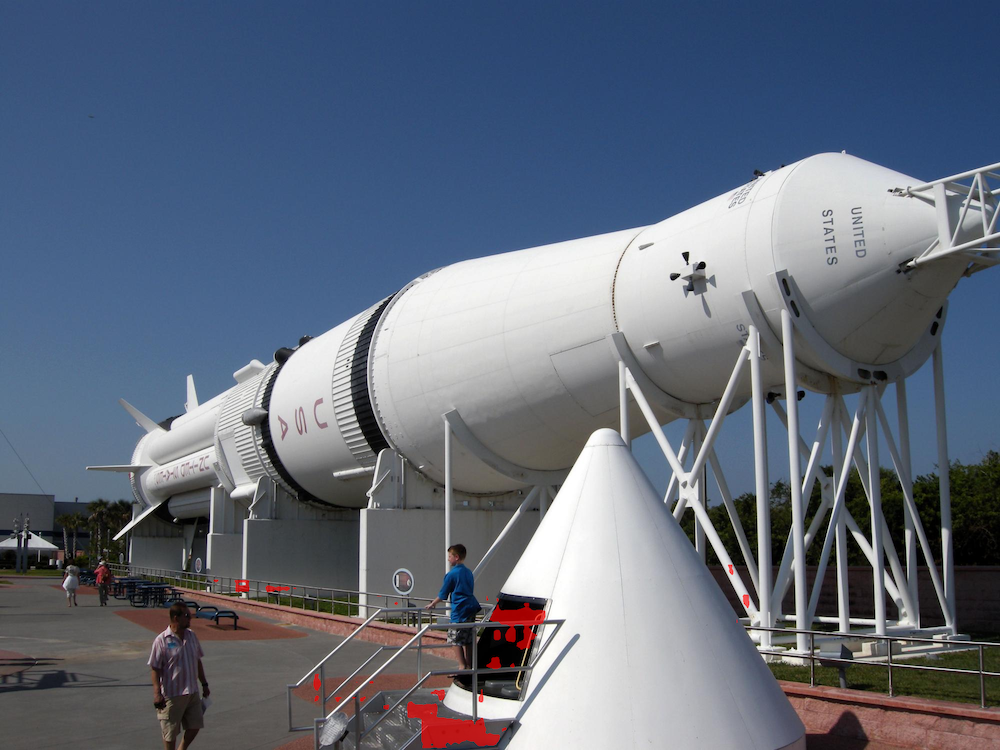} &
        \includegraphics[height=1.8cm]{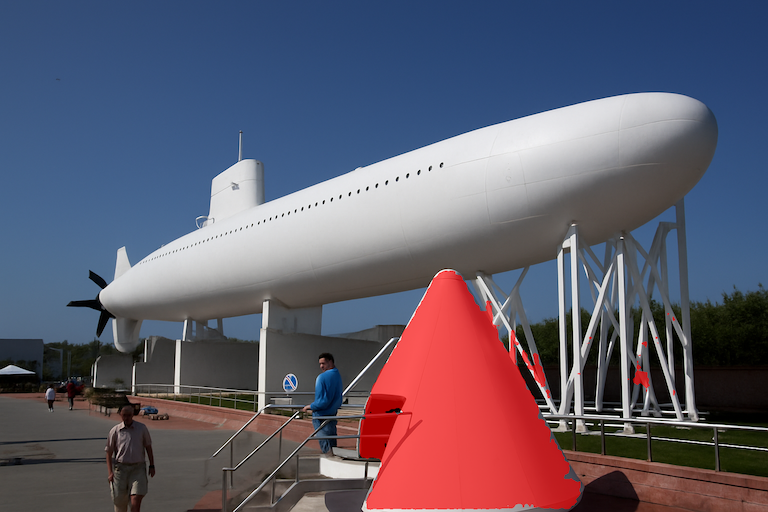} &
        \includegraphics[height=1.8cm]{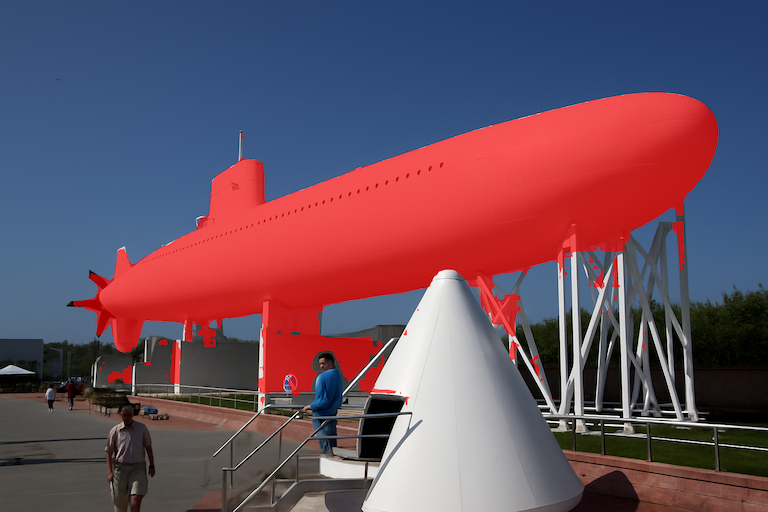} \\

        \textbf{PixelLM} &
        \includegraphics[height=1.8cm]{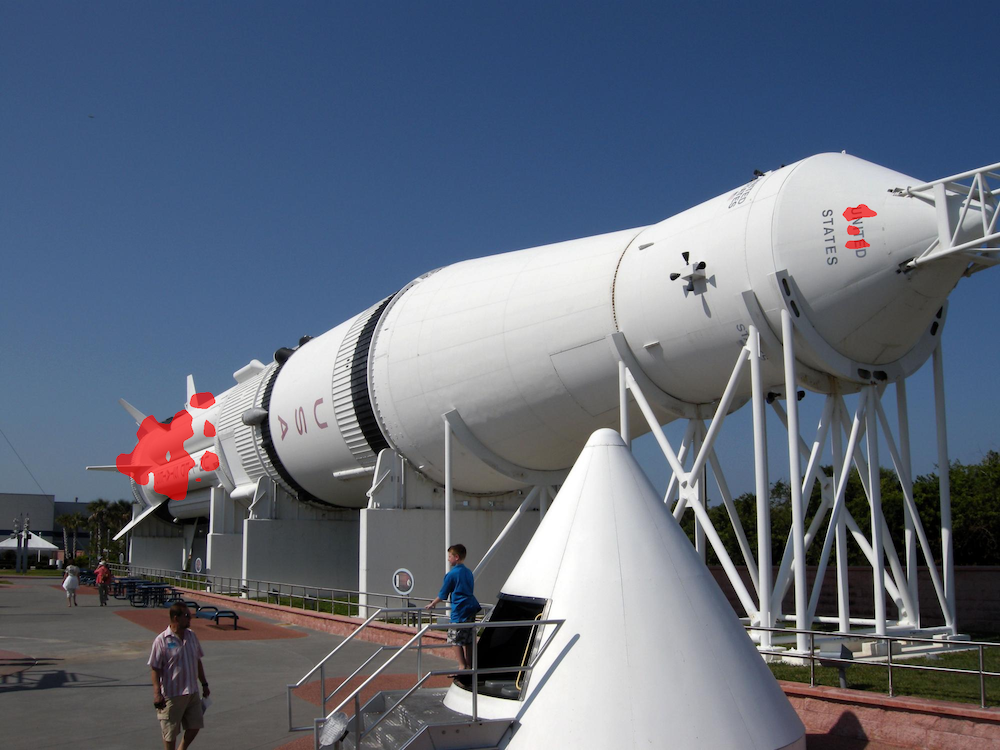} &
        \includegraphics[height=1.8cm]{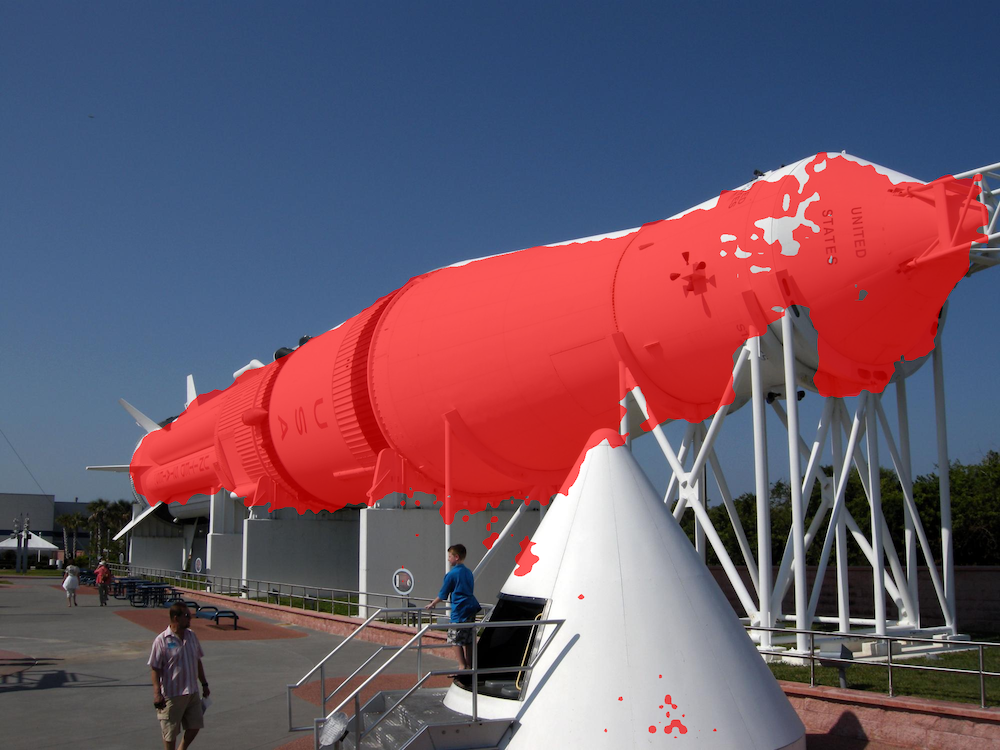} &
        \includegraphics[height=1.8cm]{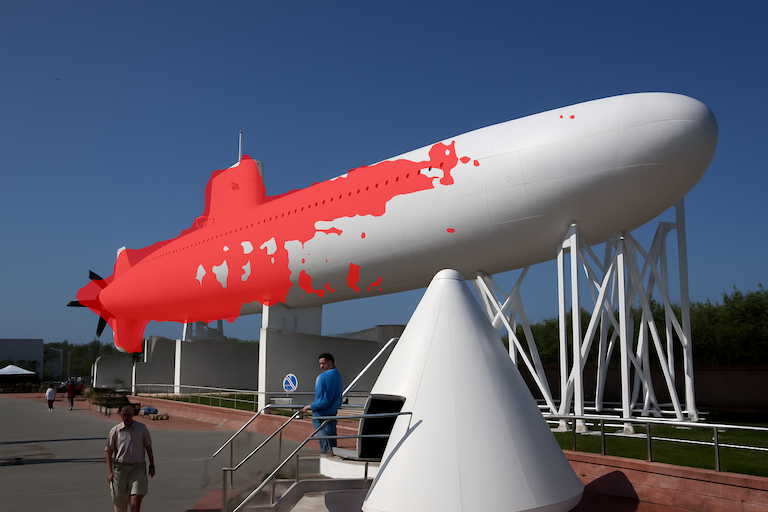} &
        \includegraphics[height=1.8cm]{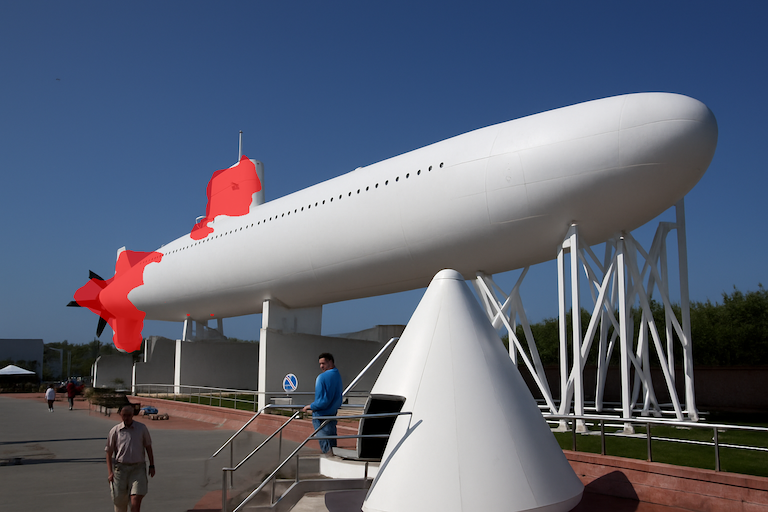} \\

        \textbf{GLaMM} &
        \includegraphics[height=1.8cm]{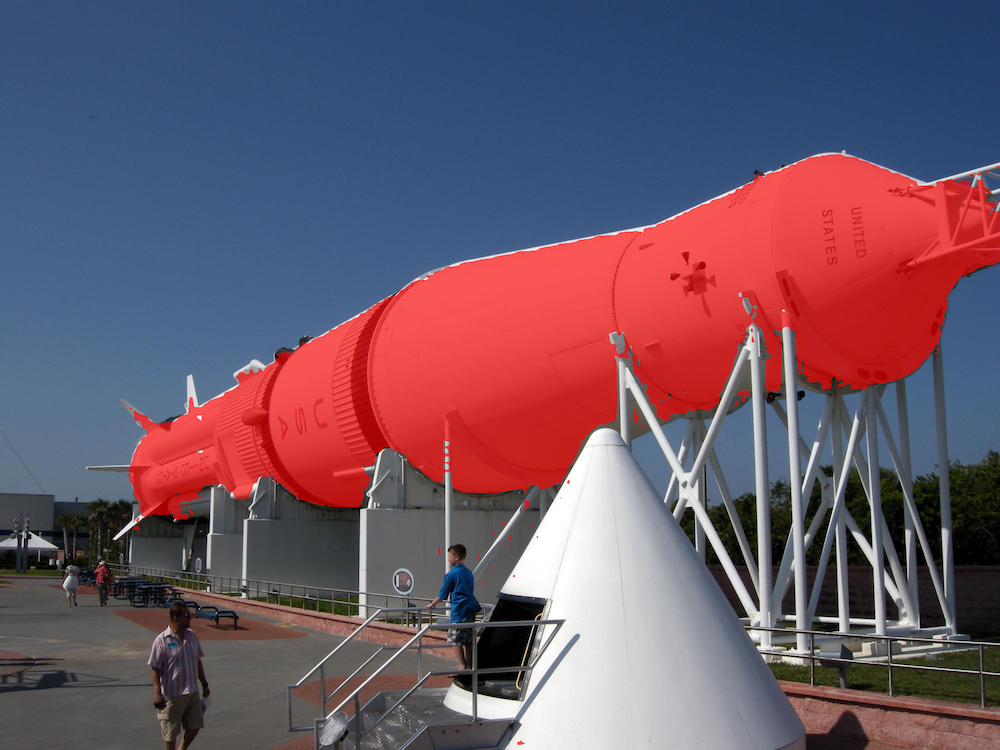} &
        \includegraphics[height=1.8cm]{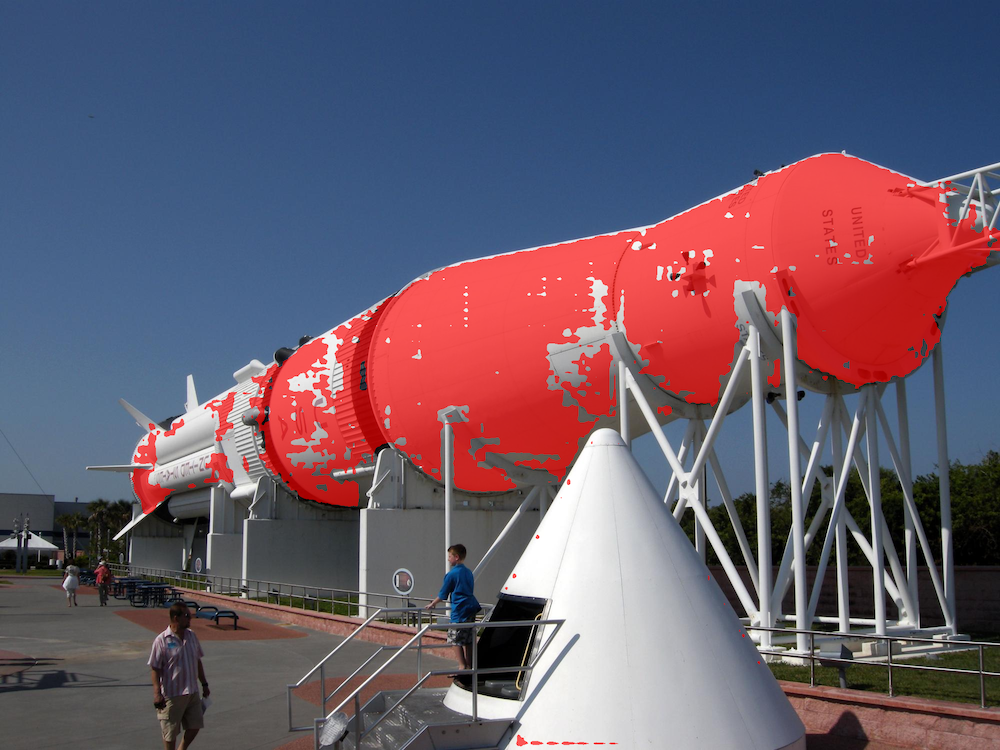} &
        \includegraphics[height=1.8cm]{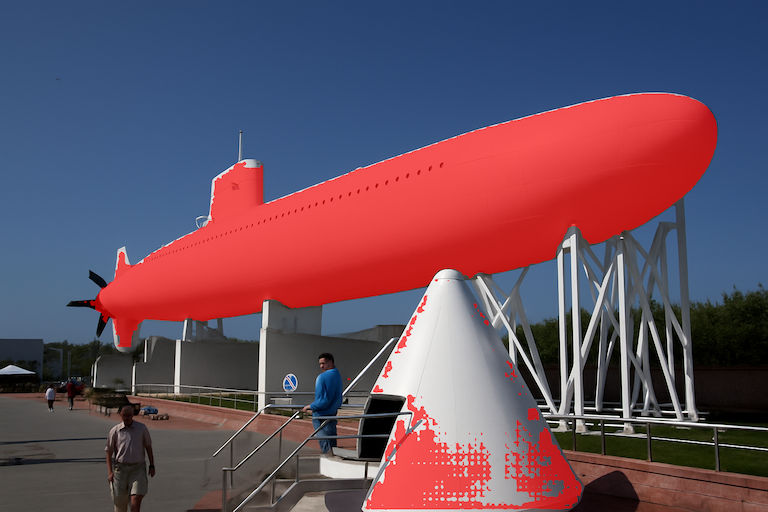} &
        \includegraphics[height=1.8cm]{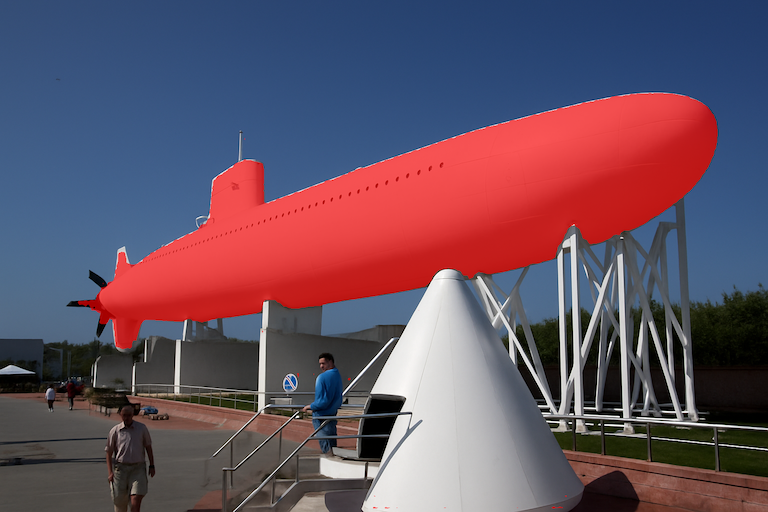} \\

        \textbf{Seg-Zero} &
        \includegraphics[height=1.8cm]{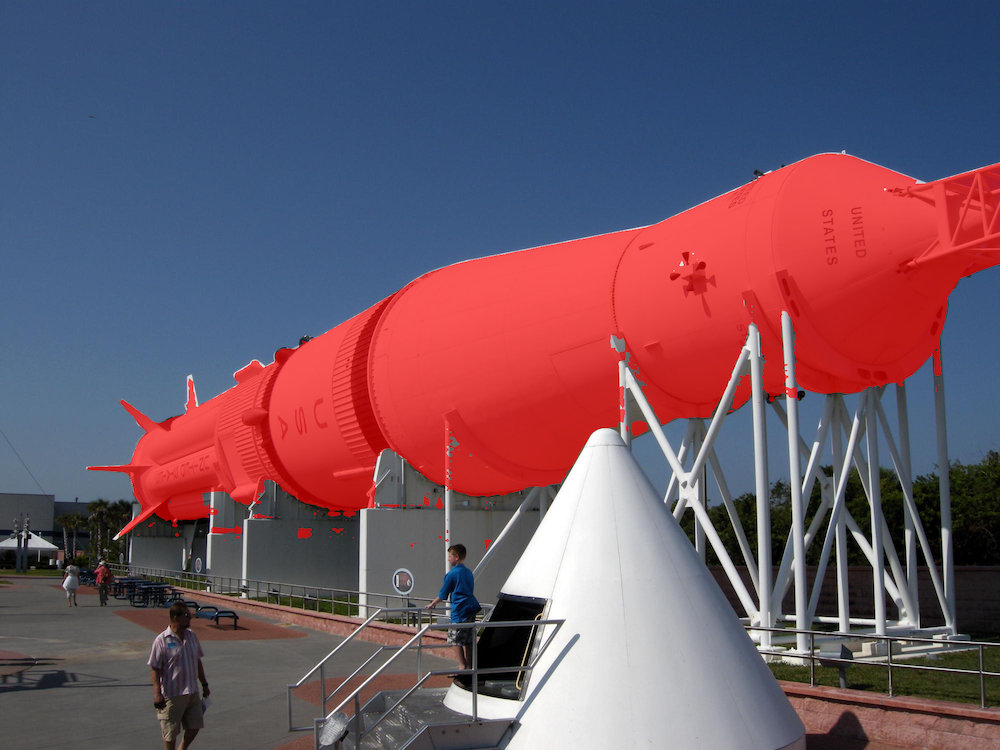} &
        \includegraphics[height=1.8cm]{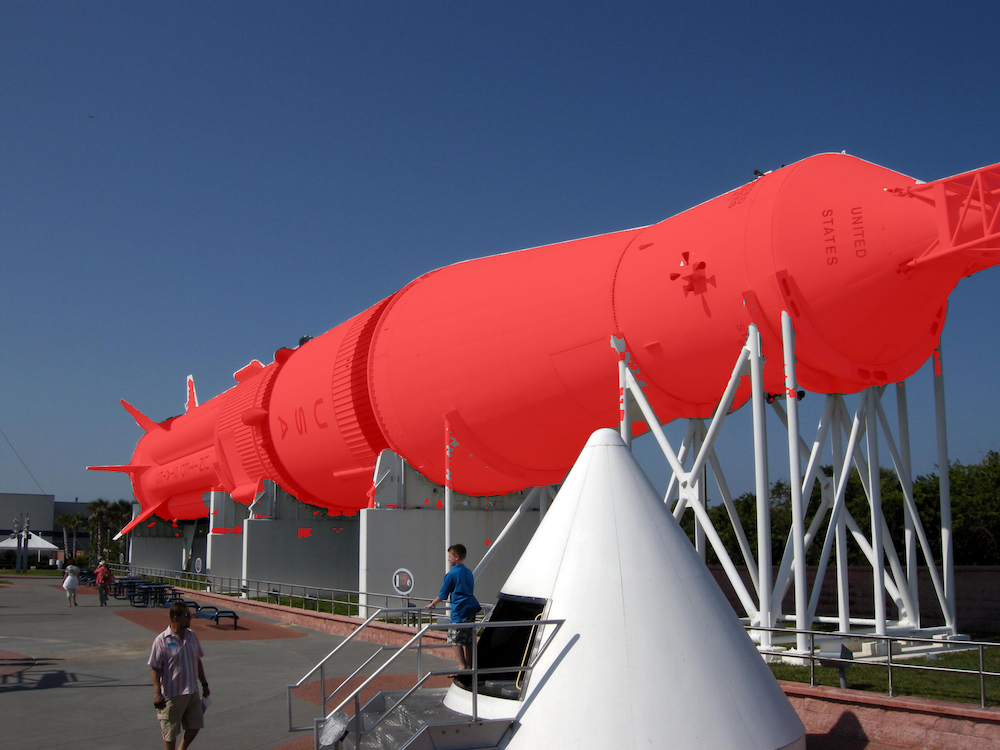} &
        \includegraphics[height=1.8cm]{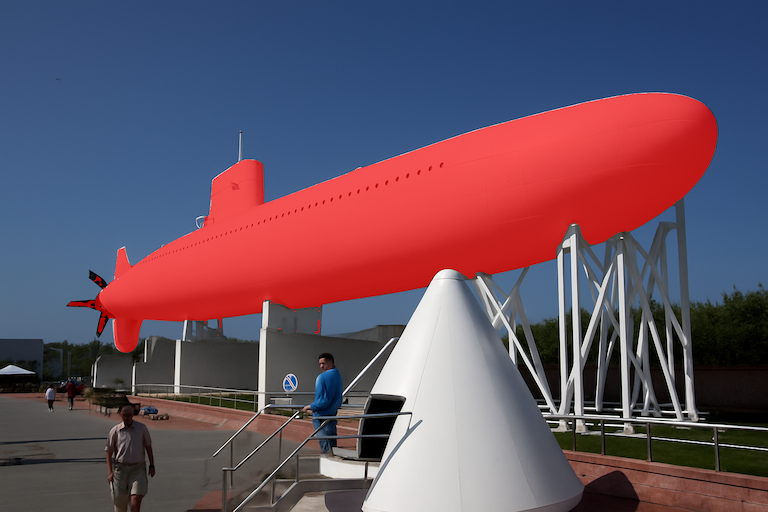} &
        \includegraphics[height=1.8cm]{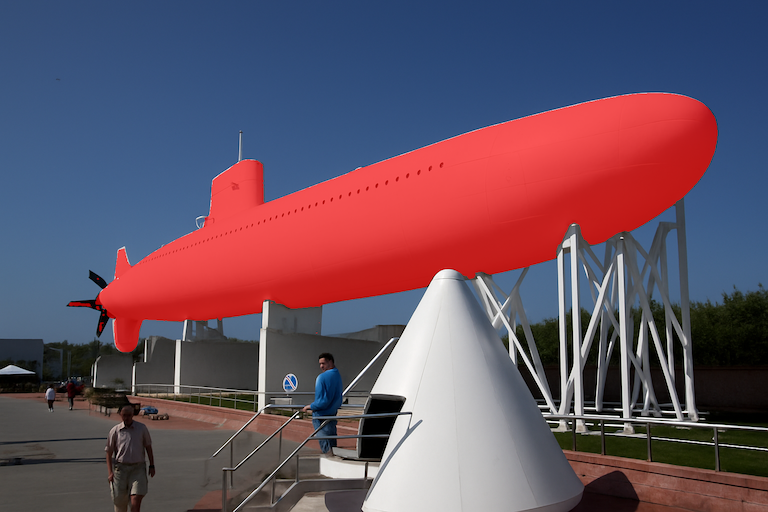} \\

        \textbf{SESAME} &
        \includegraphics[height=1.8cm]{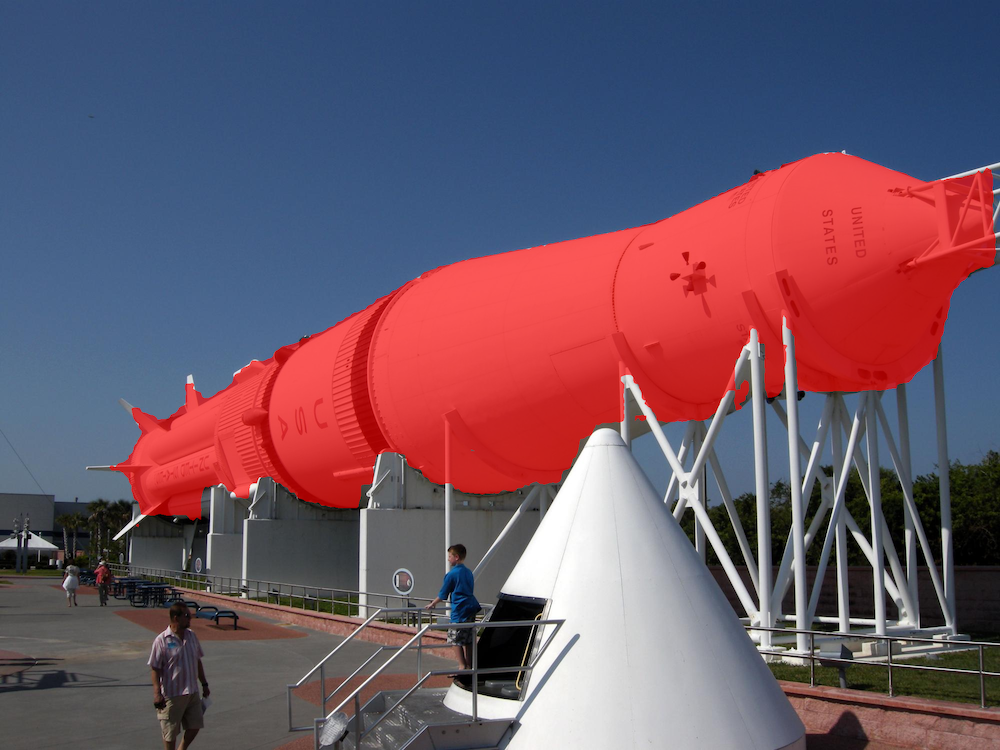} &
        \includegraphics[height=1.8cm]{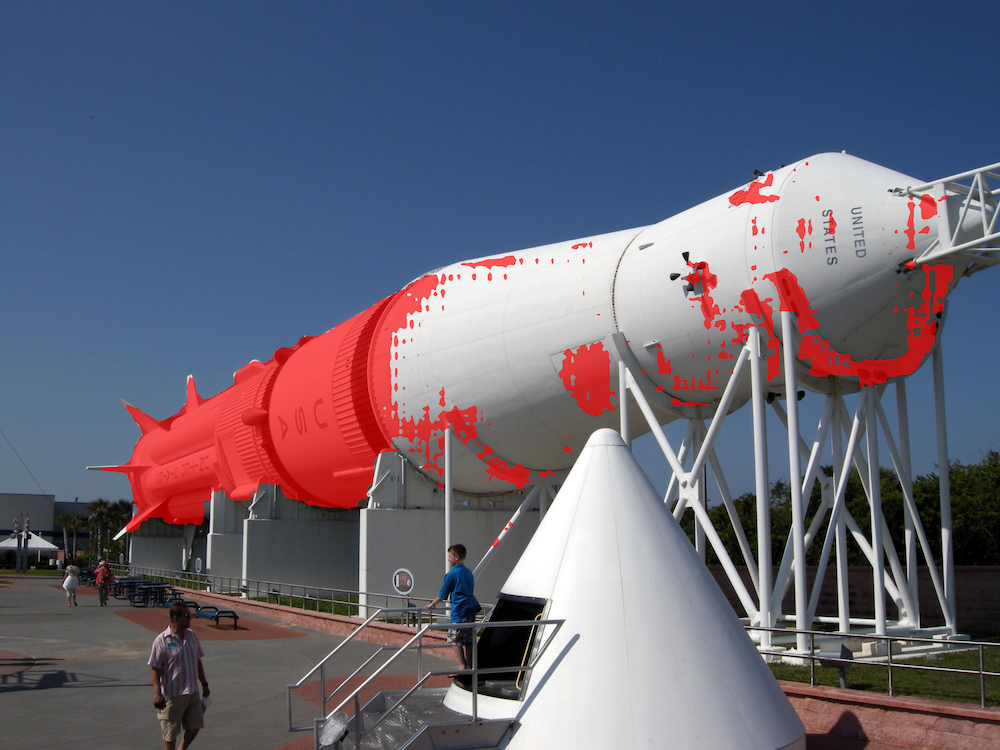} &
        \includegraphics[height=1.8cm]{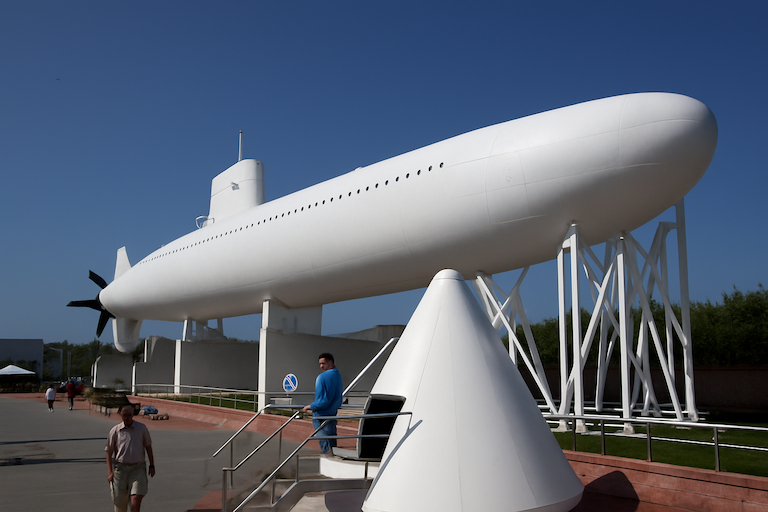} &
        \includegraphics[height=1.8cm]{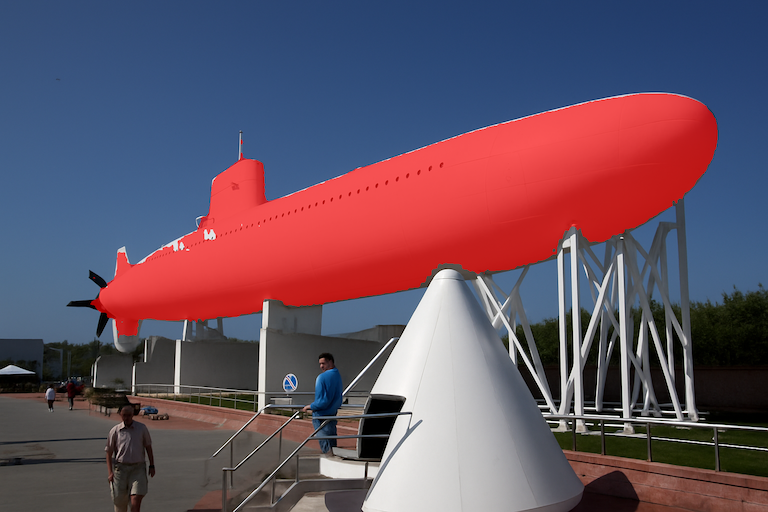} \\

        \textbf{\modelname} &
        \includegraphics[height=1.8cm]{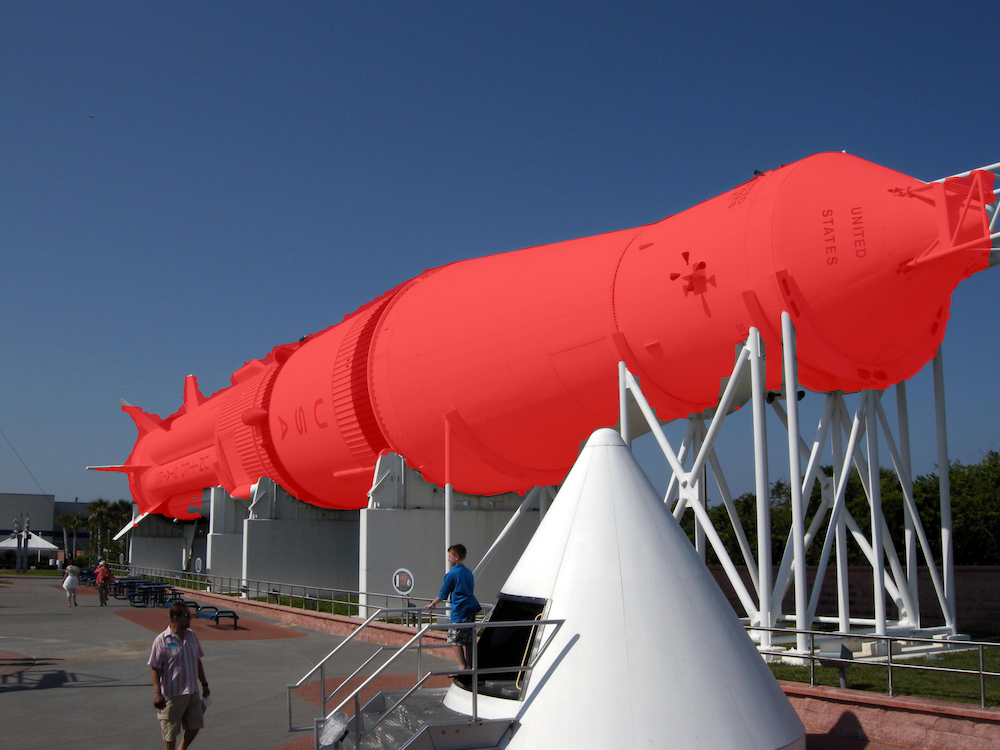} &
        \includegraphics[height=1.8cm]{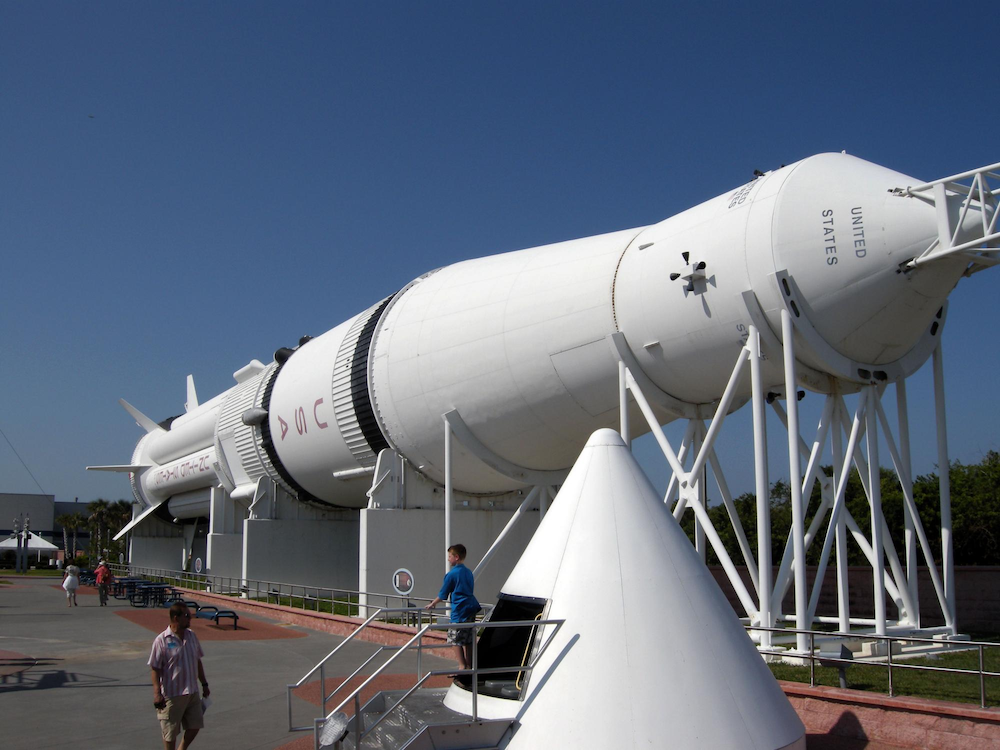} &
        \includegraphics[height=1.8cm]{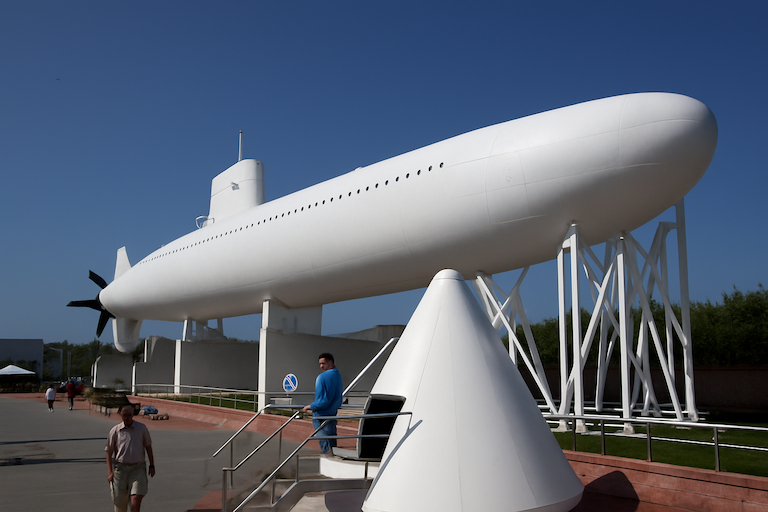} &
        \includegraphics[height=1.8cm]{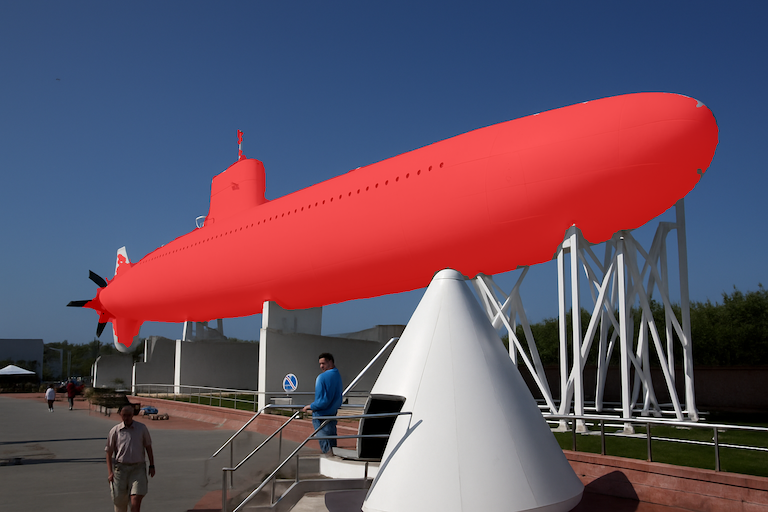} \\
    \end{tabular}}
    \vspace{-0.3cm}
    \caption{\textbf{Qualitative comparison of reasoning segmentation models across factual and counterfactual pairs.} Here, \(c=\) ``something that can fly out of the earth'' and \(c'=\) ``something that can travel beneath ocean surface'', and \(c\), \(c'\) denote the query prompts.}
    \label{fig:qualitative}
\end{figure}

%% file: assets/tables/fprefcoco.tex
\begin{table}[t!]
\centering
\caption{\textbf{Performance comparison on augmented FP-RefCOCO(+/g) dataset~\cite{wu2024see} for hallucination robustness.} We report localization accuracy ("See" task) and cIoU segmentation performance ("Segment" task); higher values indicate stronger grounding fidelity. Best performance highlighted with \colorbox{LightBlue}{\phantom{\rule{1ex}{1ex}}}, second best performances underscored with \_}
\vspace{-0.4cm}
\resizebox{0.99\columnwidth}{!}{
\begin{tabular}{lcccccc}
\hline
Method      & \multicolumn{2}{c}{\textbf{FP-RefCOCO}} & \multicolumn{2}{c}{\textbf{FP-RefCOCO+}} & \multicolumn{2}{c}{\textbf{FP-RefCOCOg}} \\ \cmidrule(lr){2-3} \cmidrule(lr){4-5} \cmidrule(lr){6-7}  
      
            & See           & Segment         & See           & Segment          & See           & Segment         \\ \hline
LISA-7B~\cite{lai2024lisa}  & 51.36         & 44.00           & 51.32         & 39.62            & 51.25         & 39.64           \\
Seg-Zero~\cite{liu2025seg} & 51.37 & 44.47 & 51.33 & 41.07 & 51.25 & 41.22 \\
VisionReasoner~\cite{liu2025visionreasoner} & 51.41 & 39.22 & 51.34 & 34.23 & 51.25 & 35.82 \\
Cascading~\cite{wu2024see} & 75.59         & 55.18           & 75.03         & 48.64            & 76.07         & 49.98           \\
SESAME~\cite{wu2024see}    & \underline{79.84}         & \underline{57.93}          & \underline{80.00}         & \underline{50.81}            & \underline{81.78}         & \underline{53.79}           \\ 
\hline
\textbf{\modelnamegradient{} (Ours)} &  \colorbox{LightBlue}{\textbf{83.37}} & \colorbox{LightBlue}{\textbf{59.57}} & \colorbox{LightBlue}{\textbf{83.00}} & \colorbox{LightBlue}{\textbf{52.91}} & \colorbox{LightBlue}{\textbf{84.21}}  & \colorbox{LightBlue}{\textbf{54.76}} \\
\end{tabular}}
\label{tab:fprefcoco}
\vspace{-0.2cm}
\end{table}

%% file: sections/6_conclusion.tex
\section{Conclusion}
In this work, we introduce the Counterfactual Segmentation Reasoning (CSR) task to evaluate grounding fidelity under controlled visual interventions. To support this task, we curate \benchmarkname{}, a diverse paired factual–counterfactual benchmark with new metrics that directly measure pixel-level hallucination in referring and reasoning segmentation settings. In addition, we introduce \modelname{}, a segmentation VLM trained with Counterfactual Finetuning (CFT) to reinforce accurate object grounding while suppressing segmentation in the absence of visual evidence. Experiments demonstrate that \modelname{} substantially reduces hallucinations, establishing a strong foundation for robust and reliable grounded segmentation.

\section*{Acknowledgments}
This research was partially supported by the U.S. Defense Advanced Research Projects Agency (DARPA) under award numbers HR00112390062 and HR001125C0303, and the U.S. Army under contract W5170125CA160. The views and conclusions contained herein are those of the authors and should not be interpreted as necessarily representing the official policies, either expressed or implied, of DARPA, the U.S. Army, or the U.S. Government. The U.S. Government is authorized to reproduce and distribute reprints for governmental purposes notwithstanding any copyright annotation therein. This work also used the Delta GPUs provided by the National Center for Supercomputing Applications through allocations CIS240752, CIS240817, and CIS250130 from the Advanced Cyberinfrastructure Coordination Ecosystem: Services \& Support (ACCESS, \citet{boerner2023access}) program, supported by National Science Foundation grants \#2138259, \#2138286, \#2138307, \#2137603, and \#2138296.

%% file: sections/X_suppl.tex
\clearpage
\setcounter{page}{1}
\maketitlesupplementary

\section{\benchmarkname{} Details}\label{ref:datasetdetails}

\textbf{Motivation.}
\benchmarkname{} introduces a counterfactual visual reasoning framework to evaluate segmentation Vision Language Models (VLMs) under controlled object-level interventions. Each factual image is paired with a counterfactual variant in which the target object is replaced by a semantically distinct, visually similar alternative, while the rest of the scene remains unchanged. This controlled pairing of visually coherent yet semantically altered scenes represents the first dataset explicitly designed for counterfactual evaluation of pixel-grounding segmentation models, enabling systematic study of vision-driven hallucinations. By comparing model outputs across these image–query pairs, \benchmarkname{} isolates whether predictions are grounded in the visual evidence or driven by semantic priors. Unlike label-only perturbations, which often fail to challenge visual grounding, our counterfactual edits introduce minimal yet meaningful visual changes, making hallucinations, such as segmenting the replaced object, directly observable. This formulation enables fine-grained, instance-level analysis of hallucination robustness in reasoning-based segmentation VLMs and motivates a need for a benchmark that can systematically stress-test grounding fidelity.

Importantly, \benchmarkname{} is not only limited to post-hoc evaluation, but it also provides a structured resource for training segmentation VLMs to better disentangle visual evidence from semantic expectations.
Evaluation on \benchmarkname{} reveals that existing state-of-the-art segmentation VLMs, including those with sophisticated reasoning capabilities, frequently hallucinate segments, underscoring their over-reliance on semantic priors in the absence of supporting visual evidence. 
This confirms their over-reliance on semantic priors rather than fine-grained visual evidence, leading to high rates of vision-driven hallucination. This failure pattern underscores the practical value of training on such data.\looseness-1 

To demonstrate this, we develop \modelname{}, a segmentation VLM training using Counterfactual Finetuning (CFT) objective, establishing a strong baseline for our proposed Counterfactual Segmentation Reasoning (CSR) task on \benchmarkname{}.
The CFT method leverages the paired training data from \benchmarkname{} to explicitly teach the model to ground accurately in the presence of visual evidence while abstaining from segmentation when evidence is absent. By learning when not to segment, \modelname{} directly addresses vision-driven hallucinations, establishing a robust foundation for reliable and visually grounded segmentation.

\noindent \textbf{Image and Mask Generation.}
\input{assets/figures/data_pipeline}
\Cref{fig:data_generation} illustrates the data generation pipeline used in \benchmarkname{}. Each counterfactual image \(\mathbf{I}'\) is derived from a factual image \(\mathbf{I}\) by applying a localized visual intervention: a single object instance of class \(c\) is replaced with an instance of a different class \(c'\), while keeping the rest of the scene unchanged.

For the referring task, each counterfactual image \(\mathbf{I}'\) is derived from a factual image \(\mathbf{I}\) by applying a targeted, localized visual intervention: an object of class \(c\) is replaced with a semantically distinct and visually plausible object of class \(c'\), while the rest of the scene remains unchanged. To construct a corresponding referring expression for \(\mathbf{I}'\), we follow the instruction format in \Cref{fig:modified_referring_prompt}, which ensures that the new reference to class \(c'\) aligns with the image context while preserving the structure of the original referring expression. The replacement is carried out using the editing constraints in \Cref{fig:mask_edit_prompt}, which enforce spatial coherence by limiting edits to a specified mask region. This design yields visually faithful and semantically meaningful counterfactuals, enabling precise evaluation of visual grounding robustness.

The reasoning task follows an analogous construction process. Starting from a factual image \(\mathbf{I}\) with a reasoning question that references an object of class \(c\), we identify the target object using the prompt in \Cref{fig:reason_label_prompt}. We then rewrite the question to refer to a new object of class \(c'\), introduced as a visually and semantically coherent substitute, guided by the prompt in \Cref{fig:reason_edit_prompt}. This procedure mirrors the referring expression pipeline while focusing on reasoning-based expressions, enabling targeted assessment of hallucination under counterfactual reasoning conditions. 
To create the corresponding counterfactual image \(\mathbf{I}\), we perform edits using the GPT-4o image generation model, which enables fine-grained object-level transformations while effectively preserving the overall scene layout and visual realism throughout the image.\looseness-1

To enable evaluation of grounding models, we provide segmentation masks for the relevant object instances in both the factual and counterfactual images. For each factual image \(\mathbf{I}\), we retain the original mask \(\mathbf{M}_c\) provided by RefCOCO~\cite{yu2016modeling}. For the counterfactual image \(\mathbf{I}'\), we generate the corresponding mask \(\mathbf{M}_{c'}'\) using Grounded SAM~\cite{ren2024grounded}. The same setting is applied to ReasonSeg~\cite{lai2024lisa} for reasoning task, except one more question query is generated for the counterfactual image \(\mathbf{I}'\) and the original question for \(\mathbf{I}\) is obtained from ReasonSeg, the masked objects of \(c\) and \(c'\) are labelled. To ensure high-quality edits and mask alignment, all counterfactual examples are manually reviewed and filtered to discard samples that introduce visual artifacts or exhibit incorrect grounding in the predicted masks. The counterfactual image–query pairs and corresponding object masks described above form the basis for evaluating model robustness to hallucination under targeted visual interventions. In the following section, we provide additional discussion on the properties of our proposed metrics, highlighting their interpretability, range, and how they support fine-grained analysis.\looseness-1

\section{Discussion of Evaluation Metrics}\label{appendix:eval}

\noindent \textbf{Range and Interpretation.}
We design our metrics to support fine-grained, instance-level analysis of hallucination behavior under controlled counterfactual interventions. The IoU-based delta metrics (\(\Delta \text{IoU}_{\text{textual}}\) and \(\Delta \text{IoU}_{\text{visual}}\)) are bounded within \([-1, 1]\), though values in practical scenarios are typically within \([0, 1]\). Higher values indicate reduced hallucination, as they correspond to greater divergence between factual predictions and those under counterfactual conditions, implying that the model appropriately suppresses predictions in the absence of supporting visual evidence.

In contrast, the Confusion Mask Score (CMS) is non-negative and unbounded above, but is normalized by the size of the corresponding ground truth object. This normalization ensures comparability across examples with varying object sizes. While the absolute value may vary based on image content and object area, CMS remains effective in capturing hallucination severity through the weighted penalty of overlapping and non-overlapping errors. In our evaluations, we set \(\alpha\!=\!3\) to emphasize overlapping errors more heavily, ensuring \(\alpha > 1\) for sharper contrast in failure cases.

\input{assets/figures/metric_distribution} 
\input{assets/tables/confidience_interval}
\input{assets/figures/qualitative_app_1}

\noindent
\textbf{Metric Distributions and Summary Statistics.}
\Cref{fig:metric_dists} illustrates the empirical distribution of \(\diou\) across all examples in \modelname{} and all baselines. The distribution of \(\dioutext\) and \(\diouvis\) reveals a bimodal pattern: one peak near \(1.0\) corresponding to successful suppression of hallucination, and a larger peak concentrated near \(0\), indicating a high frequency of hallucination cases. Notably, visual \(\diou\) scores exhibit higher density below zero compared to textual \(\diou\), whereas the inverse holds for higher values, corroborating our earlier observation that vision-driven hallucinations are more persistent across models.

\Cref{tab:metric_summary} summarizes the behavior of proposed metrics using 95\% confidence intervals. The reported means align with earlier findings that hallucination is more severe under visual counterfactual settings, with lower \(\diouvis\) and higher \(\cmscf\) compared to their textual and factual counterparts. The relatively narrow confidence intervals suggest statistical reliability and low variance across the dataset, supporting their robustness for evaluating hallucination sensitivity across diverse image-query pairs in \benchmarkname.

\section{\modelname{} Details}
\noindent\textbf{Implementation details.}
\label{implement}
Our model architecture (shown in ~\Cref{fig:model_cft} main paper) follows the modular segmentation-VLM design of prior works~\citep{lai2024lisa,wu2024see}. The architecture consists of a vision encoder, a language model and a lightweight fusion module followed by a segmentation head. Specifically, we finetune the language model component using Low-Rank Adaptation (LoRA)~\citep{hu2022lora} with rank \(r=8\), scaling factor \(\alpha=16\), and a dropout of \(0.05\), applied to the query and value of projection matrices. We additionally fine-tune the pixel decoder~\cite{kirillov2023segment} to better adapt to mask supervision.

Training uses a mixture of datasets spanning both conventional and hallucination-aware grounding tasks, namely, RefCOCO~\cite{yu2016modeling}, LLaVA VQA~\citep{liu2024visual}, FP-RefCOCO~\citep{wu2024see}, ReasonSeg~\citep{lai2024lisa}, and our proposed \benchmarkname{}, sampled in the ratio \(12\!:\!2\!:\!3\!:\!12\!:\!2\). For FP-RefCOCO, we use a \(2\!:\!1\) ratio of correct to hallucinated (negative) samples following SESAME~\citep{wu2024see}. Within \benchmarkname{}, referring and reasoning examples are used in a \(1\!:\!1\) ratio with each sample yielding four image-text pairs based on the dataset's structural design, enabling Counterfactual Finetuning.

We optimize \modelname{} using AdamW optimizer with learning rate of \(1\!\times\!10^{-4}\) and an effective batch size of \(96\), training for under \(6\) hours on \(8\) NVIDIA L40S 48GB GPUs. For fairness, all validation/test images from every dataset are excluded from finetuning \modelname{}.

\noindent \textbf{Radar Chart Details.} 
\Cref{fig:teaser} (c) in the main paper demonstrates the metrics of selected models on our benchmark \benchmarkname{} and FP-RefCOCO. For visualization purposes, CMS values are inverted to ensure that higher values correspond to better performance in the figure.

\section{Qualitative Examples}\label{appendix:qualexamples}

\input{assets/figures/qualitative_app_2}
\input{assets/tables/size_ablation}

\Cref{fig:qualitative_app_1} and \Cref{fig:qualitative_app_2} present qualitative comparisons of baseline segmentation VLMs along with \modelname{} on paired factual and counterfactual examples for both referring and reasoning settings. Each setting shows the four key configurations in our benchmark: the factual pair \((\mathbf{I}, c)\), label-replacement pair \((\mathbf{I}, c')\), counterfactual intervention pair \((\mathbf{I}', c)\), and the true counterfactual pair \((\mathbf{I}', c')\).

\Cref{fig:qualitative_app_1}, demonstrate how both label-replacement, \((\mathbf{I}, c')\) and counterfactual intervention \((\mathbf{I}', c)\), expose hallucinations across various models. In the referring example on the left (``front zebra'' vs ``front cow''), several baselines continue to segment the cow even when the query refers to ``front zebra'' revealing strong vision-driven hallucinations that our factual-counterfactual construction is designed to elicit. In the reasoning example on the right (``storing wine'' vs ``resting feet''), many models latch onto salient but incorrect regions, such as segmenting the footstool when only the barrel satisfies the given query. In contrast, \modelname{} more consistently suppresses masks for visually absent concepts and segments only when the queried region is truly present.

\Cref{fig:qualitative_app_2} illustrates queries that involve attribute-level referring and reasoning. For reasoning prompts that ask for reflective versus ornamental regions of the image, most models respond with masks on the ornamental region rather than abstaining due to the absence of a reflective region in the image. \modelname{}, however, abstains in both cases to segment the absent object and presents more localized masks that consistently track the intended object or region across all cases, as opposed to SESAME, which learns to over-mitigate the masks even in the presence of visual evidence.

Together, these qualitative examples demonstrate the failure modes that counterfactual segmentation reasoning is designed to capture with the help of \benchmarkname{} and our proposed \(\cms\) and \(\diou\) metrics. such as persistent masks for absent or replaced concepts, over-mitigation of hallucination, and sensitivity to counterfactual interventions. They also highlight how \benchmarkname{}, together with our metrics, provides consistent and interpretable signals that capture the nature and severity of hallucinations under controlled image–query manipulations. These qualitative trends corroborate our quantitative results on \benchmarkname{}, where \modelname{} shows reduced hallucination and more stable behavior across both referring and reasoning segmentation tasks compared to prior models.

\input{assets/tables/size_ablation2}

\section{Ablations}

\noindent \textbf{\benchmarkname{} Ablations} To assess how model performance varies across different spatial granularity, we group objects into small, medium, and large mask size categories and evaluate all metrics at each scale. \Cref{tab:size_ablation_ref} on the referring task and \Cref{tab:size_ablation_reason} on the reasoning task show that all models consistently exhibit the largest performance degradation both in terms of \(\Delta \text{IoU}_{\text{visual}}\) and \(\text{CMS}_{\text{counterfact}}\) for small objects, making hallucinations for small regions particularly challenging. For instance, PixelLM-13B is already the most stable baseline model regarding the size change. It still shows stronger \(\Delta \text{IoU}_{\text{visual}}\) for large objects, but better for larger ones, while its hallucination scores (CMS) worsen accordingly. In contrast, the trend for \(\text{CMS}_{\text{fact}}\) is less size-sensitive, likely due to its normalization by ground-truth mask area, which can dampen the relative penalty of hallucinations for smaller objects. CoT methods, such as Seg-Zero and VisionReasoner, demonstrate more robust performance with their thinking chain in \(\Delta \text{IoU}_{\text{visual}}\), especially for larger objects. However, this improved performance does not guarantee a better CMS performance, indicating the tradeoff that their thinking process may introduce hallucination. It is also obvious that a smaller object size may result in worse segmentation performance, even for the medium-sized, such as the \(\text{CMS}_{\text{counterfact}}\) of VisionReasoner in the reasoning task.

\input{assets/figures/loss_abl_iou}

\modelname{} demonstrates the strongest or the second-best hallucination suppression behavior across almost all object sizes, achieving low factual and counterfactual hallucination scores throughout different sizes, especially in the large size. SESAME also performs well in CMS by suppressing hallucination. However, this suppression comes at the cost of segmentation performance, as reflected by lower \(\Delta \text{IoU}\) scores across object sizes. In contrast, our model still maintains relatively high \(\Delta \text{IoU}_{\text{textual}}\) and \(\Delta \text{IoU}_{\text{visual}}\), indicating a more balanced tradeoff between hallucination avoidance and segmentation fidelity. Notably, LISA-7B/13B, GlaMM-7B, and SESAME-7B models yield  \(\Delta \text{IoU}_{\text{visual}}\) scores below \(\Delta \text{IoU}_{\text{textual}}\) across most mask sizes, indicating greater susceptibility to vision-driven hallucinations, underscoring the unique diagnostic value of counterfactual visual reasoning in \benchmarkname{}. While our model demonstrates that CFT enhances hallucination suppression capability in both cases with counterfactual reasoning, we still require a more targeted and effective method for mitigating visual hallucinations, especially when the mask size is small, which requires more attention to details from the model.

\noindent \textbf{\modelname{} Ablations} 
We also conduct an ablation study over \(\lambda_{neg}\), which scales the negative branch of our counterfactual finetuning loss. This term explicitly penalizes hallucinations by discouraging segmentation or textual output in the absence of visual evidence. \Cref{fig:cms_lambda_ablation} reports the effect of varying \(\lambda_{neg}\) on our proposed \(\cms\) metric on \benchmarkname{}. \Cref{fig:delta_iou_lambda_ablation} shows the corresponding trends for the hallucination error metric \(\diou\), allowing us to assess hallucination mitigation performance and post-mitigation consistency on \benchmarkname{}.

\input{assets/figures/loss_abl_cms}

We ablate the negative loss weights with \(\lambda_{neg} \in \{0.8, 1.0, 1.2\}\). Overall, varying \(\lambda_{neg}\) shows that \(\lambda_{neg} = 1.0\) offers the most favorable behavior across our metrics. In the Referring setting, it yields noticeably lower \(\cms\) and improved IoU scores for both factual and counterfactual masks compared to the other choices, indicating stronger suppression of hallucinated regions while maintaining accurate segments in the presence of visual evidence. In the Reasoning setting, all three values of \(\lambda_{neg}\) behave similarly, with \(\lambda_{neg} = 1\) remaining consistently competitive across \(\cms\) and \(\diou\). Based on these trends, we set \(\lambda_{neg} = 1\) for all subsequent experiments as it provides the best balance between referring and reasoning performance. 

\section{Broader Impacts}

\benchmarkname{} provides the first Counterfactual Segmentation Reasoning (\(\csr\)) evaluation framework and benchmark for pixel-level counterfactual visual reasoning, enabling fine-grained diagnosis of hallucination behaviors in vision-language segmentation models. By explicitly probing model behavior under controlled visual interventions, our benchmark facilitates greater transparency and diagnostic insight into segmentation failures and their underlying causes.
Understanding when and why models segment objects that are no longer visually present is essential for developing trustworthy systems. 
Beyond application-specific relevance, our work contributes to the broader vision-language community by emphasizing the need to evaluate grounding robustness under structured counterfactuals, not merely aggregate accuracy. We believe this direction is crucial for advancing safe, reliable, and generalizable multimodal AI systems that behave consistently under visual perturbations.

Along with the evaluation framework, \benchmarkname{} also provides a training resource for robust model development. The paired factual-counterfactual structure supports explicit counterfactual supervision, enriching model training with fine-grained grounding perturbations, and making it well-suited for contrastive or preference learning. Our proposed model, \modelname{}, serves as a strong baseline trained on this resource via a Counterfactual Finetuning strategy to reinforce robust grounding behavior, showing that grounding robustness can be significantly improved by supervising models with factual-counterfactual data. 

However, as with any diagnostic benchmark, there is a possibility of misuse. For example, while these tools are designed for research, the ability to manipulate object presence in otherwise faithful scenes may be misused to create misleading or deceptive visual content. We therefore emphasize that our editing pipeline is restricted to non-human, non-sensitive categories and should not be applied to real-world identity or surveillance data. Additionally, systems optimized for abstention-based safety may under-detect rare or safety-critical objects if deployed without calibration. We recommend responsible, domain-appropriate deployment practices and transparent reporting when using counterfactual data for model evaluation or training.


\lstdefinestyle{promptstyle}{
  backgroundcolor=\color{gray!5},
  basicstyle=\ttfamily\scriptsize,
  frame=single,
  breaklines=true,
  showstringspaces=false,
  literate={•}{{\textbullet\ }}1
}

\begin{figure*}[t]
    \centering
    \begin{minipage}{\linewidth}
        \begin{lstlisting}[style=promptstyle]
You are given two images:
    1. The full scene.
    2. A binary mask marking an object labeled "{label}". The referring description is "{description}". In case of vague or incorrect descriptions, follow the image and mask.

Task:
- Locate the masked object precisely.
- Create a replacement instruction that:
    • Uniquely identifies the object (position, color, size, shape, etc.)
    • Swaps it for a new object that is not already present
    • Ensures the new object is meaningful, similar in size and shape, but different in identity

CRITICAL:
Your instruction must leverage the label/description to precisely identify the masked object.

Requirements:
- The new object must not exist anywhere in the image.
- The new object must be a common object, not an abstract concept.
- Avoid vague objects like "fruit" or "vegetable"; be specific.
- Avoid changes that are too similar or unnoticeable, such as changing glass to pokal. The new object cannot be a name or description still correct for the original object.
- The replacement must be reasonable for the original object's location (e.g., animal to another animal in a zoo, but not a car).
- If the original description is nonsense (e.g., "yep"), use the mask and image to determine the object, do not hallucinate unnecessary details.
- The proposed new object MUST NOT satisfy the original description "{description}".
- If one tried to use "{description}" to refer to the new object, it MUST be incorrect and not make sense.
- Do NOT use quantity words like "a", "an", "the".

Additional Cautions:
1. Ensure the proposed replacement does NOT occur elsewhere in the image, even partially.
2. Match the approximate size, shape, and spatial position of the masked region to maintain realism.
3. The instruction must correspond exactly to the masked region, not a nearby similar object.

Output:
Only one line in this exact format:
Change <original object referring description> to <new object referring description>.
        \end{lstlisting}
    \end{minipage}
    \vspace{-0.5cm}
    \caption{\textbf{Prompt for Generating Modified Referring Expressions.} 
    The prompt instructs the VLM to identify the masked object 
    and produce a contextually plausible replacement while enforcing strict 
    visual, spatial, and semantic constraints. Here, \texttt{\{label\}} and 
    \texttt{\{description\}} corresponds to the RefCOCO object label and 
    referring expression.}
    \label{fig:modified_referring_prompt}
\end{figure*}


\lstdefinestyle{promptstyle}{
  backgroundcolor=\color{gray!5},
  basicstyle=\ttfamily\scriptsize,
  frame=single,
  breaklines=true,
  showstringspaces=false,
  literate={•}{{\textbullet\ }}1
}

\begin{figure*}[t]
    \centering
    \begin{minipage}{\linewidth}
        \begin{lstlisting}[style=promptstyle]
You must only {item['gpt_instruction']}. Carefully analyze the image to find the object described in the instruction above. Pay attention to the location details (position, color, size, surrounding context) mentioned in the instruction.

You are provided with an inverse mask, where the masked regions represent parts of the image that must be strictly preserved.
You are only allowed to modify the unmasked (transparent) regions. No edits are allowed in any masked area.
  
Even if there are multiple similar objects in the image, you must only change the one located in the unmasked area, do not modify any other similar objects outside of the unmasked region.

Strictly maintain the size, position, and shape of the unmasked region: do not resize, move, or distort it. Do not zoom in or out, and do not change the aspect ratio.
All other parts of the image (including other similar objects, background, lighting, textures, and context) must remain completely unchanged.

Additional Cautions:
    1. Even if the masked object looks very similar to the target object, you must still perform the edit. Do not skip the modification or leave the object unchanged just because the two look alike. The replacement must be clearly visible and consistent with the given instruction. 
    2. Carefully verify the mask position before editing. Perform editing only on the specific object within the masked region and indicated by the referring prompt, not on any other objects even if they are identical or of the same type. The modification must occur exactly inside the masked region, not elsewhere.  
        
The final edited image must look realistic, natural, and indistinguishable from an untouched real-world photograph.
        
        \end{lstlisting}
    \end{minipage}
    \vspace{-0.4cm}
    \caption{\textbf{Prompt for Mask-Constrained Image Editing.} 
    This prompt instructs the VLM to perform localized image edits strictly within 
    the unmasked region while preserving all masked content, ensuring spatial 
    alignment and visual realism.}
    \label{fig:mask_edit_prompt}
\end{figure*}

\lstdefinestyle{promptstyle}{
  backgroundcolor=\color{gray!5},
  basicstyle=\ttfamily\scriptsize,
  frame=single,
  breaklines=true,
  showstringspaces=false,
  literate={•}{{\textbullet\ }}1
}

\begin{figure*}[t]
    \centering
    \begin{minipage}{\linewidth}
        \begin{lstlisting}[style=promptstyle]
You are given two images:
- The FIRST image is the original scene.
- The SECOND image is a WHITE binary mask highlighting the specific object region in the same scene.

Step 1:
Localize the WHITE mask area within the original image to identify the spatial region of interest.

Step 2:
Compare the localized region in the original image with the reasoning question:"{question}".

Step 3:
Determine what object is being referred to, focusing ONLY on the masked region.

CRITICAL:
You must ONLY analyze the WHITE masked region and its corresponding area in the original image. Do NOT infer or assume anything about unmasked regions or other objects outside the mask.

Task:
Identify EXACTLY which object is being referenced in the question and highlighted by the WHITE mask. Provide a concise label with key distinguishing features.

Requirements:
- Generate a simple label that includes:
    • The object name (e.g., "car", "person", "tree", "building", "ball", "bottle")
    • Relative position (upper-left, center, lower-right, etc.)
    • One key visual feature (color, size, or distinctive characteristic)
- Do NOT include quantity words like "a", "an", "the".
- Focus ONLY on the masked object; ignore all other content.

Output Format:
A single concise label combining the object name, location, and one key feature.

Examples:
"red cup in upper-left corner"
"larger cup on the right"
"blue bottle in center"
"small ball at bottom"
        \end{lstlisting}
    \end{minipage}
    \vspace{-0.5cm}
    \caption{\textbf{Prompt for Extracting the Reasoning Target Label.} 
    This prompt guides a VLM to identify the exact object referenced by a 
    reasoning question using a binary mask, ensuring localization-specific 
    and feature-aware label extraction.}
    \label{fig:reason_label_prompt}
\end{figure*}

\lstdefinestyle{promptstyle}{
  backgroundcolor=\color{gray!5},
  basicstyle=\ttfamily\scriptsize,
  frame=single,
  breaklines=true,
  showstringspaces=false,
  literate={•}{{\textbullet\ }}1
}

\begin{figure*}[t]
    \centering
    \begin{minipage}{\linewidth}
        \begin{lstlisting}[style=promptstyle]
You must only perform the specified editing operation described in the instruction above. Carefully analyze the image to find the object referenced, paying close attention to its position, color, size, and surrounding context.

You are provided with an inverse mask, where the masked regions represent areas that must be strictly preserved. You may only modify the unmasked (transparent) regions. No edits are allowed in any masked area.

Even if the image contains multiple similar objects, you must only modify the one located within the unmasked region, do not alter any similar objects outside the editable area.

Strictly maintain the size, position, and shape of the unmasked region. Do not resize, move, or distort it. Do not zoom or change the aspect ratio. All other parts of the image (including background, lighting, textures, and similar objects) must remain completely unchanged.

Additional Cautions:
1. Even if the masked object strongly resembles the target object, you must still perform the edit. Do not skip or ignore the modification. The replacement must be visible and consistent with the instruction.
2. Carefully verify the mask position before editing. Modify only the specific object inside the masked region, not nearby identical or similar objects. The change must occur exactly within the masked area.

The final edited image must appear realistic, natural, and indistinguishable from an untouched photograph.
        \end{lstlisting}
    \end{minipage}
    \vspace{-0.5cm}
    \caption{\textbf{Prompt for Generating Modified Reasoning Expressions.} 
    This prompt enforces mask-constrained and spatially aligned image edits, ensuring that reasoning-targeted modifications affect only the intended region while preserving global scene realism.}
    \label{fig:reason_edit_prompt}
\end{figure*}

%% file: assets/figures/data_pipeline.tex
\begin{figure*}[t]
  \centering
  \includegraphics[width=0.93\textwidth]{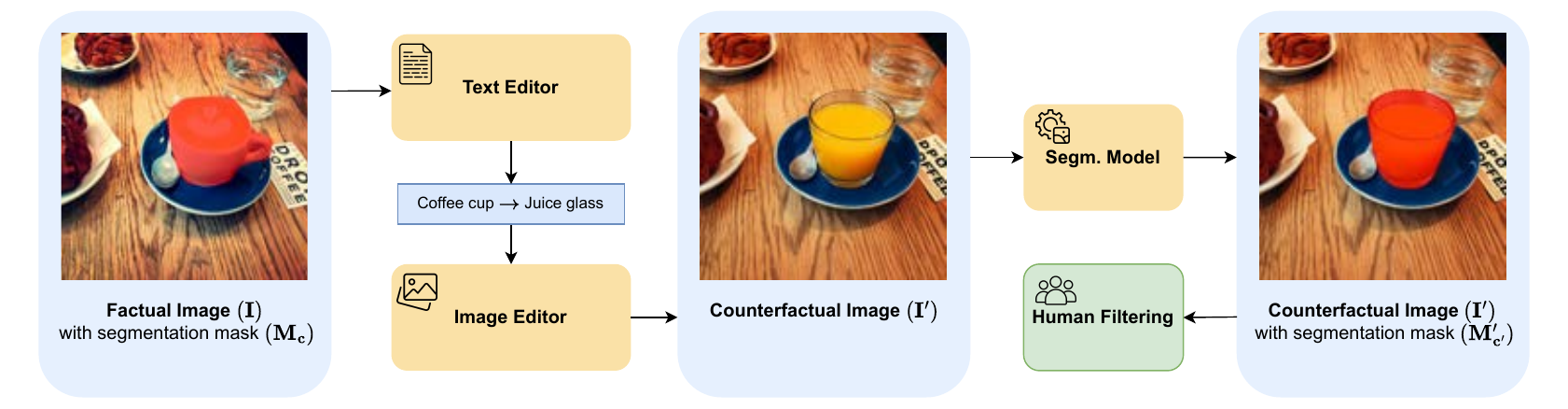}
  \vspace{-0.3cm}
      \caption{\textbf{Core Data Generation Pipeline.} Pipeline components for referral and reasoning data generation.}
  \label{fig:data_generation}
  \vspace{-0.3cm}
\end{figure*}

%% file: assets/figures/metric_distribution.tex
\begin{figure}[t!]
    \centering
    \includegraphics[width=0.99\linewidth]{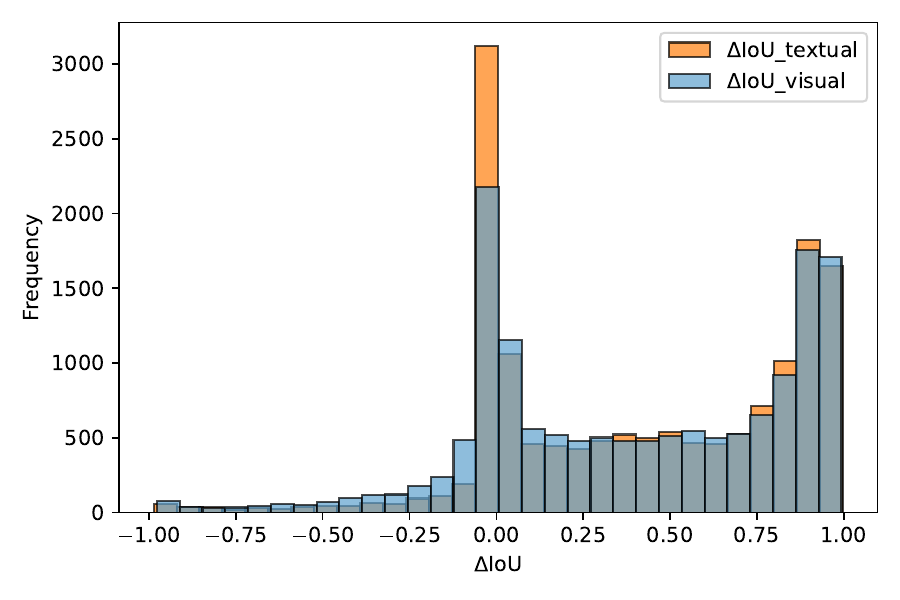}
    \vspace{-0.5cm}
    \caption{\textbf{Distribution of \(\Delta \text{IoU}\) Across All Samples.} Most \(\Delta \text{IoU}\) values lie near zero, indicating persistent hallucinations.}
    \vspace{-0.3cm}
    \label{fig:metric_dists}
\end{figure}

%% file: assets/tables/confidience_interval.tex
\begin{table}[t!]
\centering
\caption{\textbf{Mean Values and Confidence Intervals for Evaluation Metrics.} The final column reports the 95\% confidence interval half-width (\(\pm\)) as an indicator of variability.}
\label{tab:metric_summary}
\vspace{-0.3cm}
\resizebox{0.99\columnwidth}{!}{
\begin{tabular}{lcccc}
\toprule
\textbf{Metric} & \textbf{Mean} & \textbf{95\% CI Lower} & \textbf{95\% CI Upper} & \(\pm\) \textbf{CI Half-width} \\
\midrule
\(\Delta \text{IoU}_{\text{textual}}\)          & 0.4167 & 0.4100 & 0.4235 & \(\pm 0.0068\) \\
\(\Delta \text{IoU}_{\text{visual}}\)           & 0.3981 & 0.3912 & 0.4051 & \(\pm 0.0070\) \\
\(\text{CMS}_{\text{factual}}\)                & 0.4029 & 0.3949 & 0.4109 & \(\pm 0.0080\) \\
\(\text{CMS}_{\text{counterfact}}\)         & 0.6185 & 0.6075 & 0.6294 & \(\pm 0.0110\) \\
\bottomrule
\end{tabular}
}
\vspace{-0.2cm}
\end{table}

%% file: assets/figures/qualitative_app_1.tex
\begin{figure*}[t!]
    \centering
    \begin{minipage}[t]{0.49\linewidth}
        \centering
        \setlength{\tabcolsep}{1pt}
        \renewcommand{\arraystretch}{0.6}
        \resizebox{\linewidth}{!}{
        \begin{tabular}{rccccccc}
            & \textbf{GT} 
            & \textbf{LISA} 
            & \textbf{PixelLM}
            & \textbf{GLaMM}
            & \textbf{Seg-Zero}
            & \textbf{SESAME}
            & \textbf{\modelname} \\

            \raisebox{0.6cm}{\(\mathbf{I}, c\)} &
            \includegraphics[height=2.6cm]{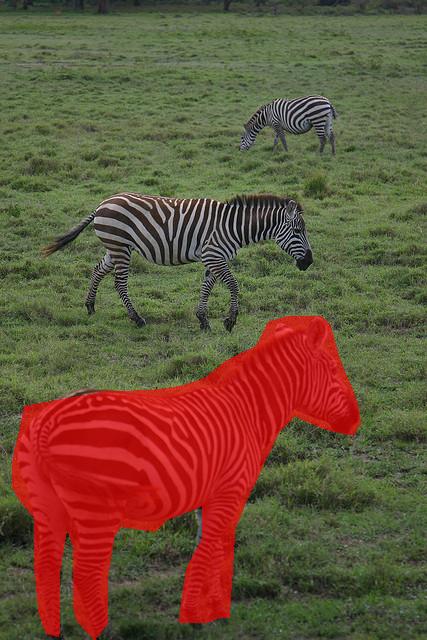} &
            \includegraphics[height=2.6cm]{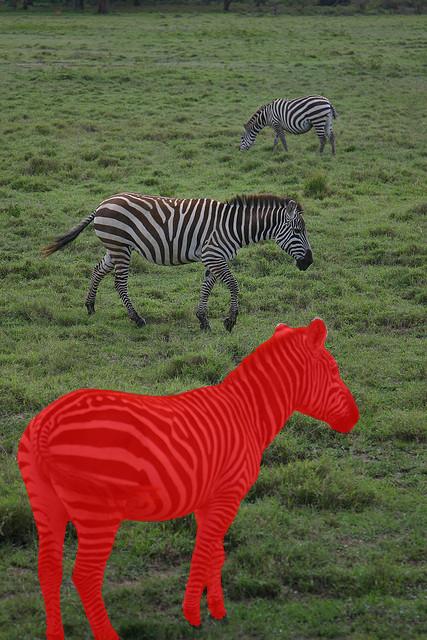} &
            \includegraphics[height=2.6cm]{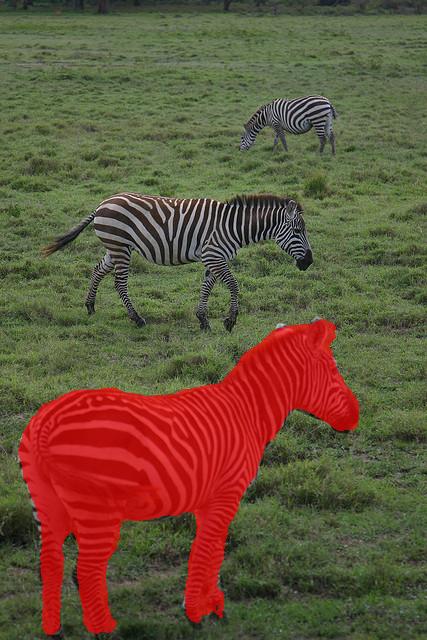} &
            \includegraphics[height=2.6cm]{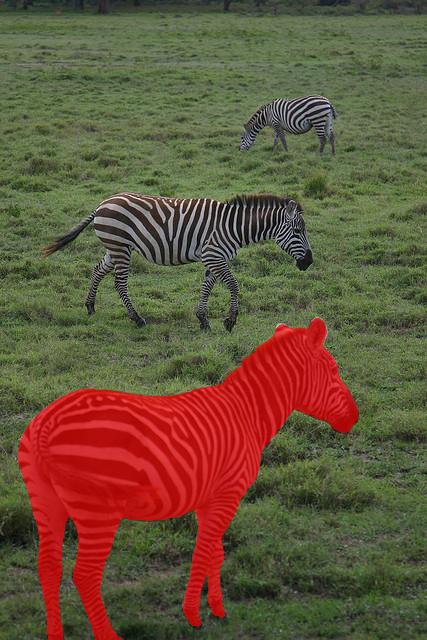} &
            \includegraphics[height=2.6cm]{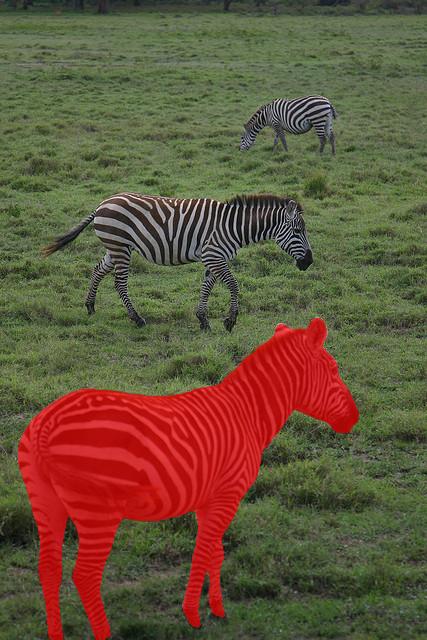} &
            \includegraphics[height=2.6cm]{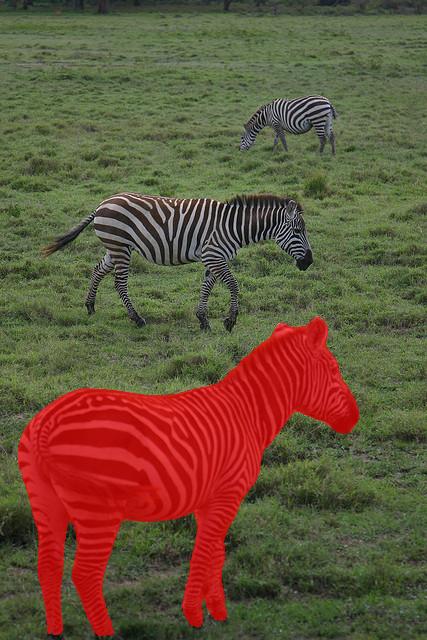} &
            \includegraphics[height=2.6cm]{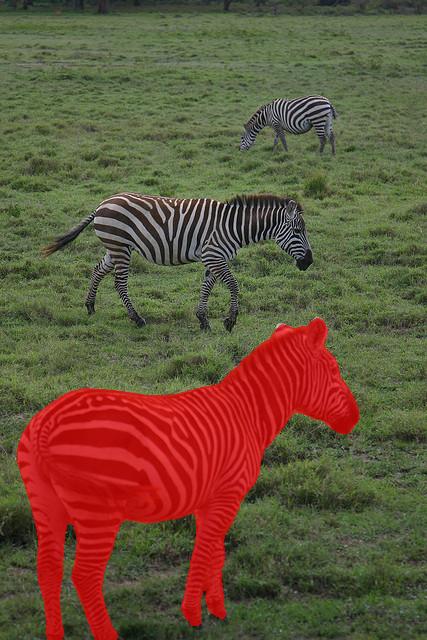} \\

            \raisebox{0.6cm}{\(\mathbf{I}, c'\)} &
            \raisebox{0.6cm}{\(\boldsymbol{\varnothing}\)} &
            \includegraphics[height=2.6cm]{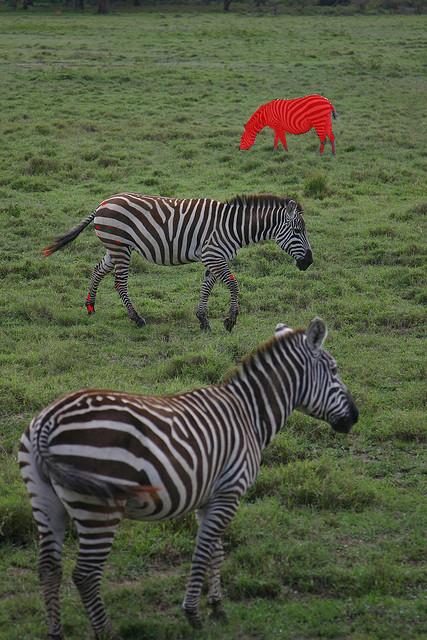} &
            \includegraphics[height=2.6cm]{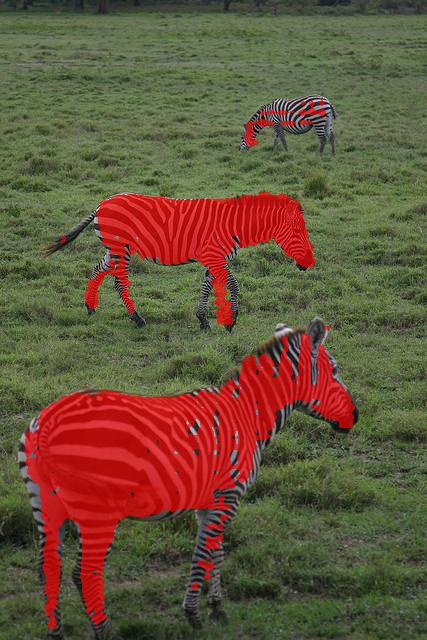} &
            \includegraphics[height=2.6cm]{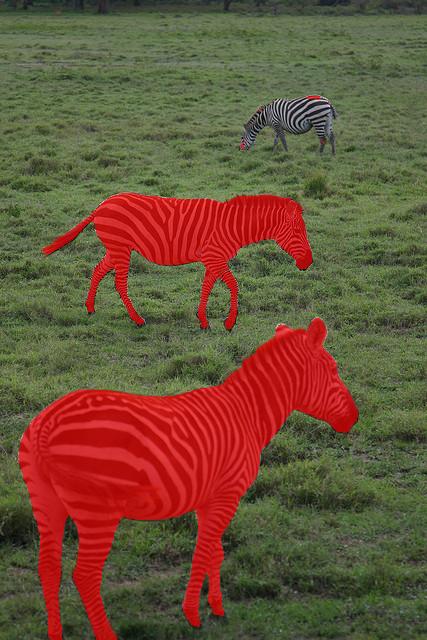} &
            \includegraphics[height=2.6cm]{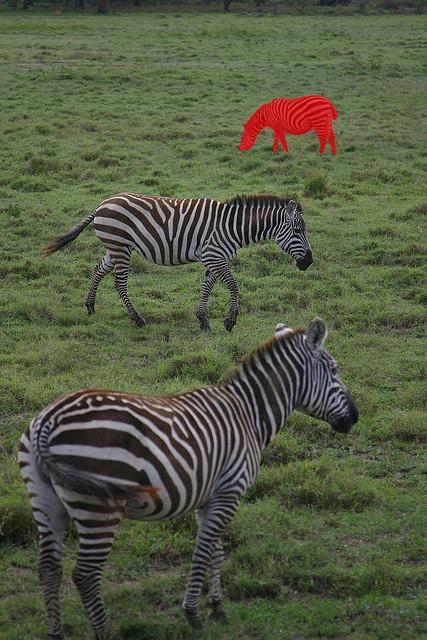} &
            \includegraphics[height=2.6cm]{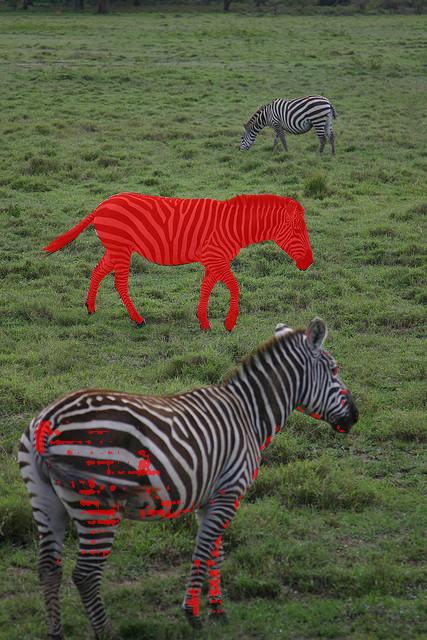} &
            \includegraphics[height=2.6cm]{assets/figures/ref_compress/doc4_ours/mask_edtl_orgi.jpg} \\

            \raisebox{0.6cm}{\(\mathbf{I}', c\)} &
            \raisebox{0.6cm}{\(\boldsymbol{\varnothing}\)} &
            \includegraphics[height=2.6cm]{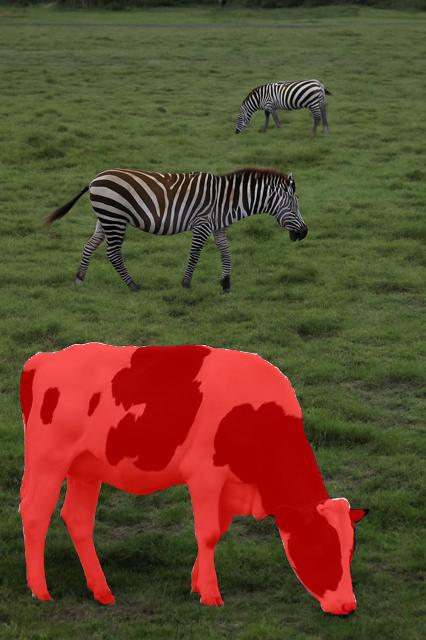} &
            \includegraphics[height=2.6cm]{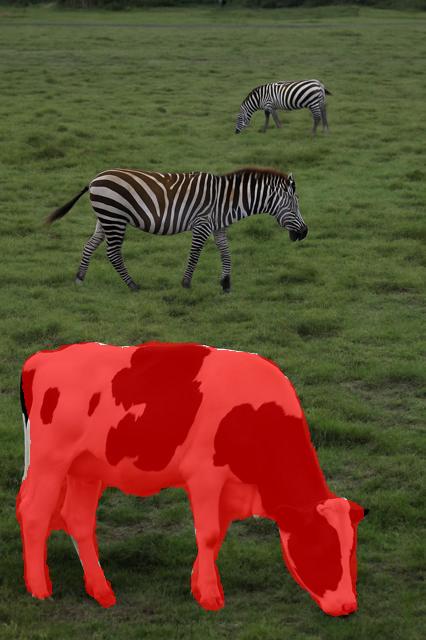} &
            \includegraphics[height=2.6cm]{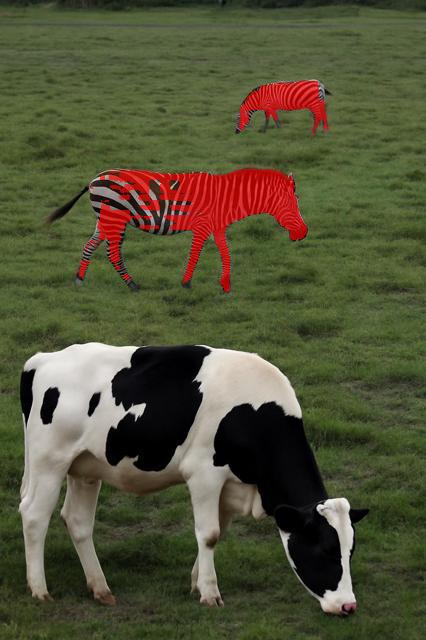} &
            \includegraphics[height=2.6cm]{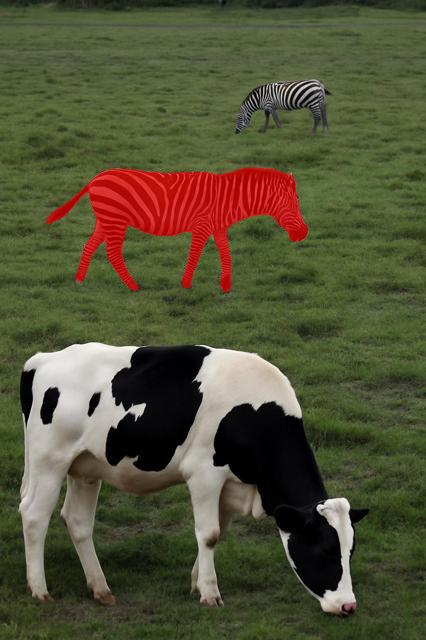} &
            \includegraphics[height=2.6cm]{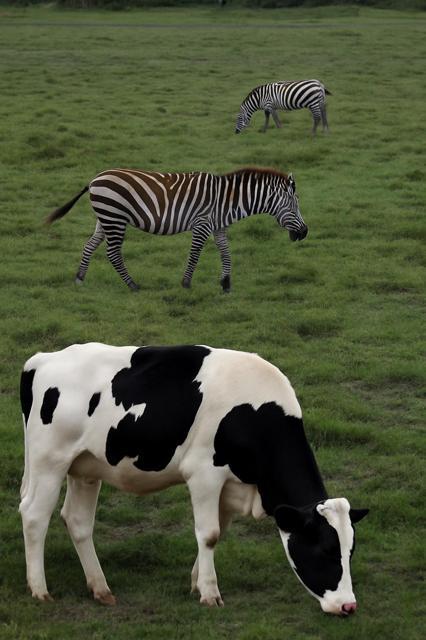} &
            \includegraphics[height=2.6cm]{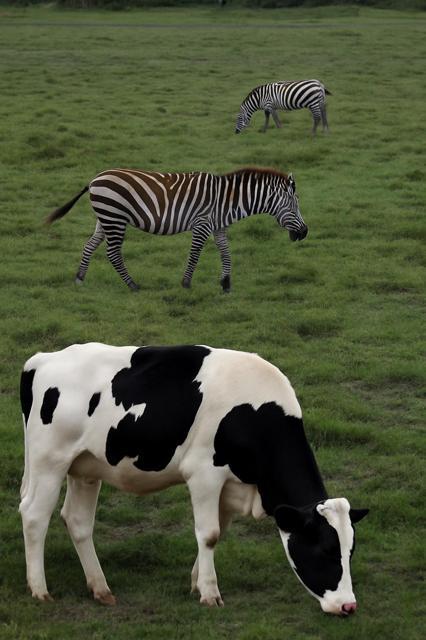} \\

            \raisebox{0.6cm}{\(\mathbf{I}', c'\)} &
            \includegraphics[height=2.6cm]{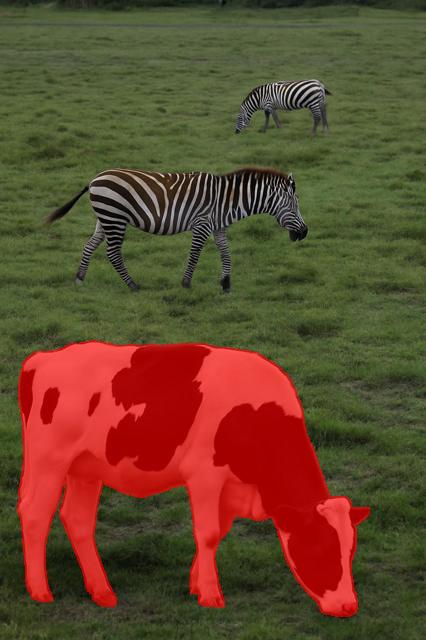} &
            \includegraphics[height=2.6cm]{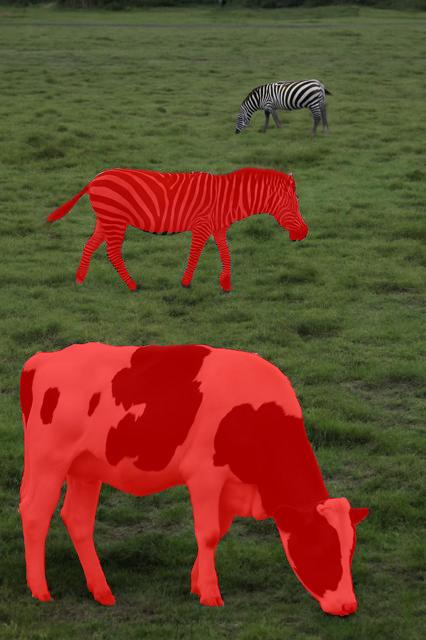} &
            \includegraphics[height=2.6cm]{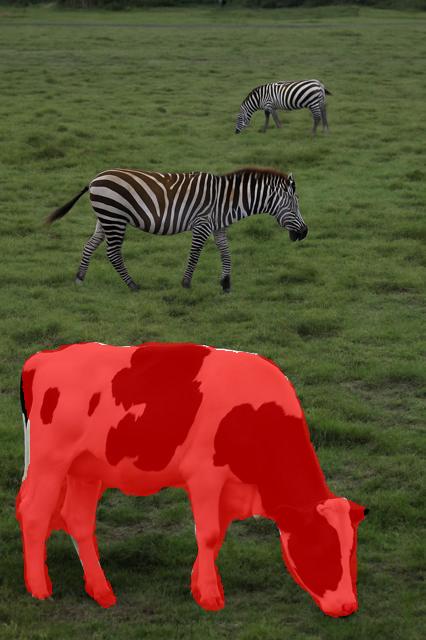} &
            \includegraphics[height=2.6cm]{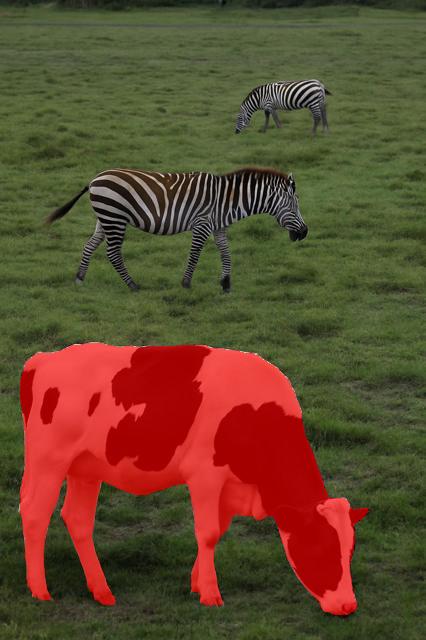} &
            \includegraphics[height=2.6cm]{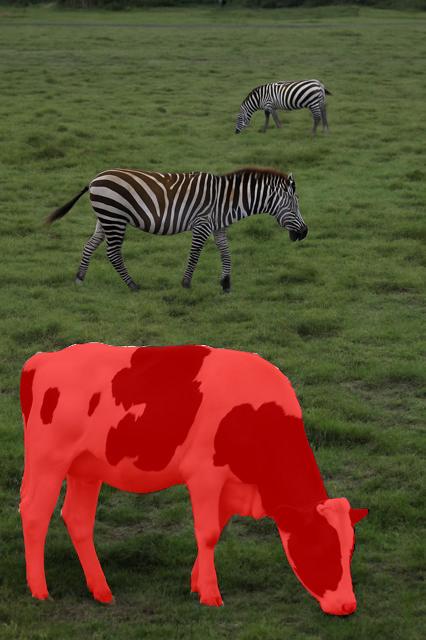} &
            \includegraphics[height=2.6cm]{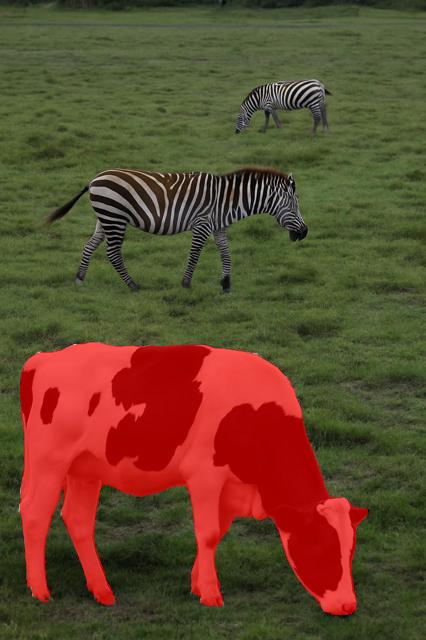} &
            \includegraphics[height=2.6cm]{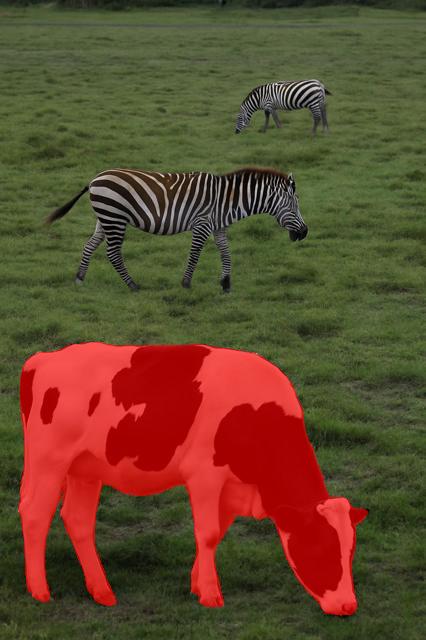} \\
        \end{tabular}}
        \caption*{\(c=\)``front zebra'' and \(c'=\)``front cow''.}
    \end{minipage}
    \hfill
    \begin{minipage}[t]{0.49\linewidth}
        \centering
        \setlength{\tabcolsep}{1pt}
        \renewcommand{\arraystretch}{0.6}
        \resizebox{\linewidth}{!}{
        \begin{tabular}{rccccccc}
            & \textbf{GT} 
            & \textbf{LISA} 
            & \textbf{PixelLM}
            & \textbf{GLaMM}
            & \textbf{Seg-Zero}
            & \textbf{SESAME}
            & \textbf{\modelname} \\

            \raisebox{0.6cm}{\(\mathbf{I}, c\)} &
            \includegraphics[height=2.6cm]{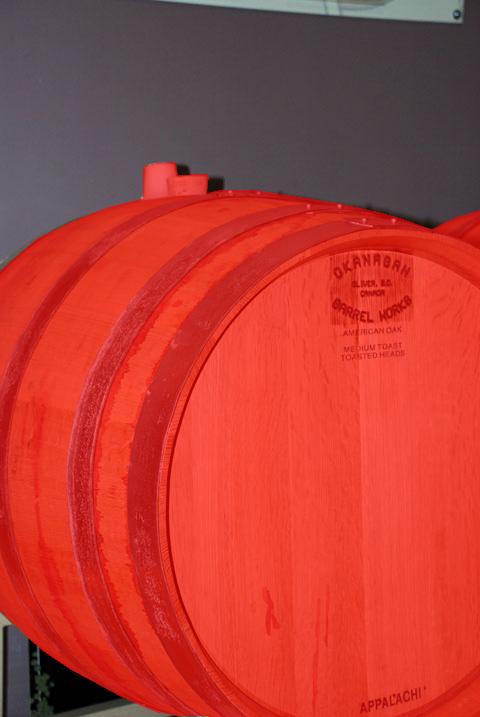} &
            \includegraphics[height=2.6cm]{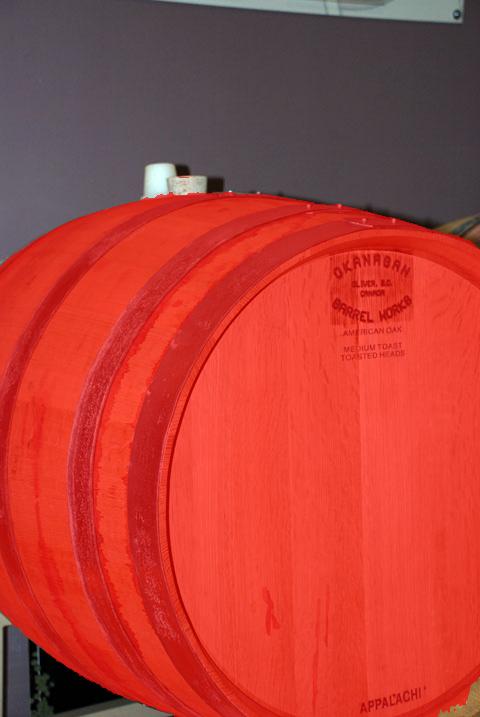} &
            \includegraphics[height=2.6cm]{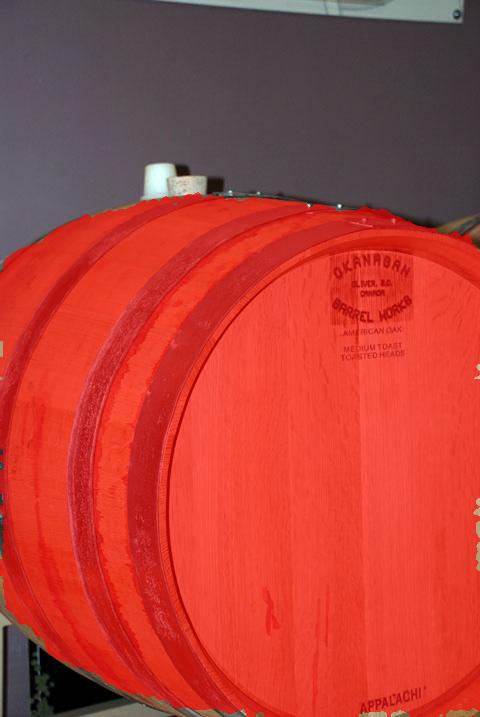} &
            \includegraphics[height=2.6cm]{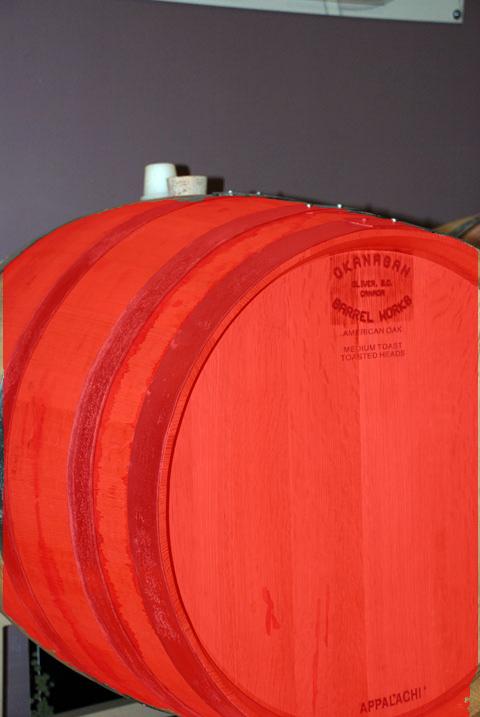} &
            \includegraphics[height=2.6cm]{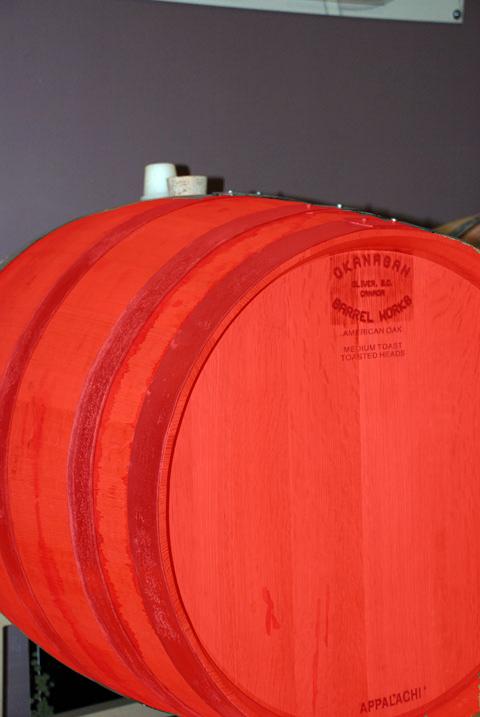} &
            \includegraphics[height=2.6cm]{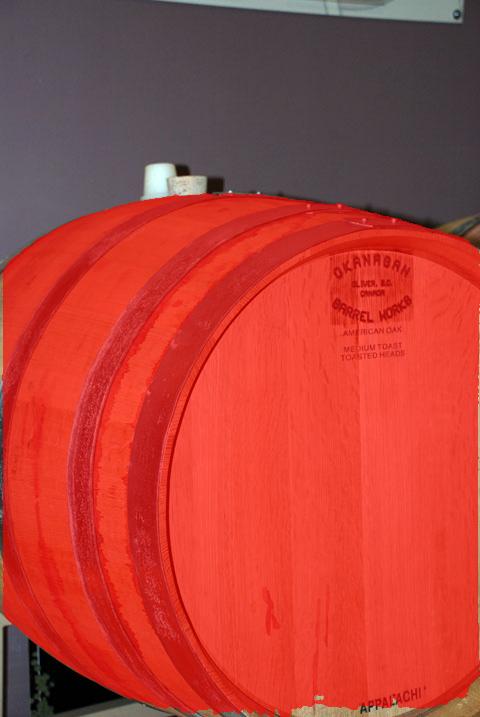} &
            \includegraphics[height=2.6cm]{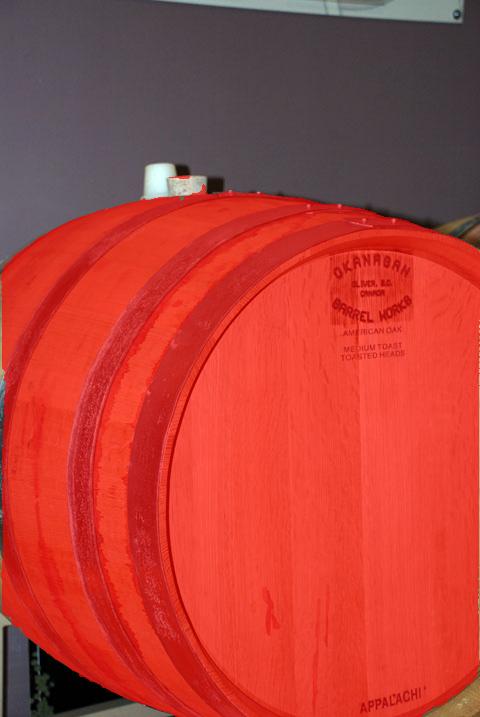} \\

            \raisebox{0.6cm}{\(\mathbf{I}, c'\)} &
            \raisebox{0.6cm}{\(\boldsymbol{\varnothing}\)} &
            \includegraphics[height=2.6cm]{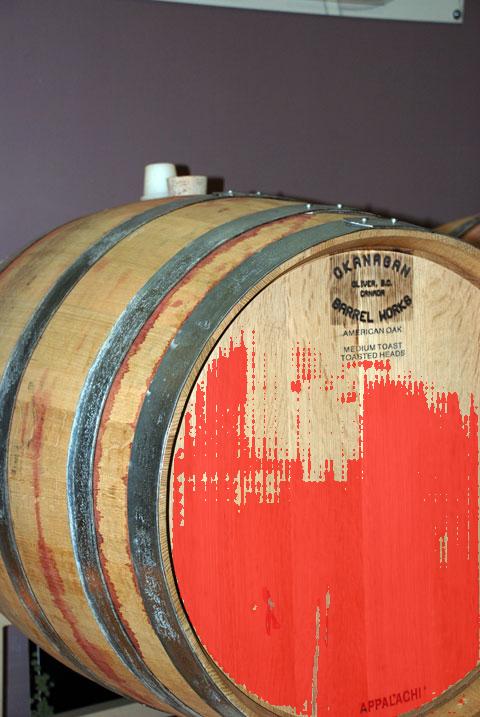} &
            \includegraphics[height=2.6cm]{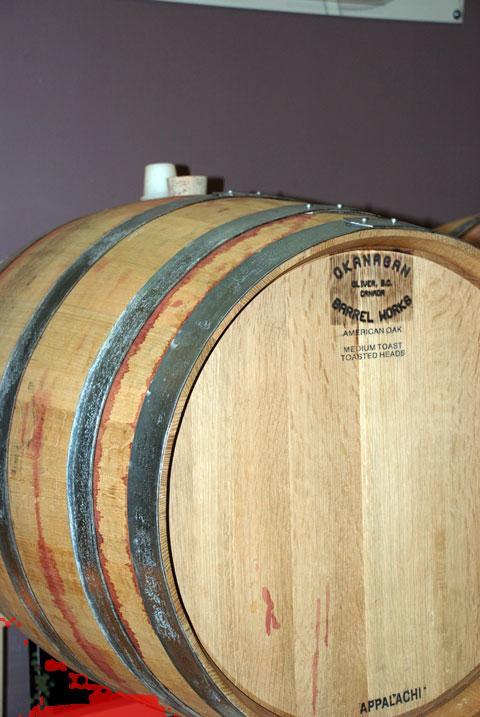} &
            \includegraphics[height=2.6cm]{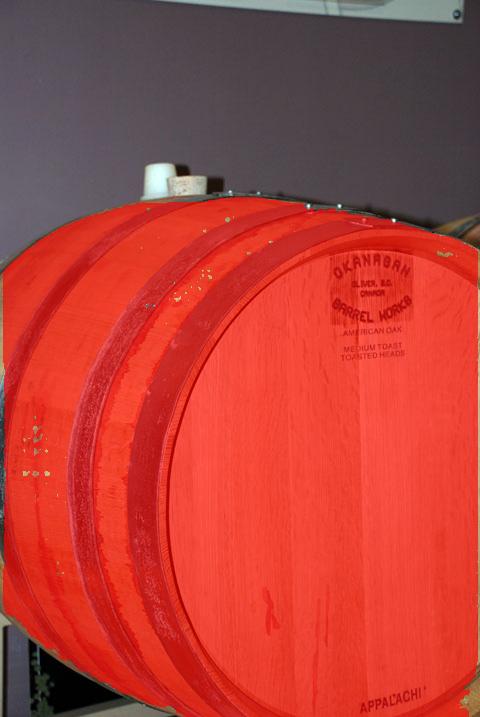} &
            \includegraphics[height=2.6cm]{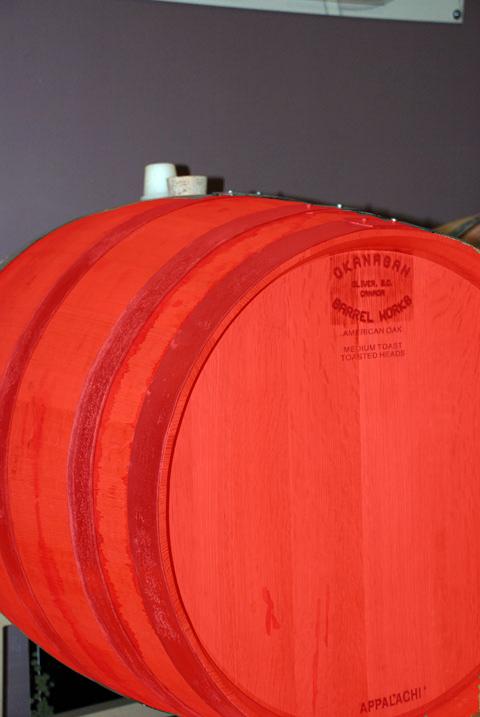} &
            \includegraphics[height=2.6cm]{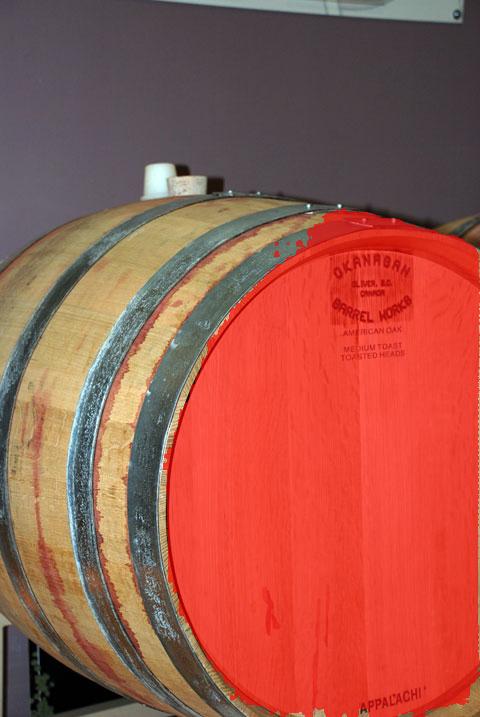} &
            \includegraphics[height=2.6cm]{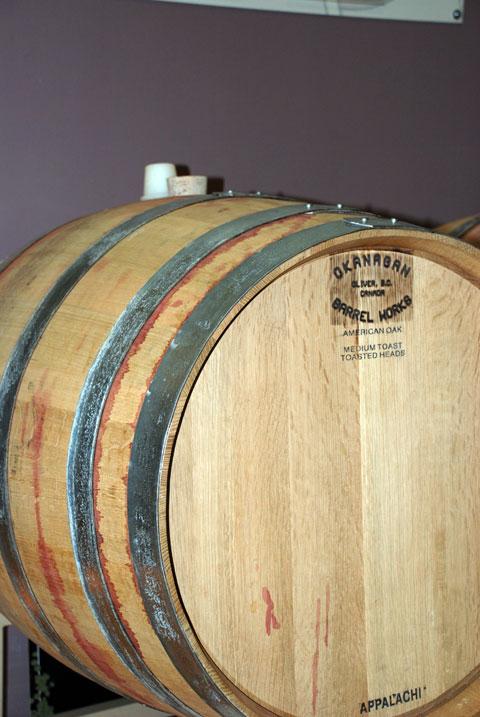} \\
        
            \raisebox{0.6cm}{\(\mathbf{I}', c\)} &
            \raisebox{0.6cm}{\(\boldsymbol{\varnothing}\)} &
            \includegraphics[height=2.6cm]{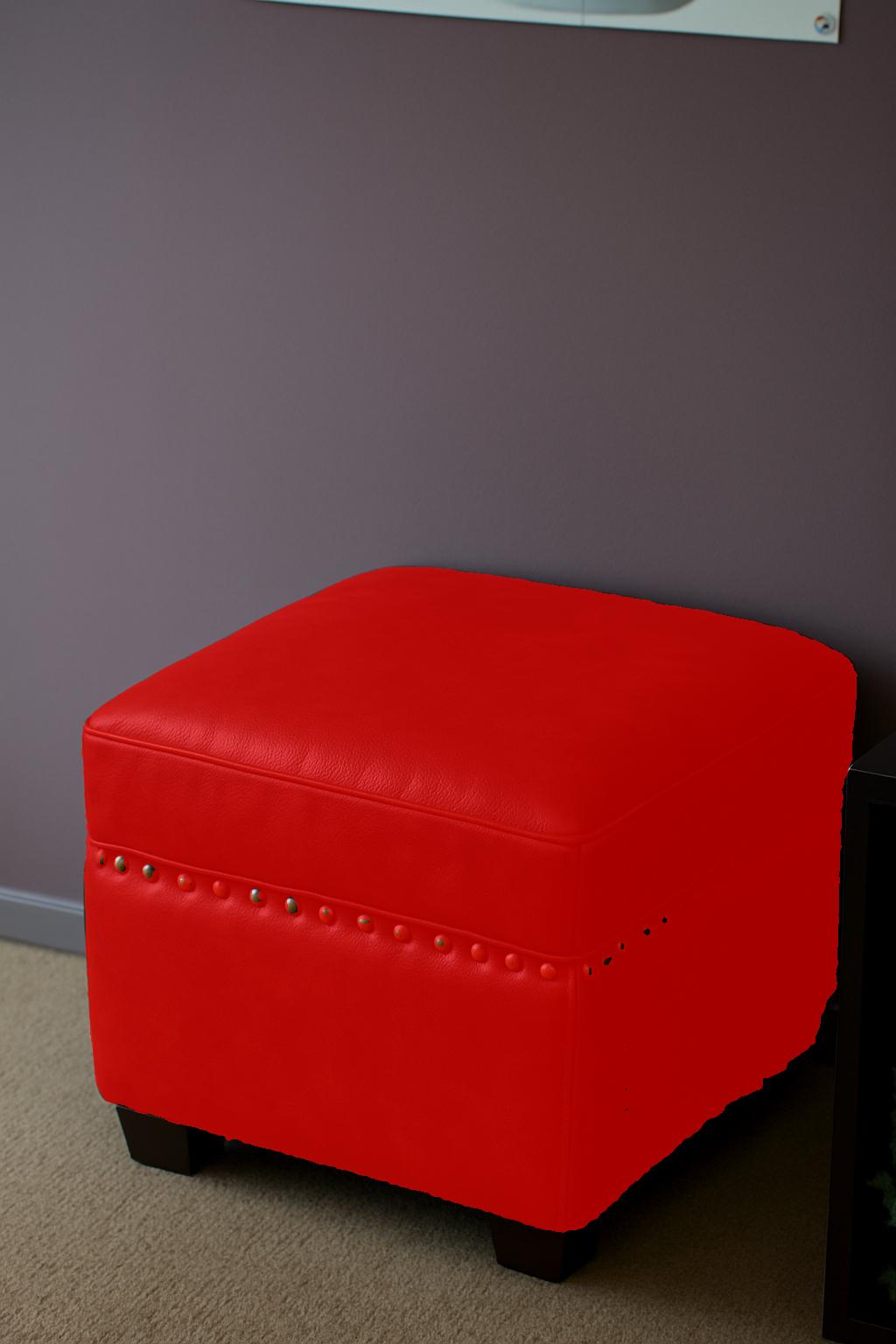} &
            \includegraphics[height=2.6cm]{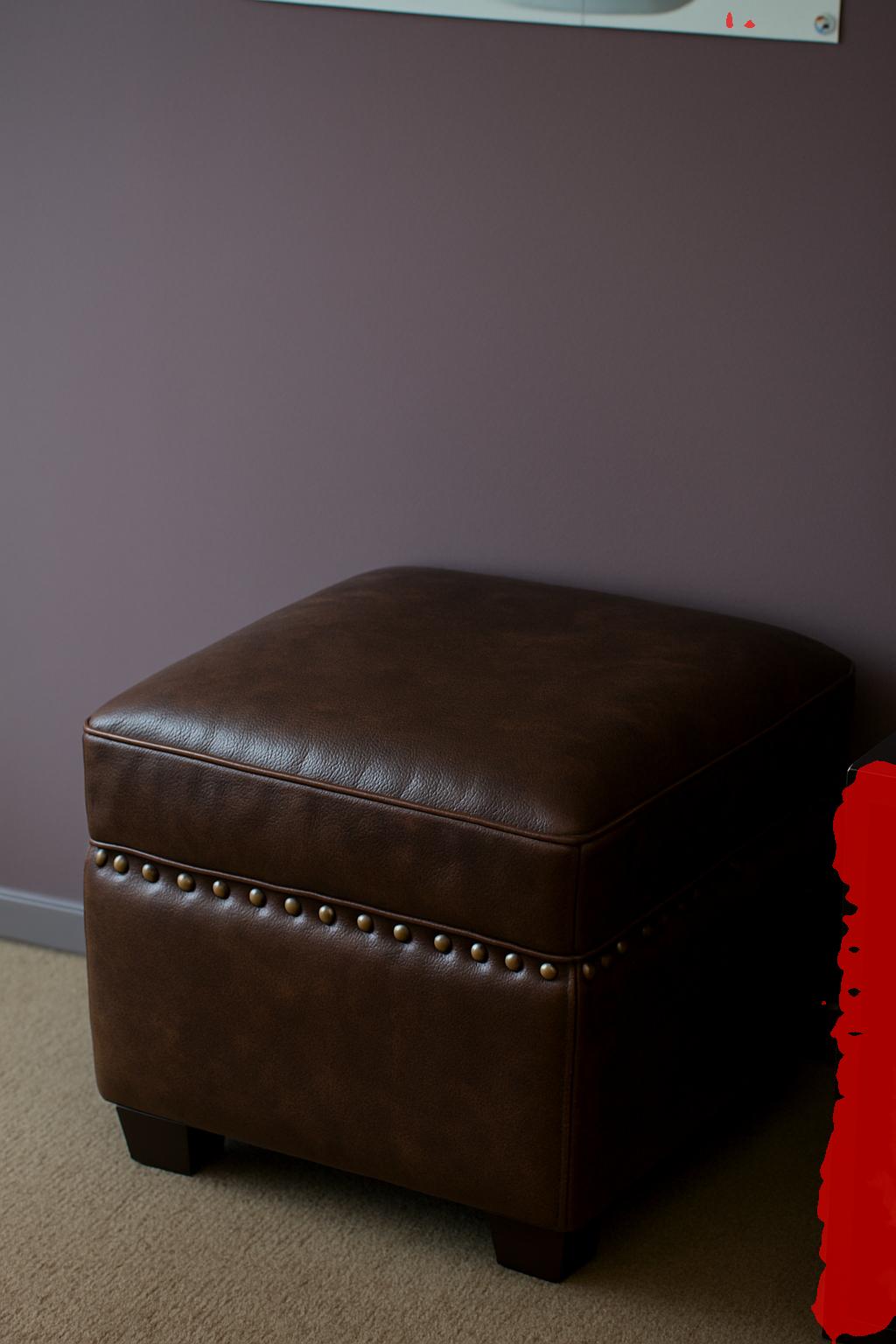} &
            \includegraphics[height=2.6cm]{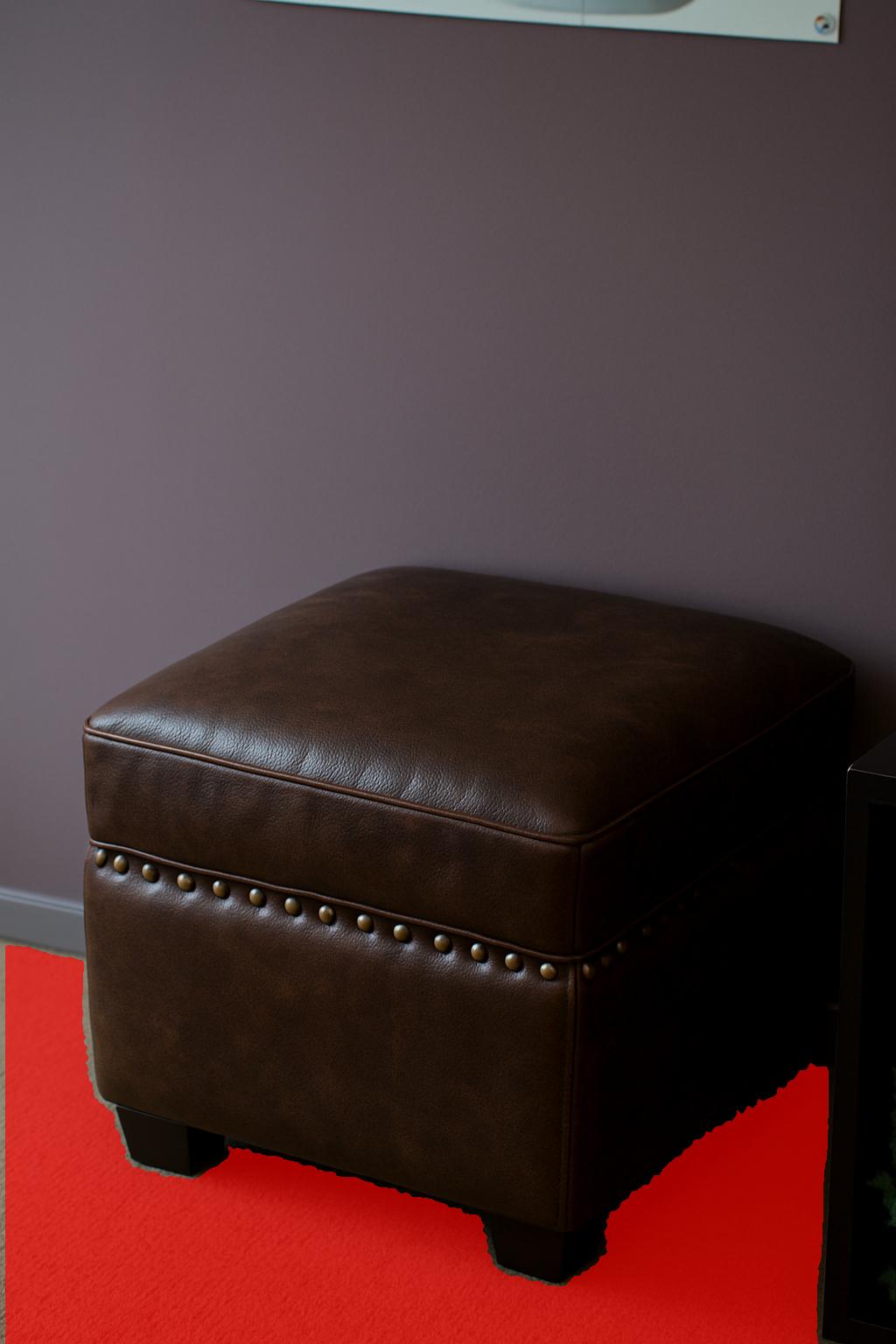} &
            \includegraphics[height=2.6cm]{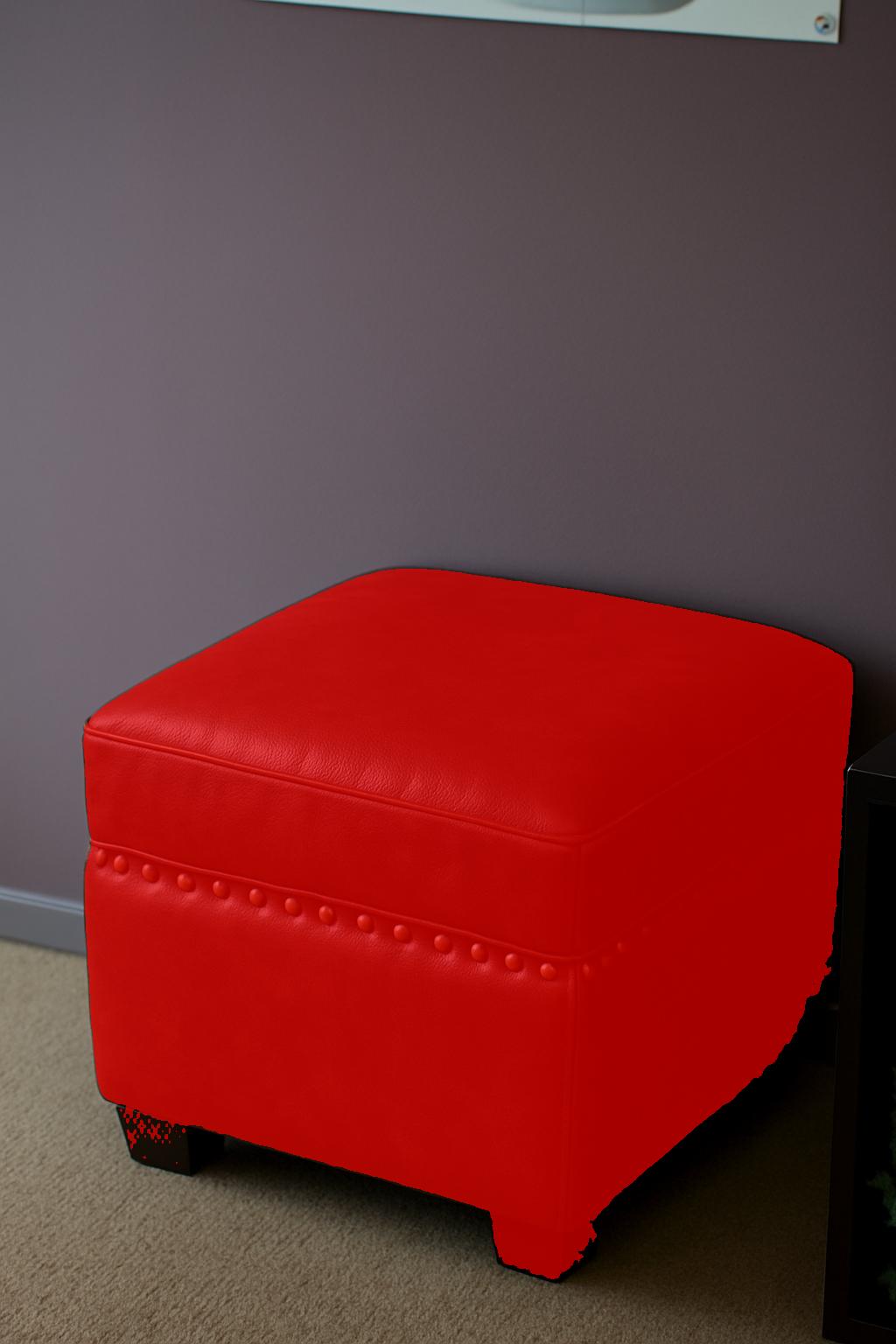} &
            \includegraphics[height=2.6cm]{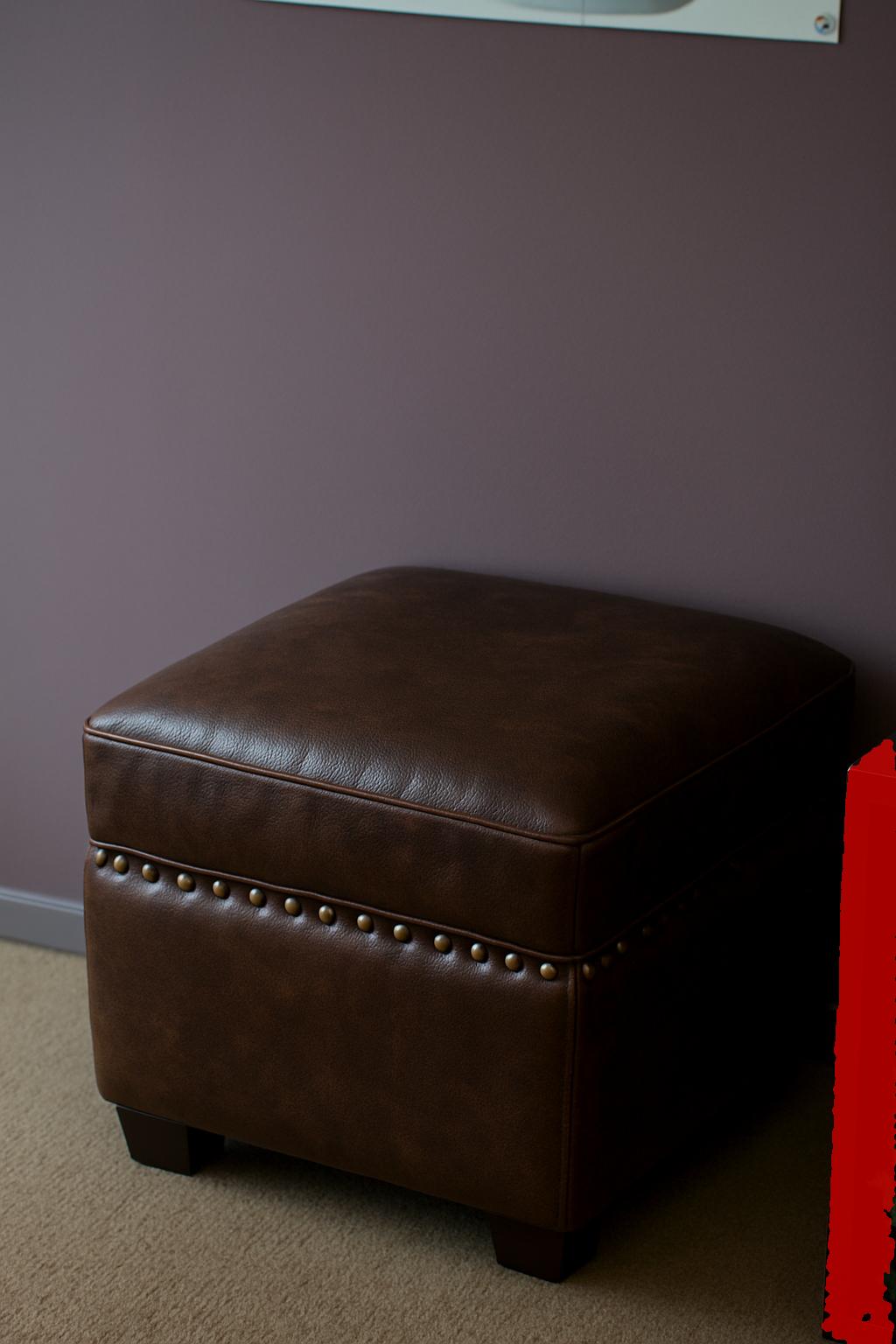} &
            \includegraphics[height=2.6cm]{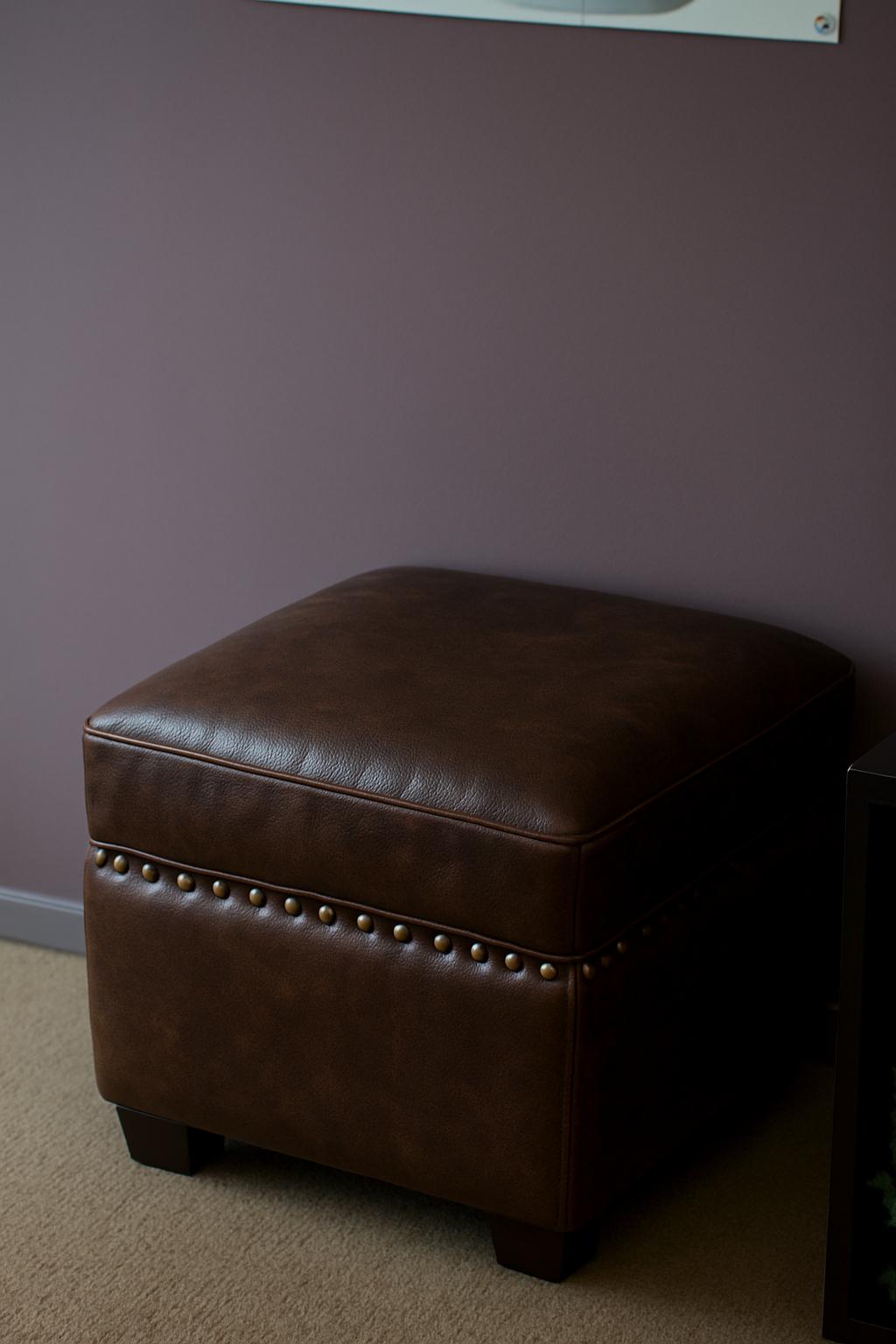} \\

            \raisebox{0.6cm}{\(\mathbf{I}', c'\)} &
            \includegraphics[height=2.6cm]{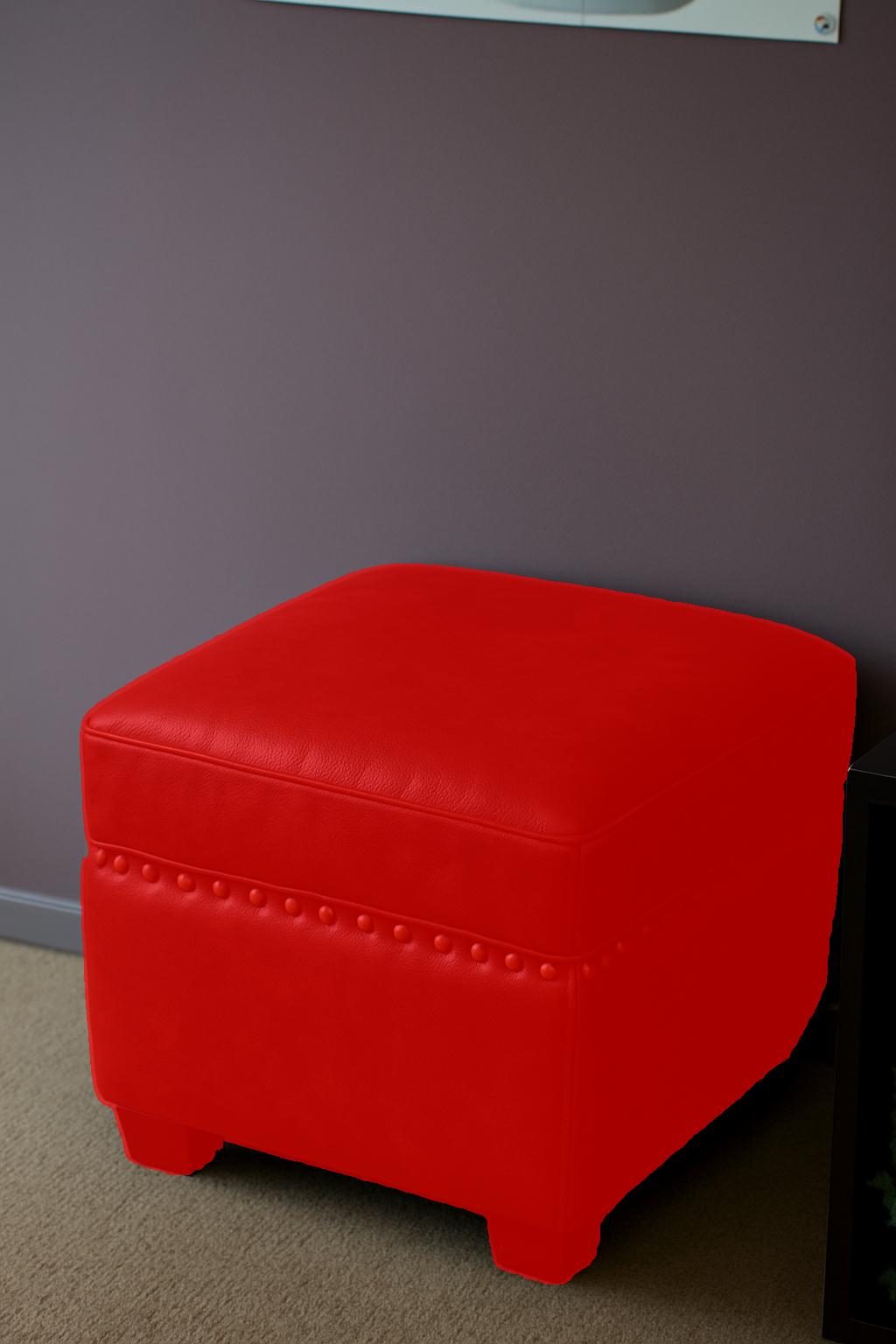} &
            \includegraphics[height=2.6cm]{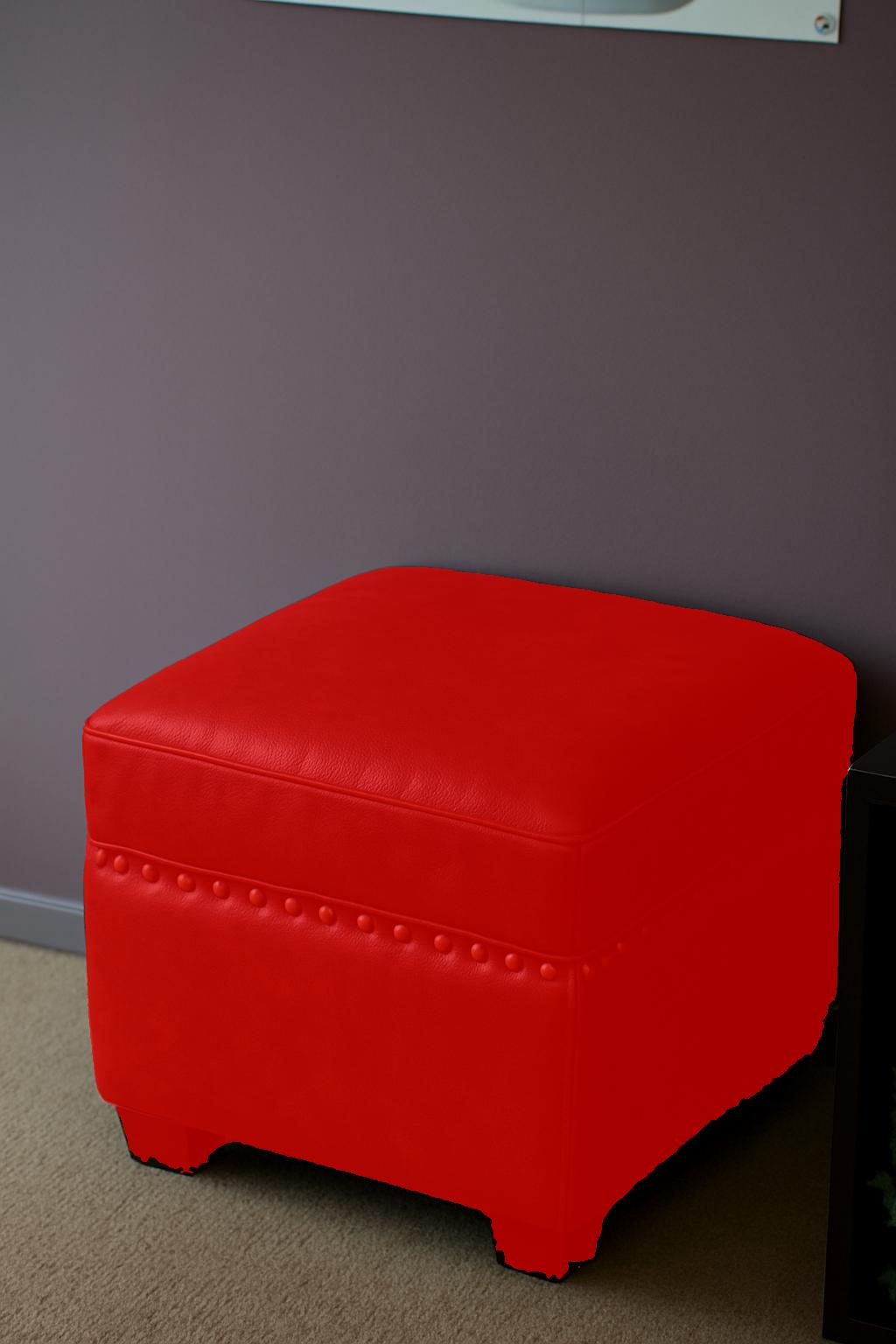} &
            \includegraphics[height=2.6cm]{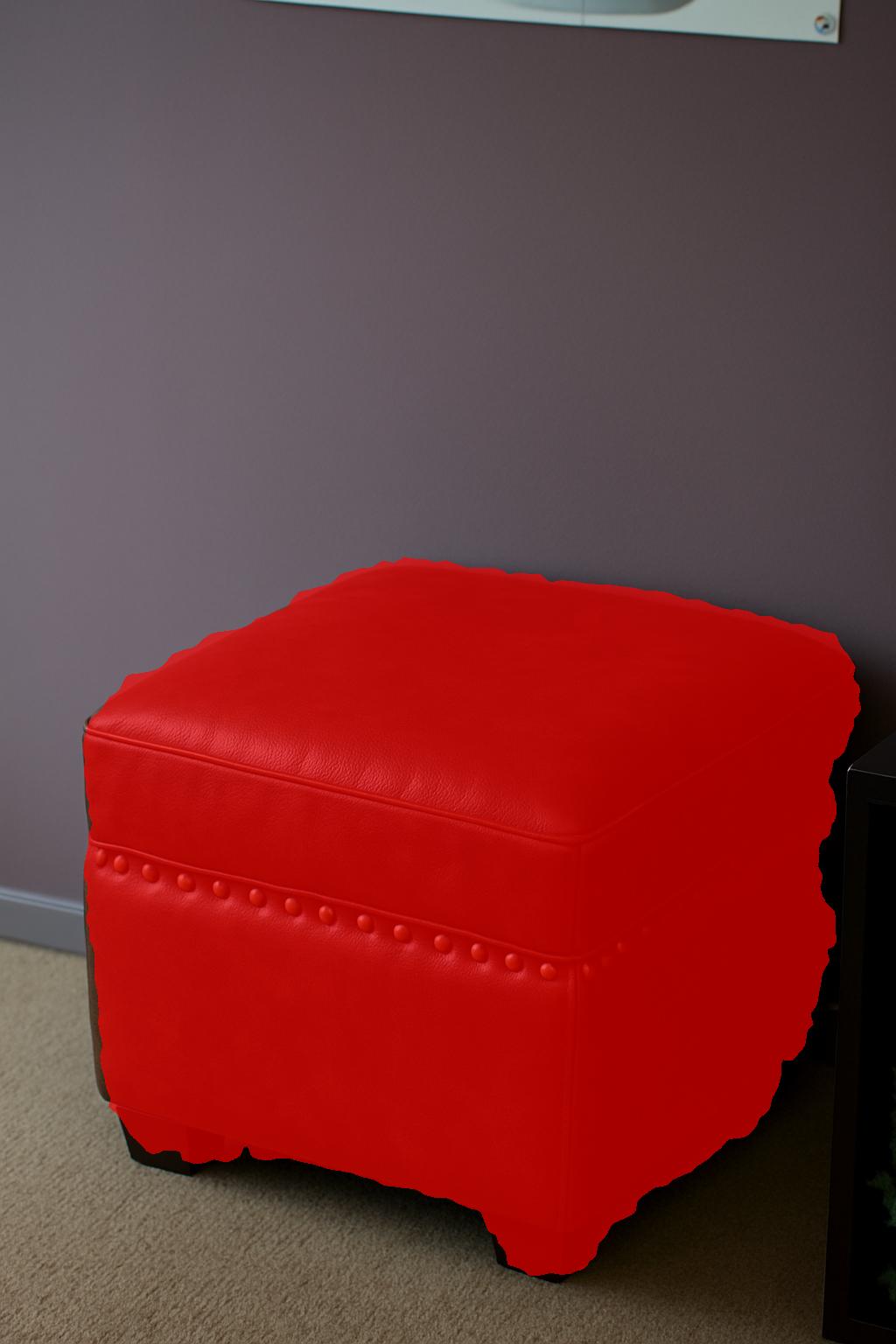} &
            \includegraphics[height=2.6cm]{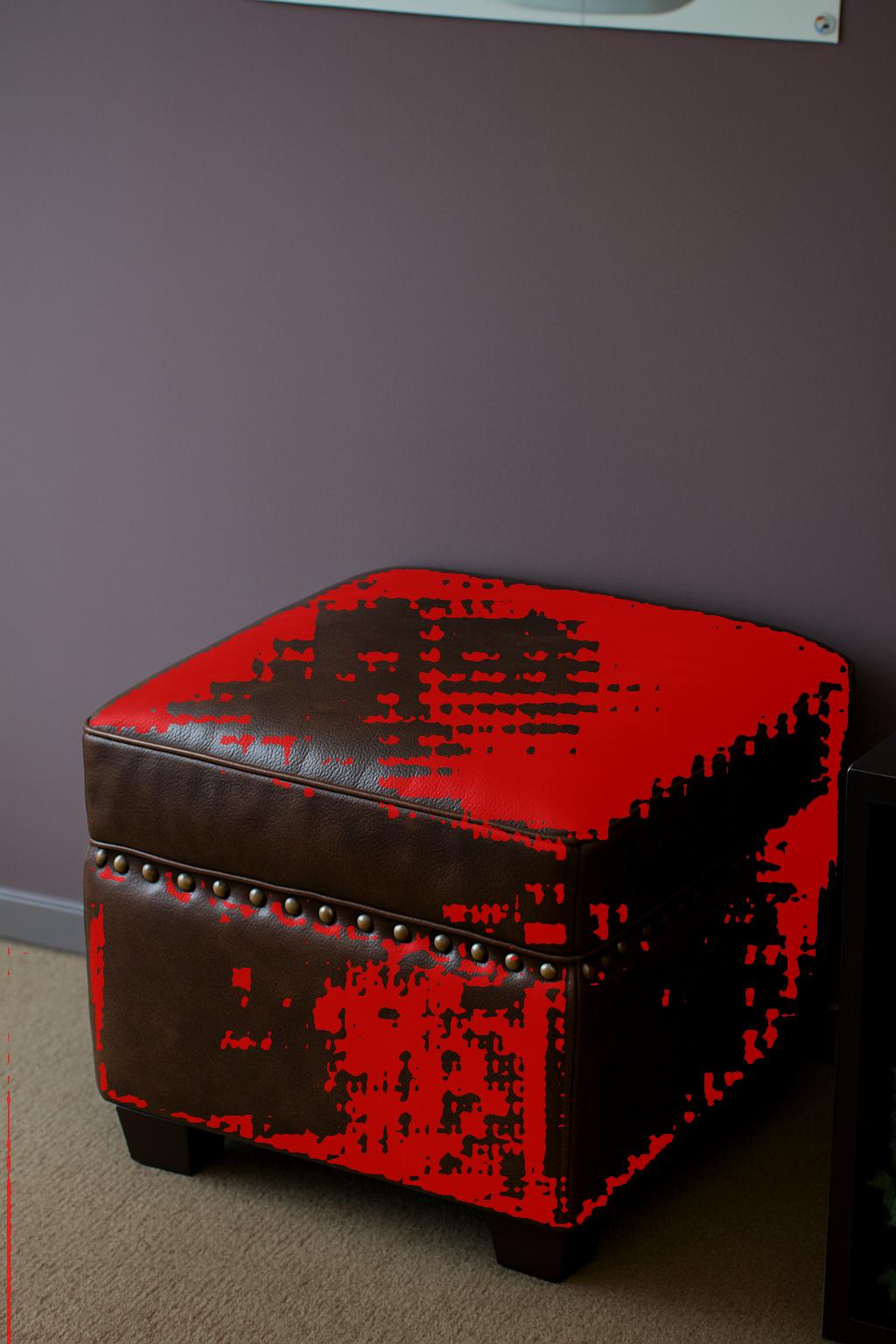} &
            \includegraphics[height=2.6cm]{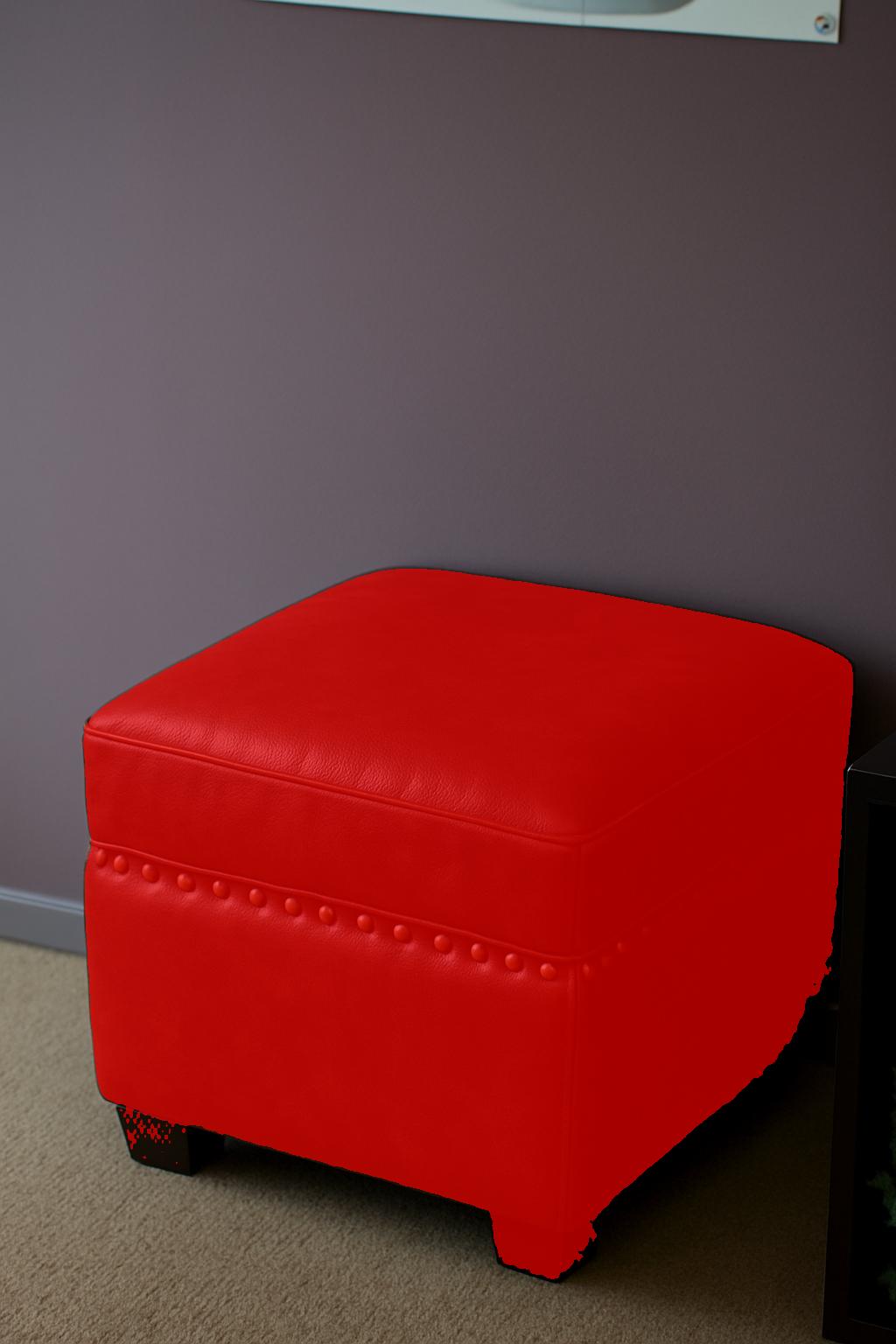} &
            \includegraphics[height=2.6cm]{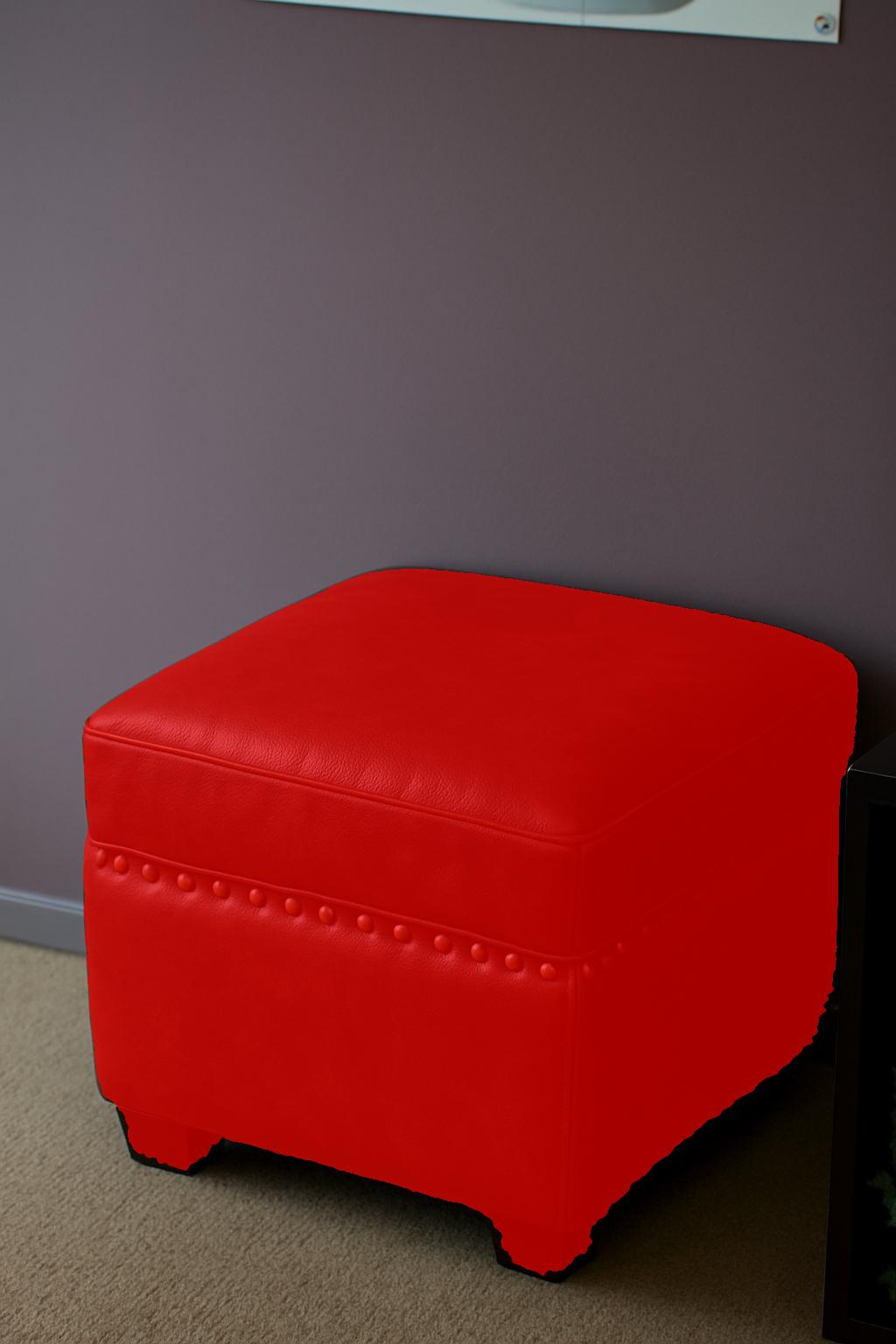} &
            \includegraphics[height=2.6cm]{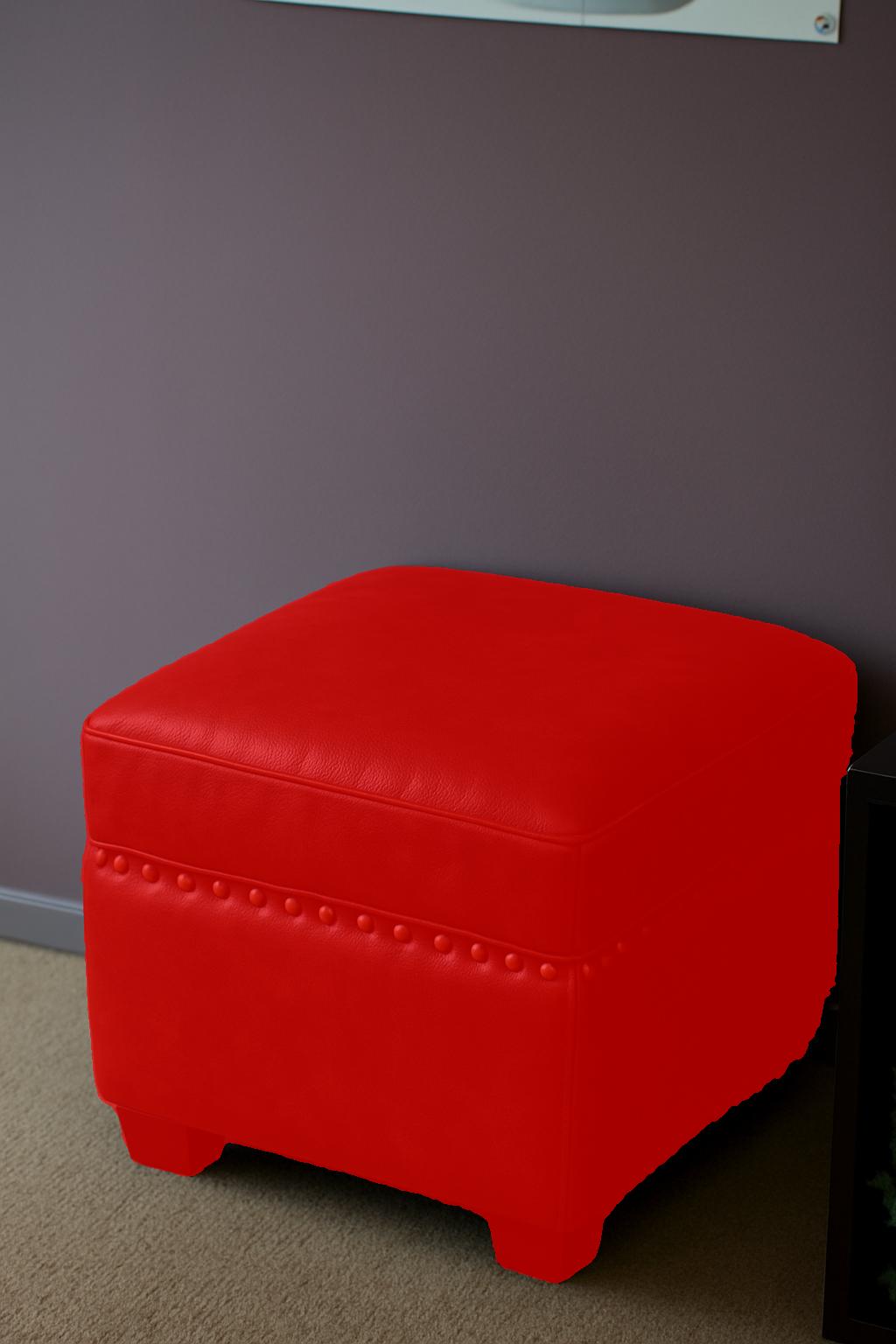} \\
        \end{tabular}}
        \caption*{\(c=\)``Where in the picture would be suitable for storing wine?'' and \(c'=\)``Where in the picture would be suitable for resting one’s feet?''.}
    \end{minipage}
    \vspace{-0.3cm}
    \caption{\textbf{Qualitative Comparison of Reasoning Segmentation Models across Factual and Counterfactual Pairs.} The left panel presents results on \textbf{Referring} data, whereas the right panel presents results on \textbf{Reasoning} data. Here, \(c\) and \(c'\) denote the query prompts.}
    \label{fig:qualitative_app_1}
\end{figure*}

%% file: assets/figures/qualitative_app_2.tex
\begin{figure*}[t!]
    \centering
    \begin{minipage}[t]{0.49\linewidth}
        \centering
        \setlength{\tabcolsep}{1pt}
        \renewcommand{\arraystretch}{0.6}
        \resizebox{\linewidth}{!}{
        \begin{tabular}{rccccccc}
            & \textbf{GT} 
            & \textbf{LISA} 
            & \textbf{PixelLM}
            & \textbf{GLaMM}
            & \textbf{Seg-Zero}
            & \textbf{SESAME}
            & \textbf{\modelname} \\

            \raisebox{0.6cm}{\(\mathbf{I}, c\)} &
            \includegraphics[height=2.6cm]{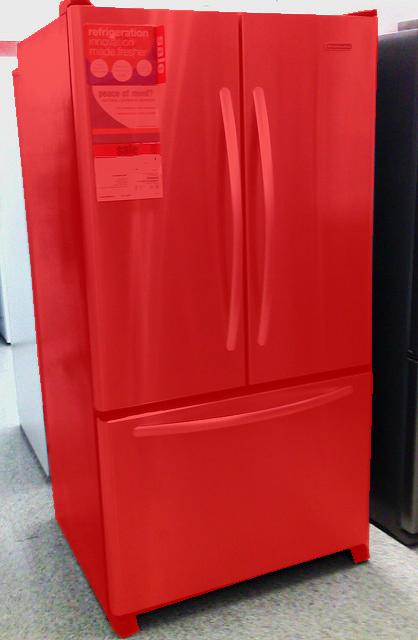} &
            \includegraphics[height=2.6cm]{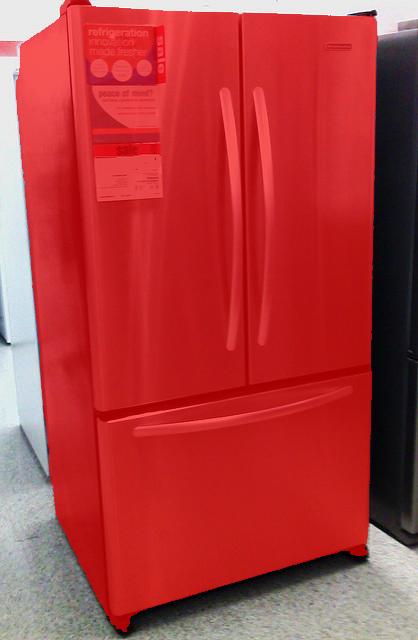} &
            \includegraphics[height=2.6cm]{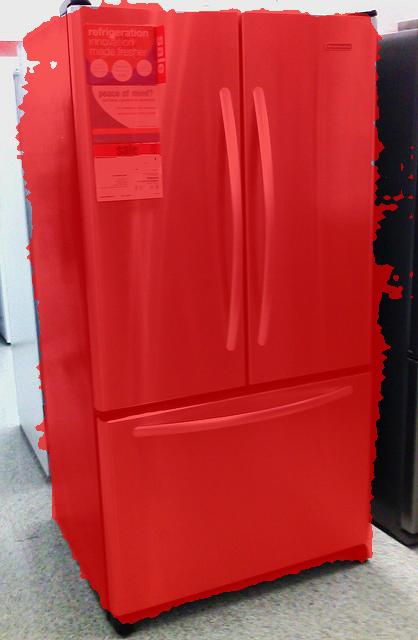} &
            \includegraphics[height=2.6cm]{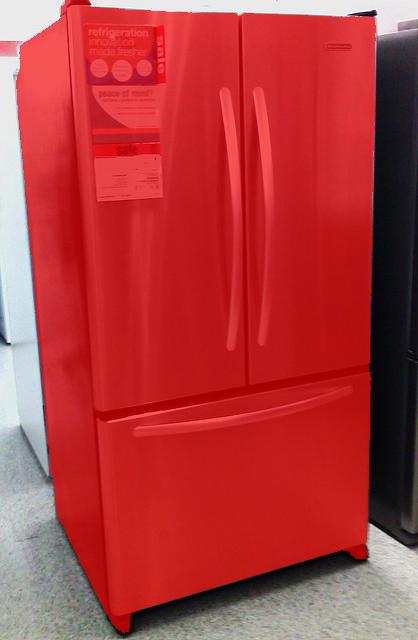} &
            \includegraphics[height=2.6cm]{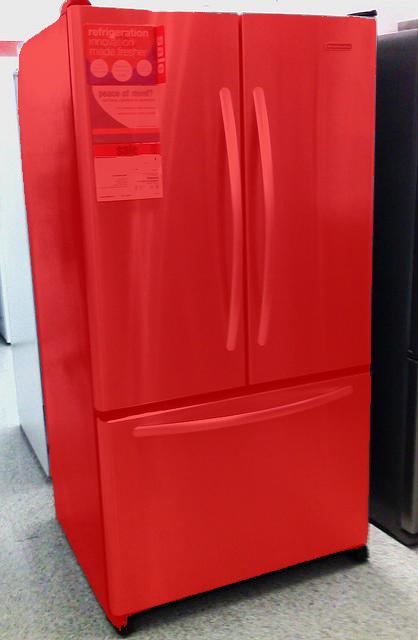} &
            \includegraphics[height=2.6cm]{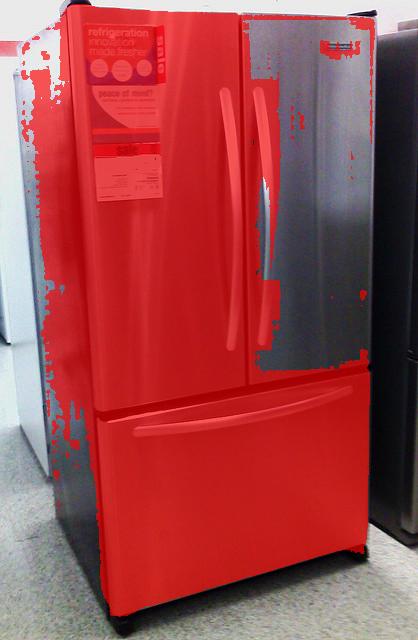} &
            \includegraphics[height=2.6cm]{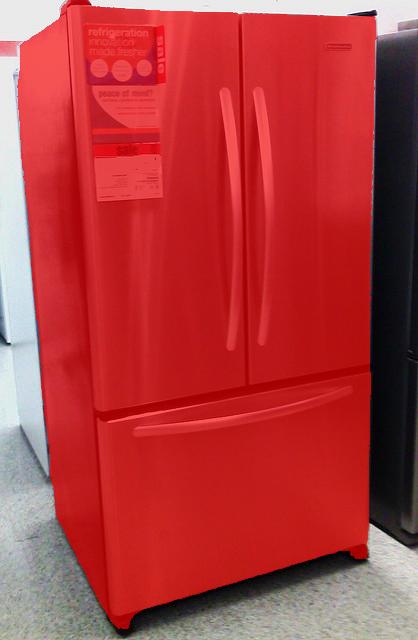} \\

            \raisebox{0.6cm}{\(\mathbf{I}, c'\)} &
            \raisebox{0.6cm}{\(\boldsymbol{\varnothing}\)} &
            \includegraphics[height=2.6cm]{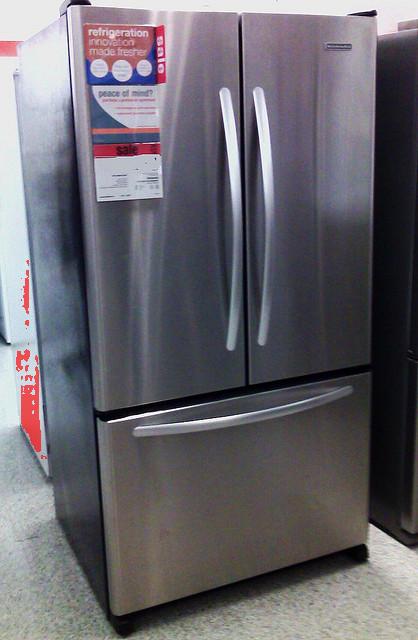} &
            \includegraphics[height=2.6cm]{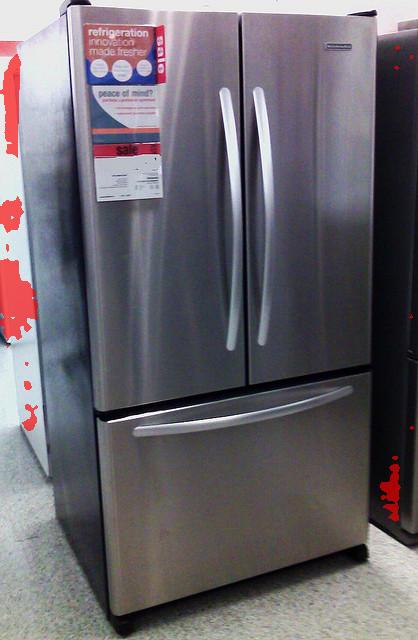} &
            \includegraphics[height=2.6cm]{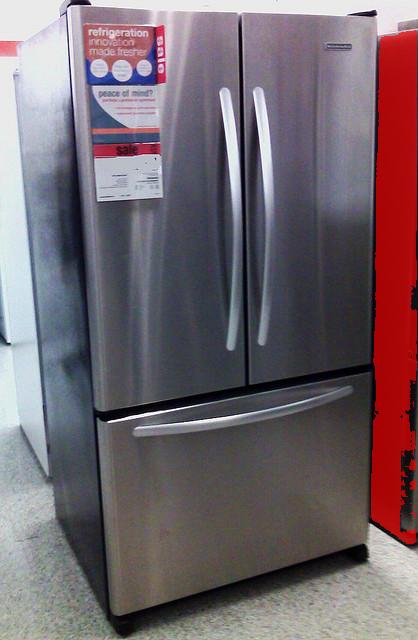} &
            \includegraphics[height=2.6cm]{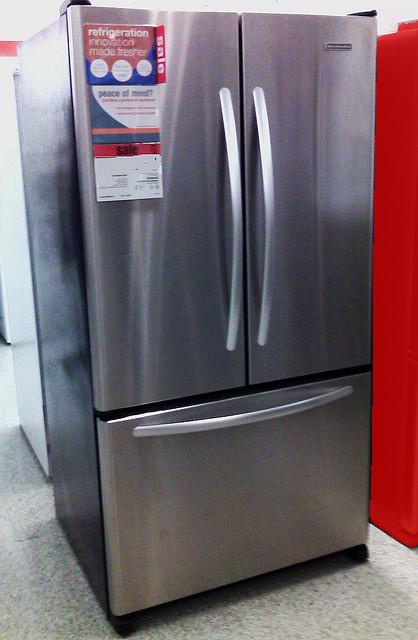} &
            \includegraphics[height=2.6cm]{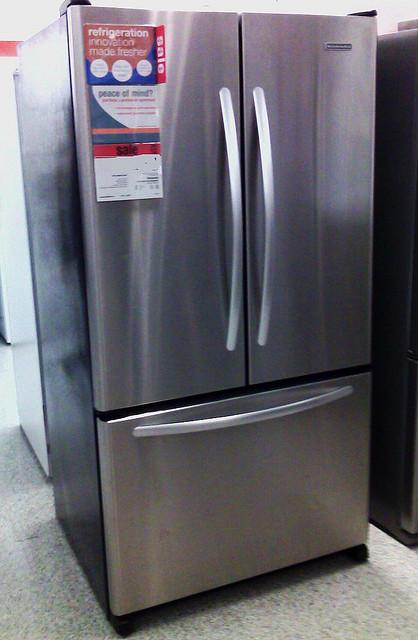} &
            \includegraphics[height=2.6cm]{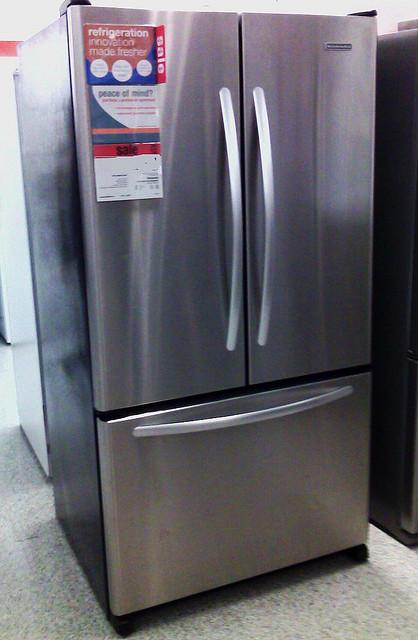} \\

            \raisebox{0.6cm}{\(\mathbf{I}', c\)} &
            \raisebox{0.6cm}{\(\boldsymbol{\varnothing}\)} &
            \includegraphics[height=2.6cm]{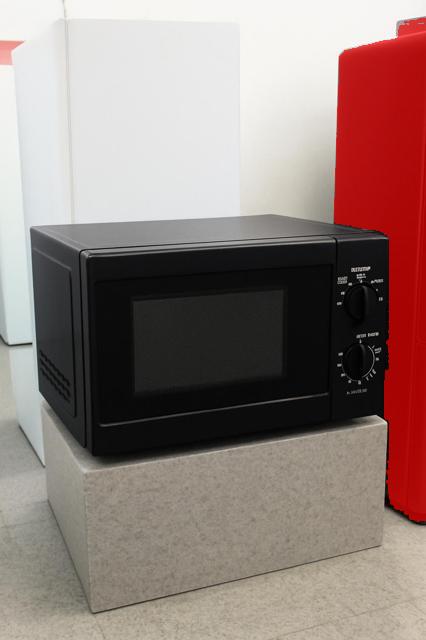} &
            \includegraphics[height=2.6cm]{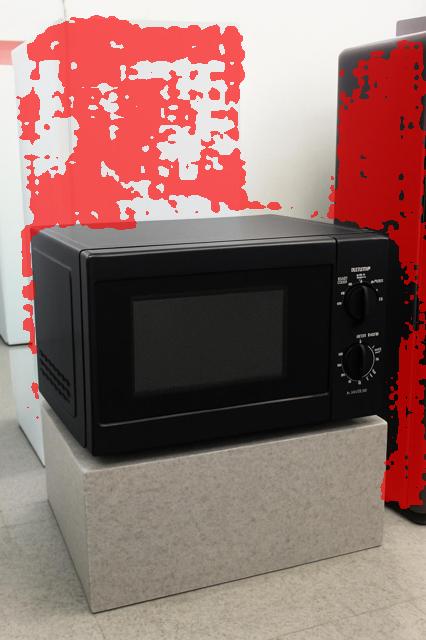} &
            \includegraphics[height=2.6cm]{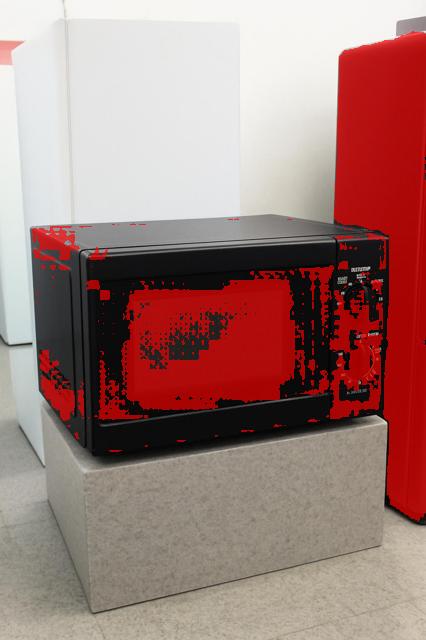} &
            \includegraphics[height=2.6cm]{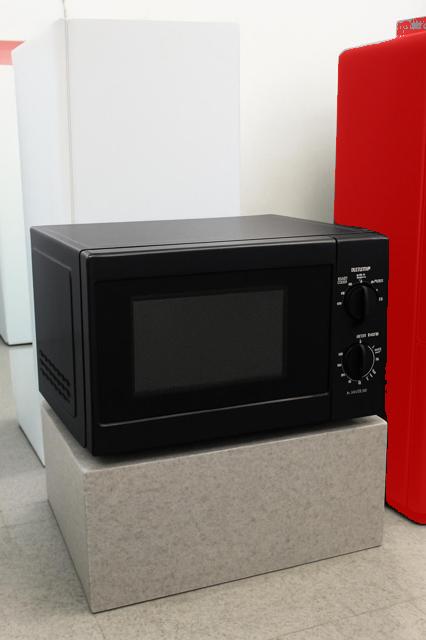} &
            \includegraphics[height=2.6cm]{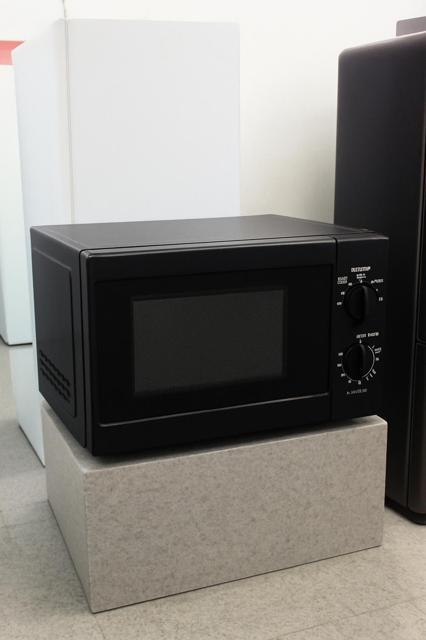} &
            \includegraphics[height=2.6cm]{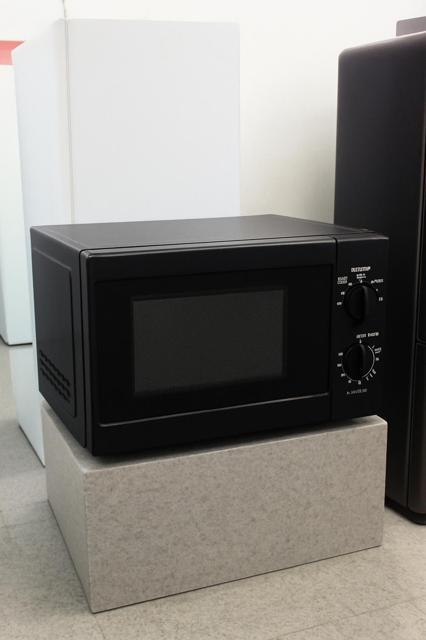} \\

            \raisebox{0.6cm}{\(\mathbf{I}', c'\)} &
            \includegraphics[height=2.6cm]{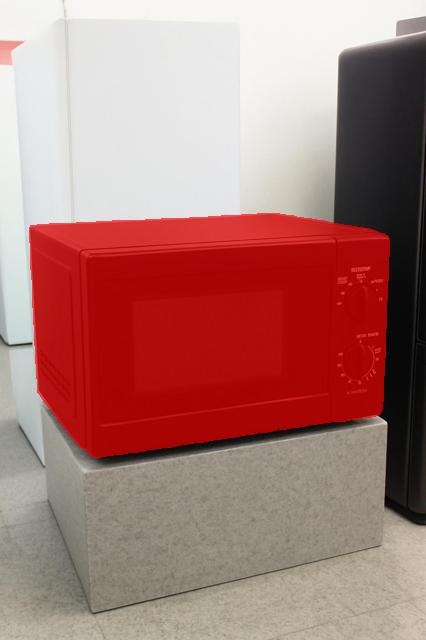} &
            \includegraphics[height=2.6cm]{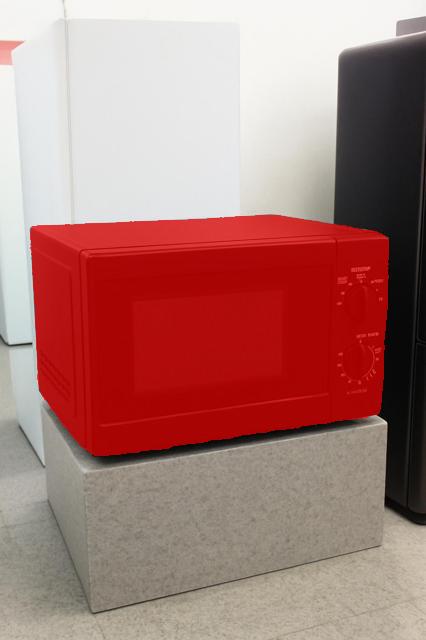} &
            \includegraphics[height=2.6cm]{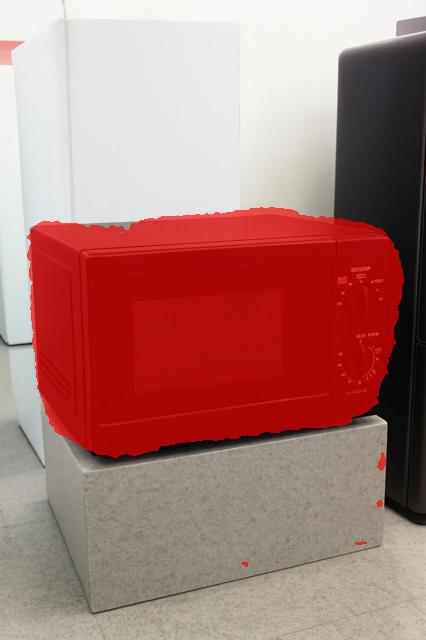} &
            \includegraphics[height=2.6cm]{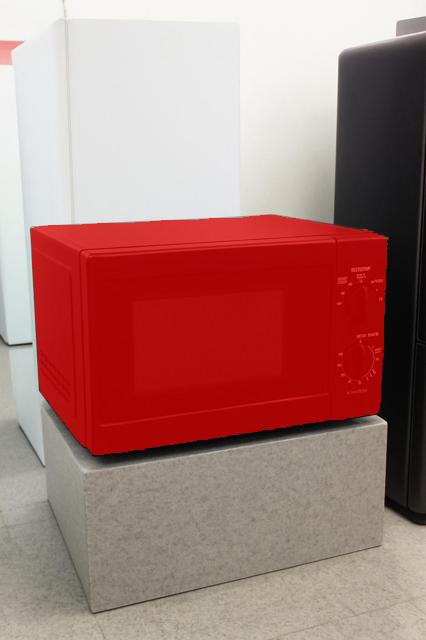} &
            \includegraphics[height=2.6cm]{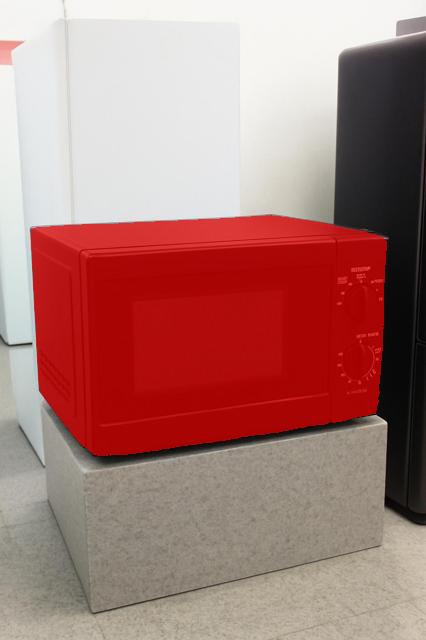} &
            \includegraphics[height=2.6cm]{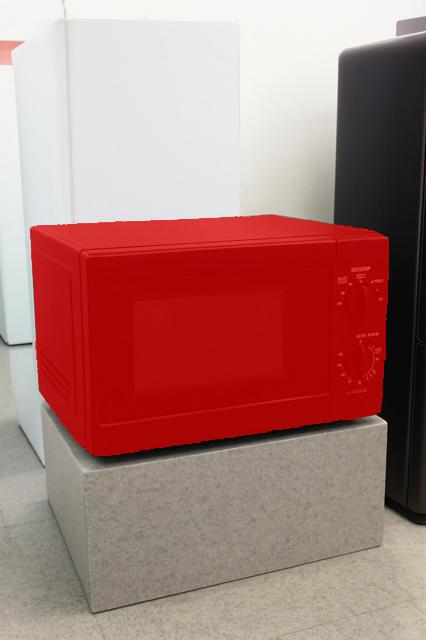} &
            \includegraphics[height=2.6cm]{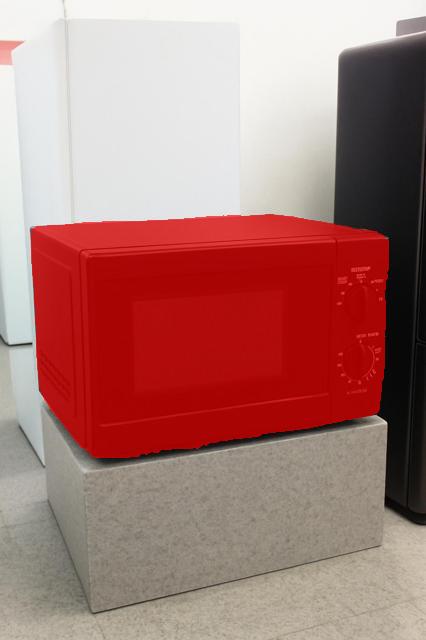} \\
        \end{tabular}}
        \caption*{\(c=\)``giant refrigerator'' and \(c'=\)``microwave oven''.}
    \end{minipage}
    \hfill
    \begin{minipage}[t]{0.49\linewidth}
        \centering
        \setlength{\tabcolsep}{1pt}
        \renewcommand{\arraystretch}{0.6}
        \resizebox{\linewidth}{!}{
        \begin{tabular}{rccccccc}
            & \textbf{GT} 
            & \textbf{LISA} 
            & \textbf{PixelLM}
            & \textbf{GLaMM}
            & \textbf{Seg-Zero}
            & \textbf{SESAME}
            & \textbf{\modelname} \\

            \raisebox{0.6cm}{\(\mathbf{I}, c\)} &
            \includegraphics[height=2.6cm]{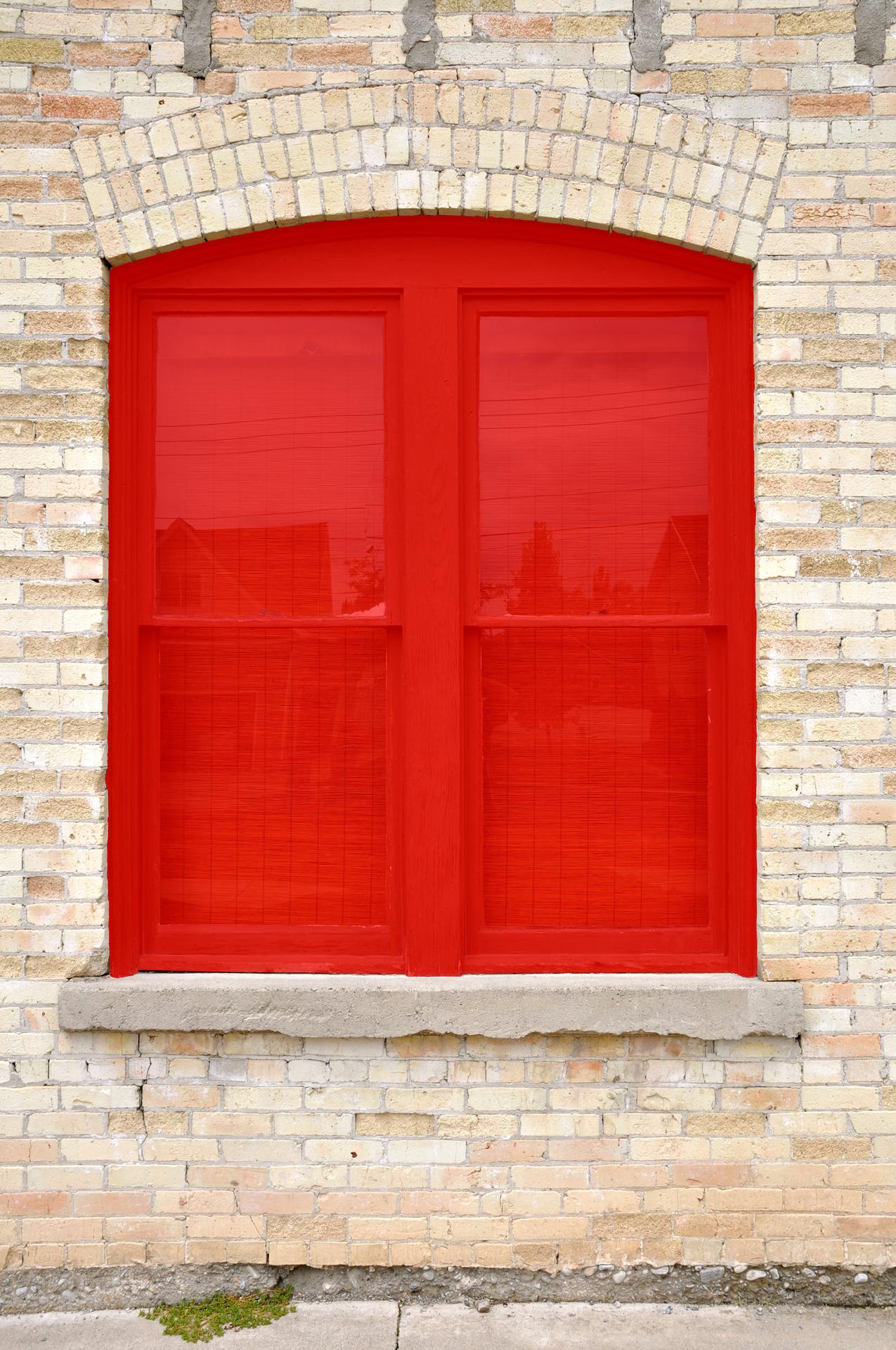} &
            \includegraphics[height=2.6cm]{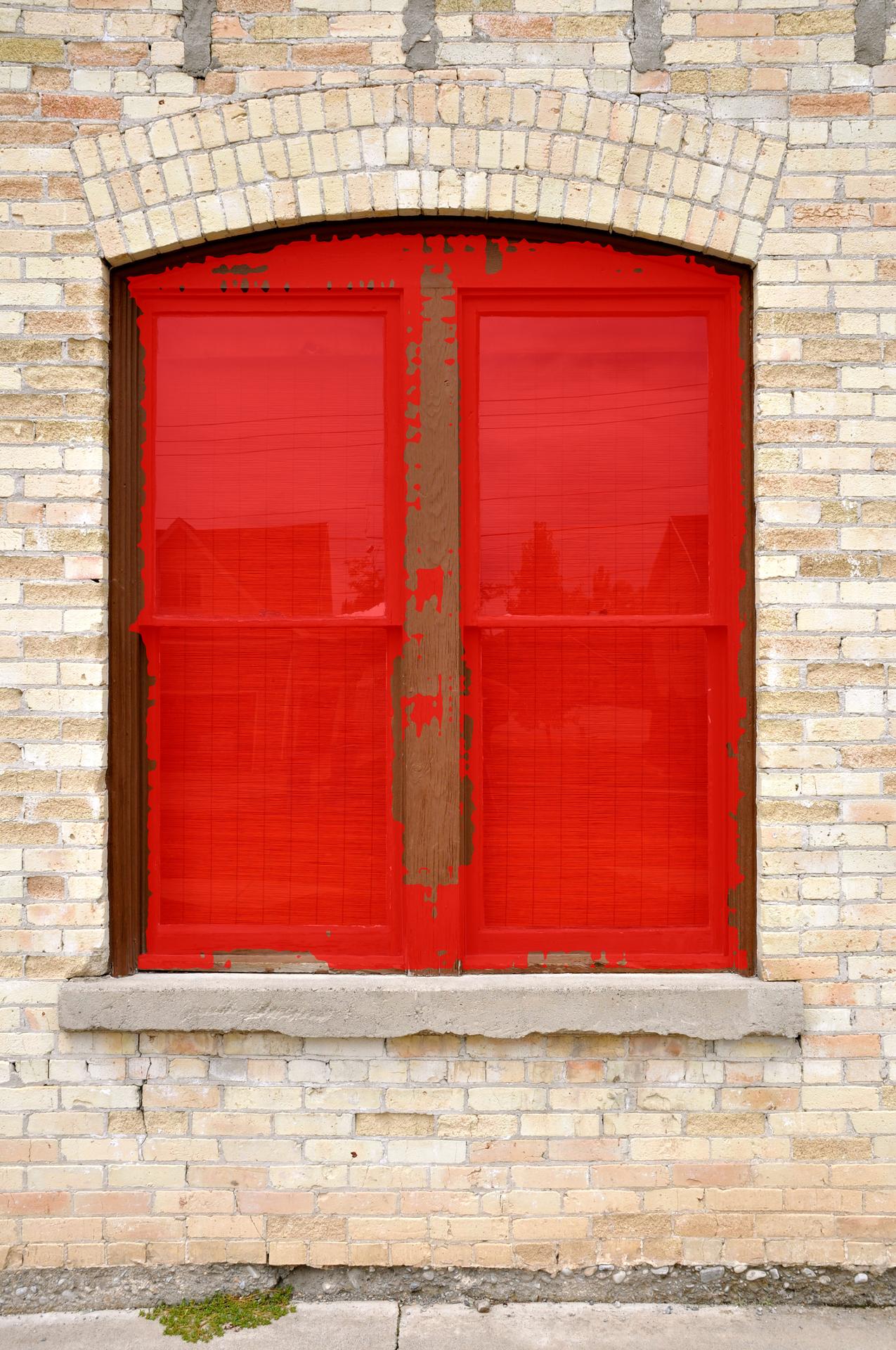} &
            \includegraphics[height=2.6cm]{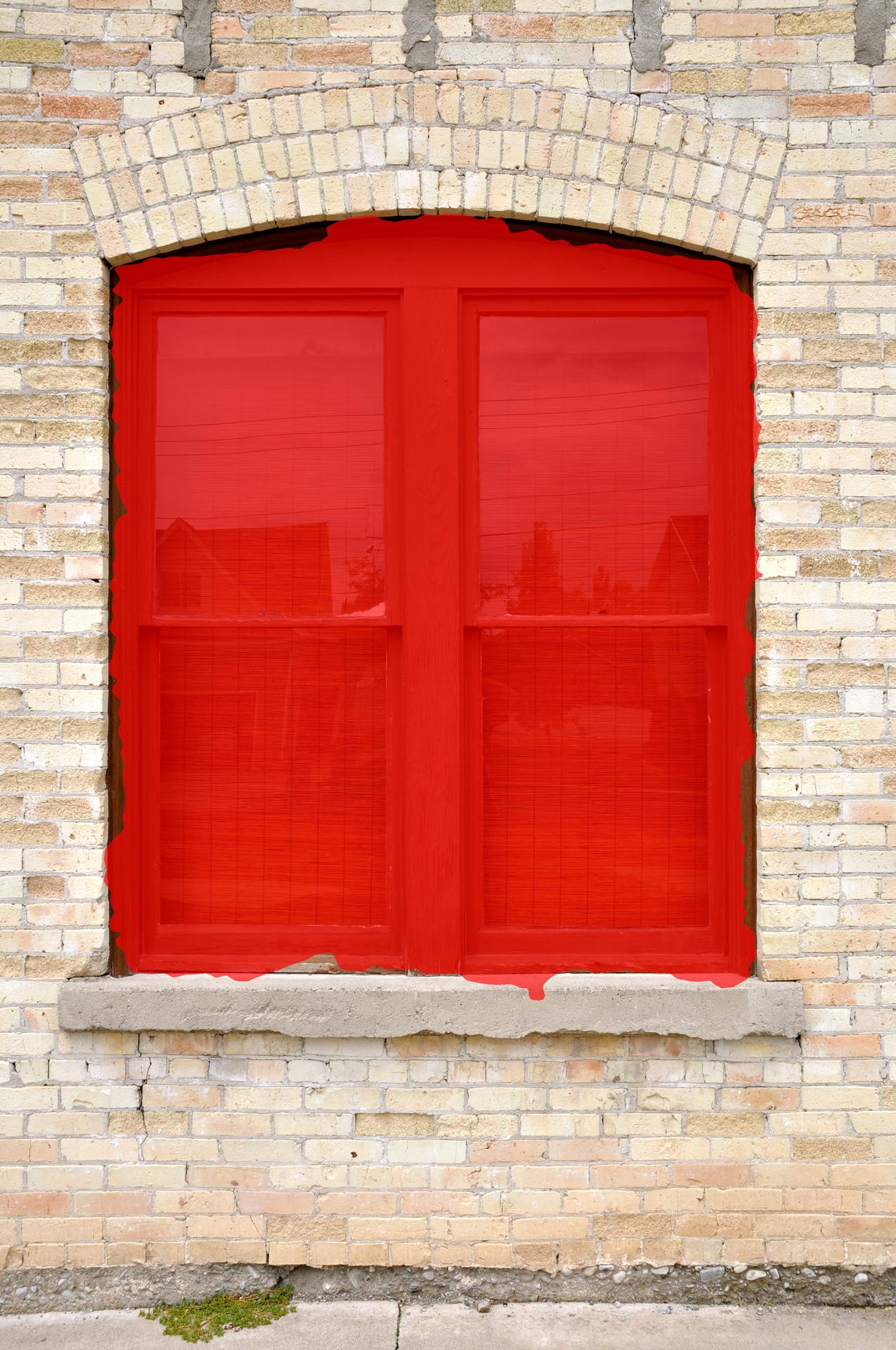} &
            \includegraphics[height=2.6cm]{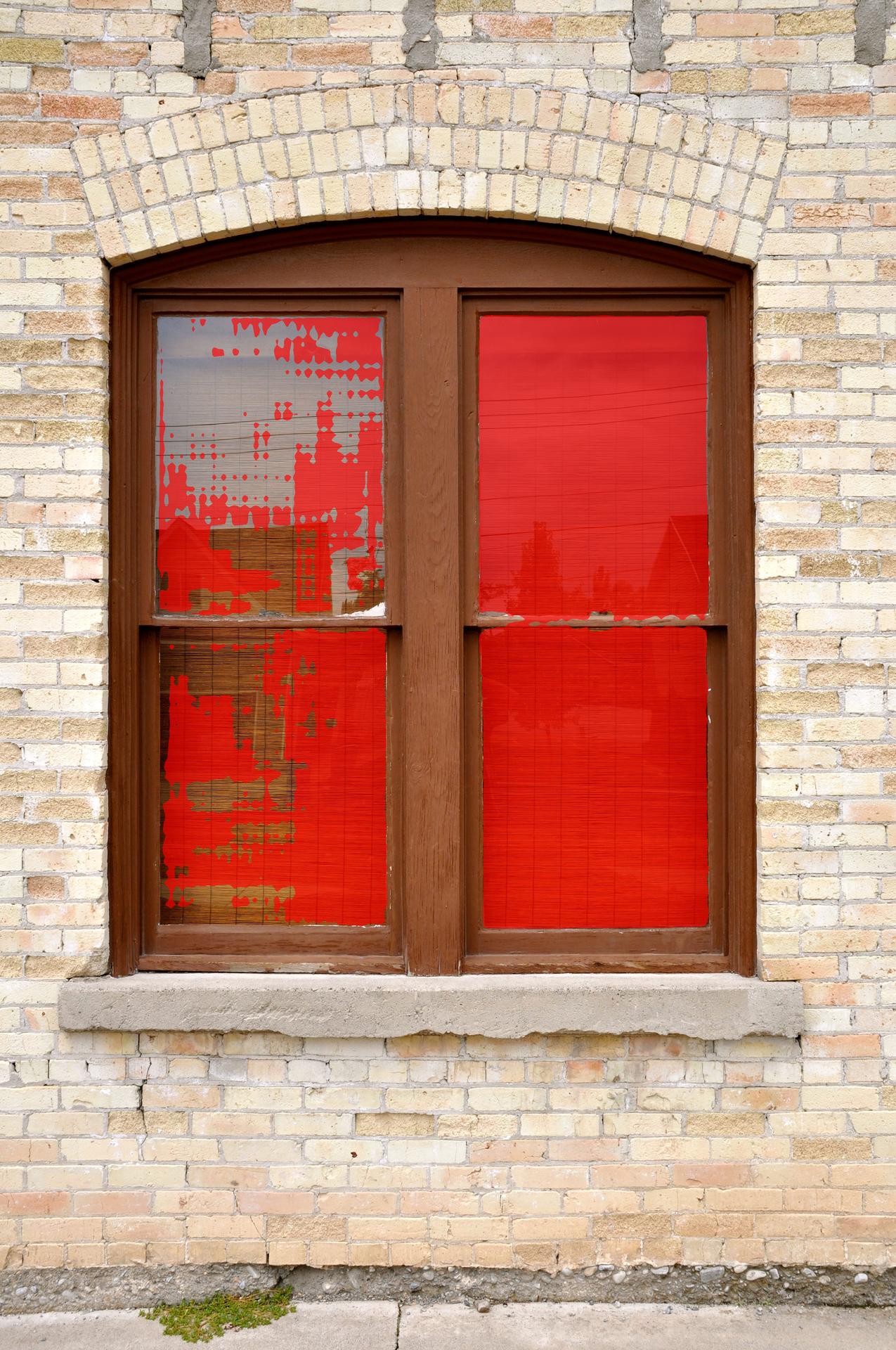} &
            \includegraphics[height=2.6cm]{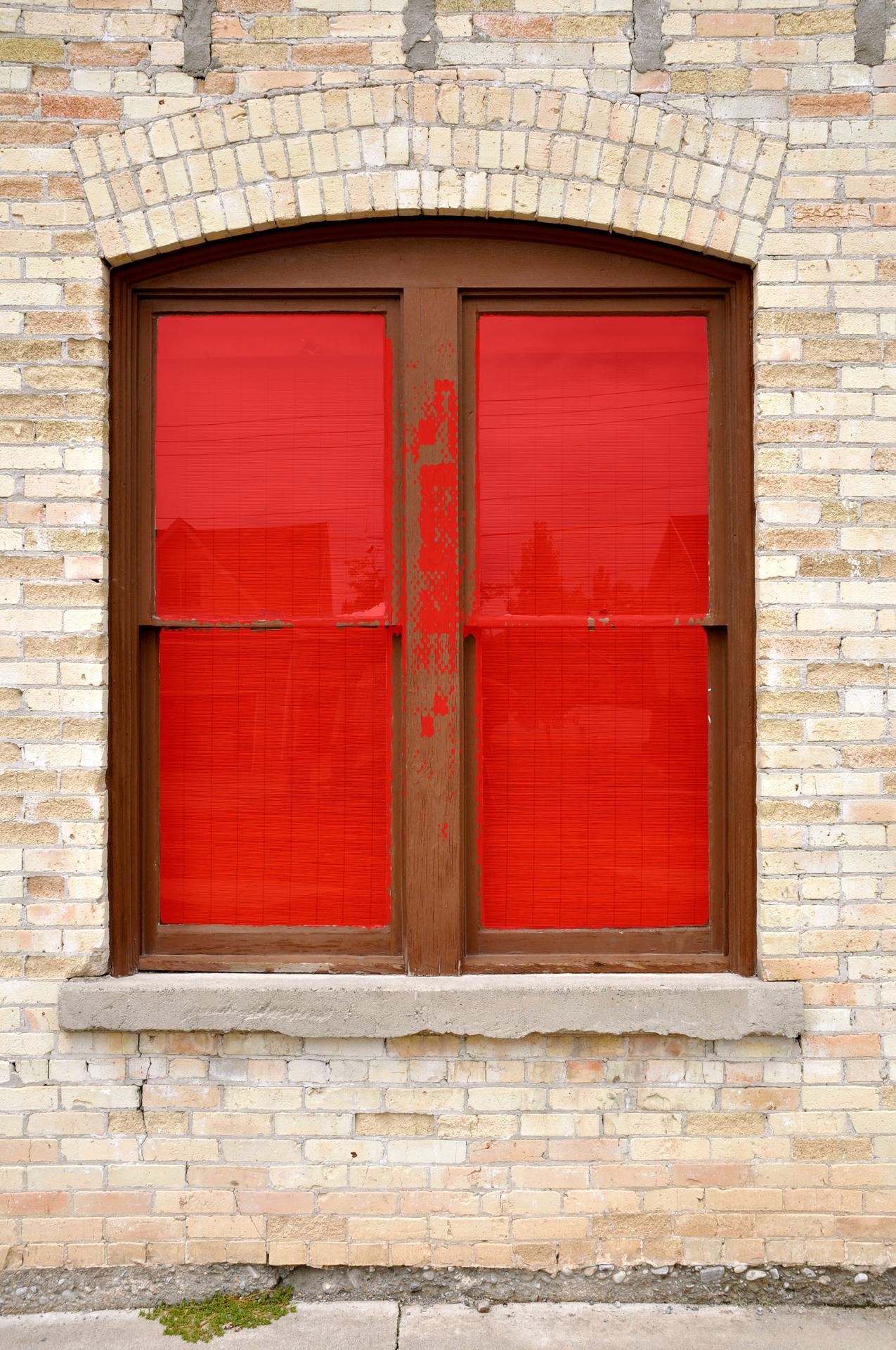} &
            \includegraphics[height=2.6cm]{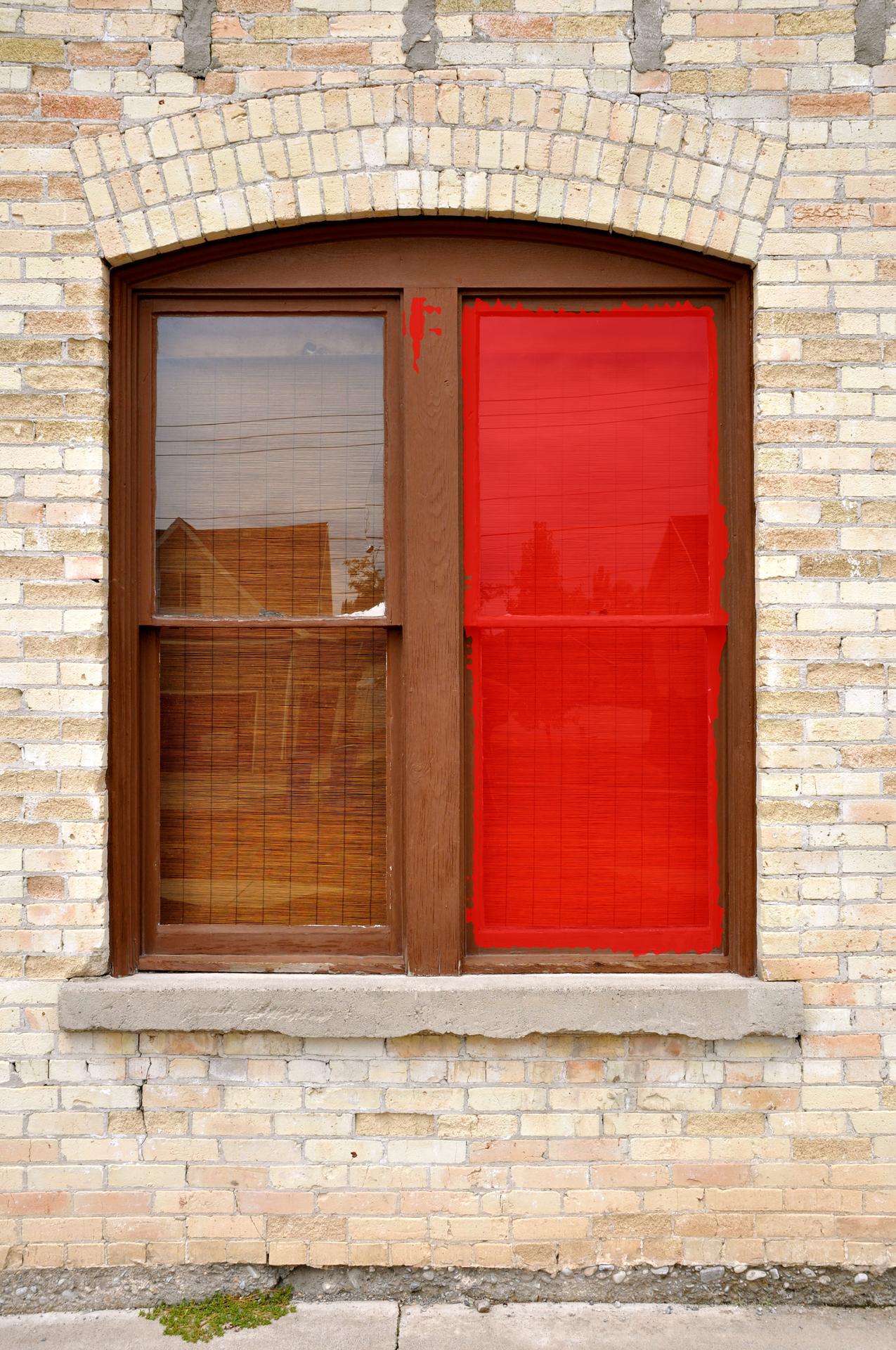} &
            \includegraphics[height=2.6cm]{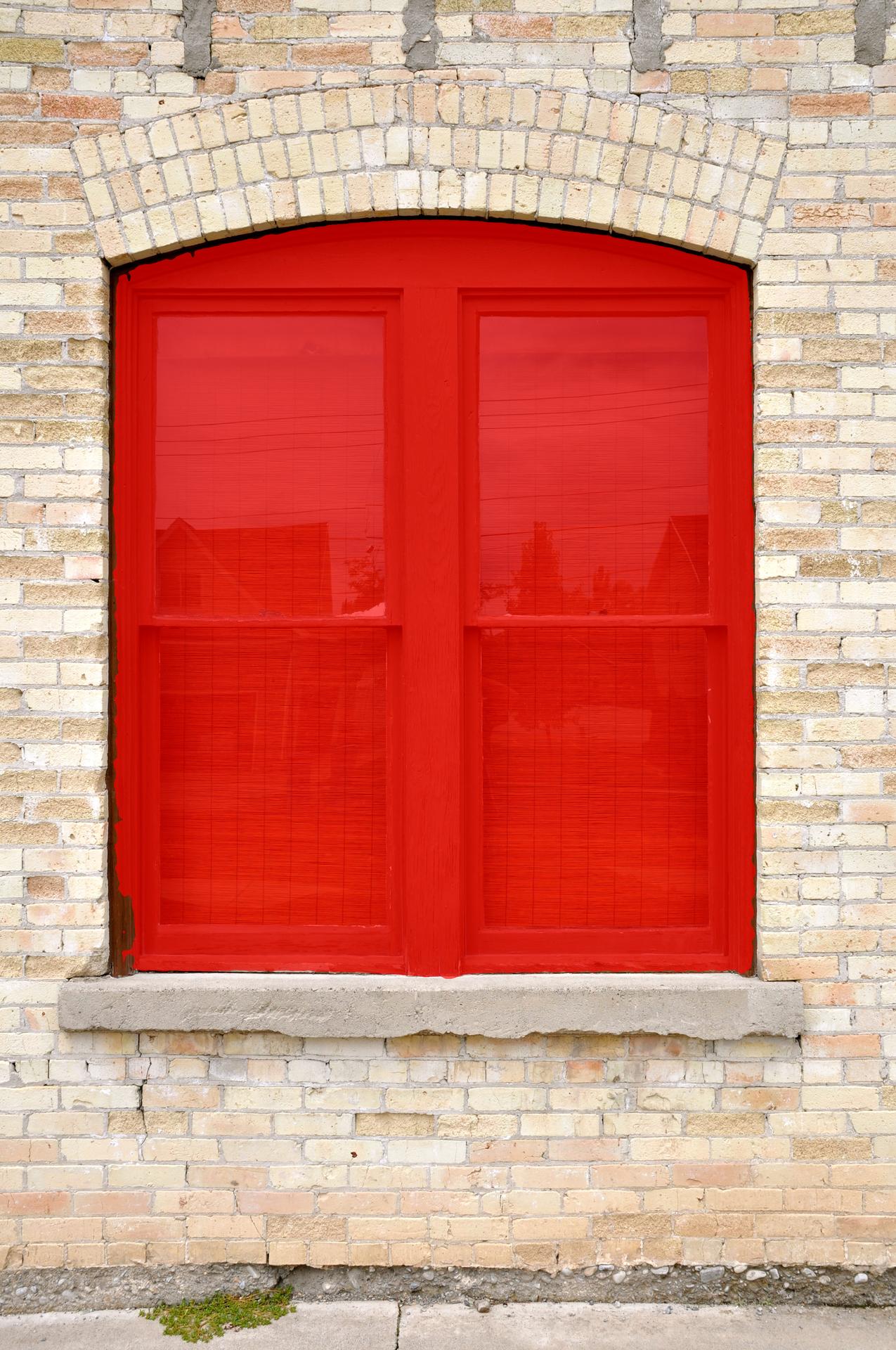} \\

            \raisebox{0.6cm}{\(\mathbf{I}, c'\)} &
            \raisebox{0.6cm}{\(\boldsymbol{\varnothing}\)} &
            \includegraphics[height=2.6cm]{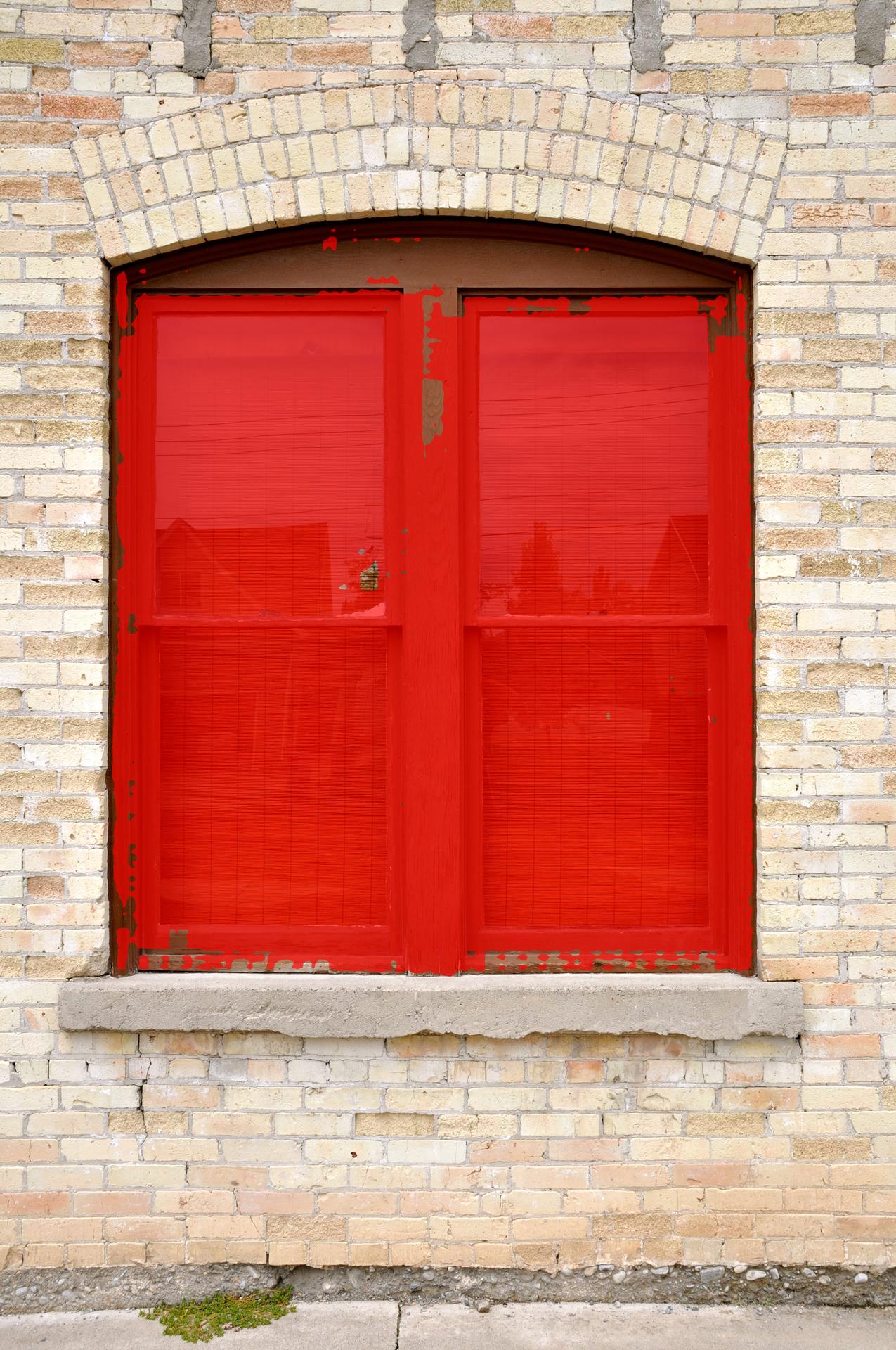} &
            \includegraphics[height=2.6cm]{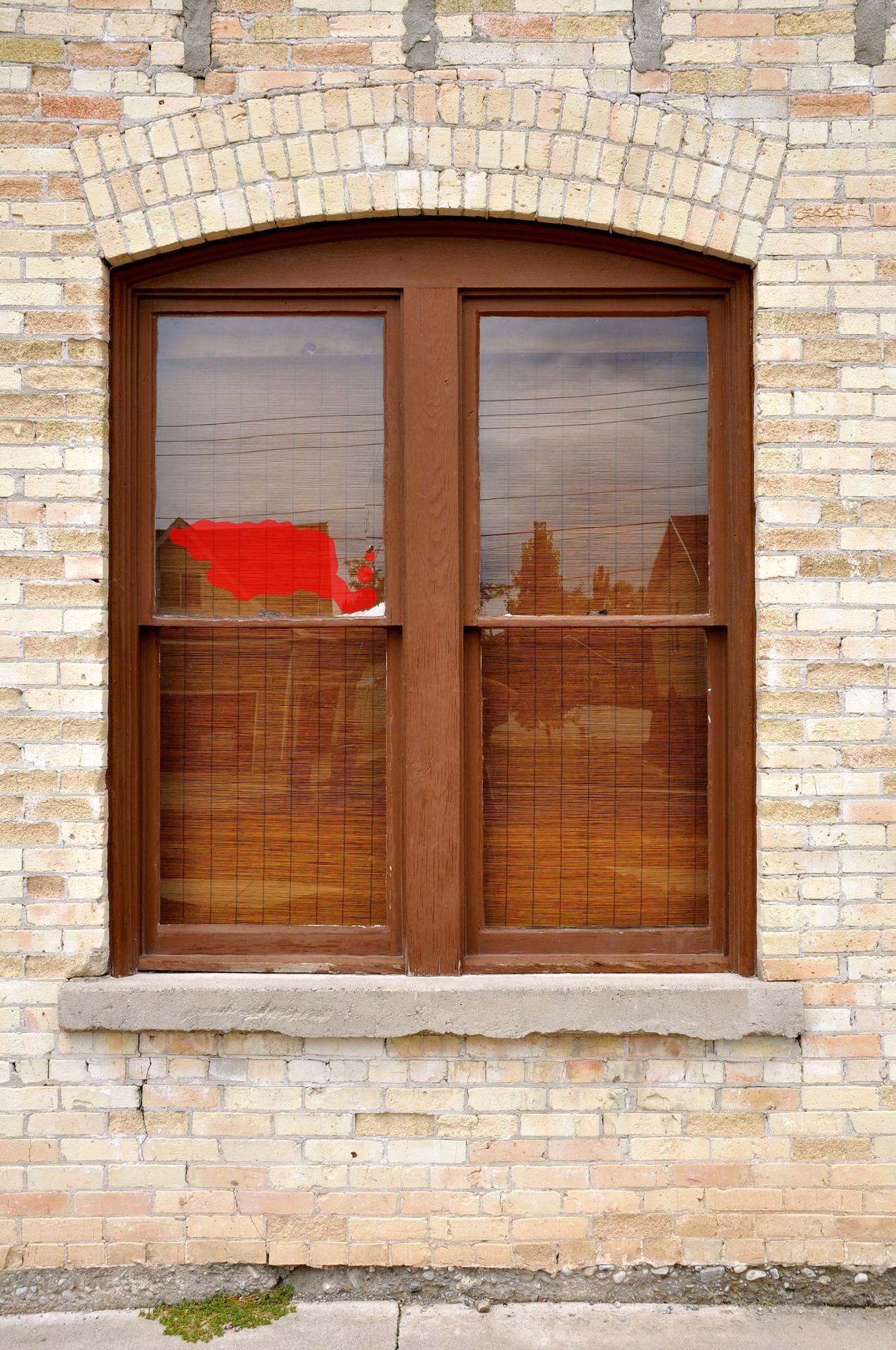} &
            \includegraphics[height=2.6cm]{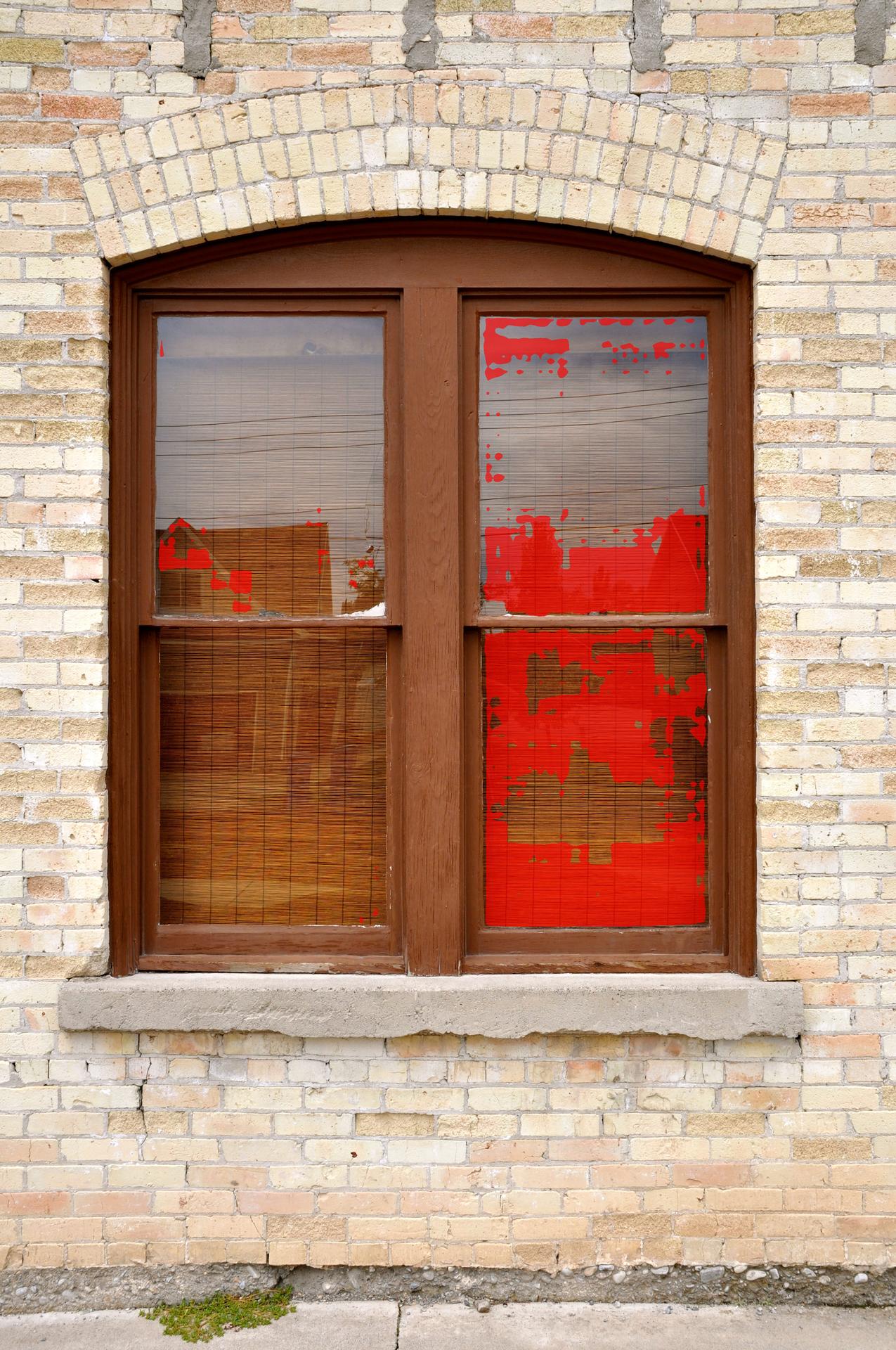} &
            \includegraphics[height=2.6cm]{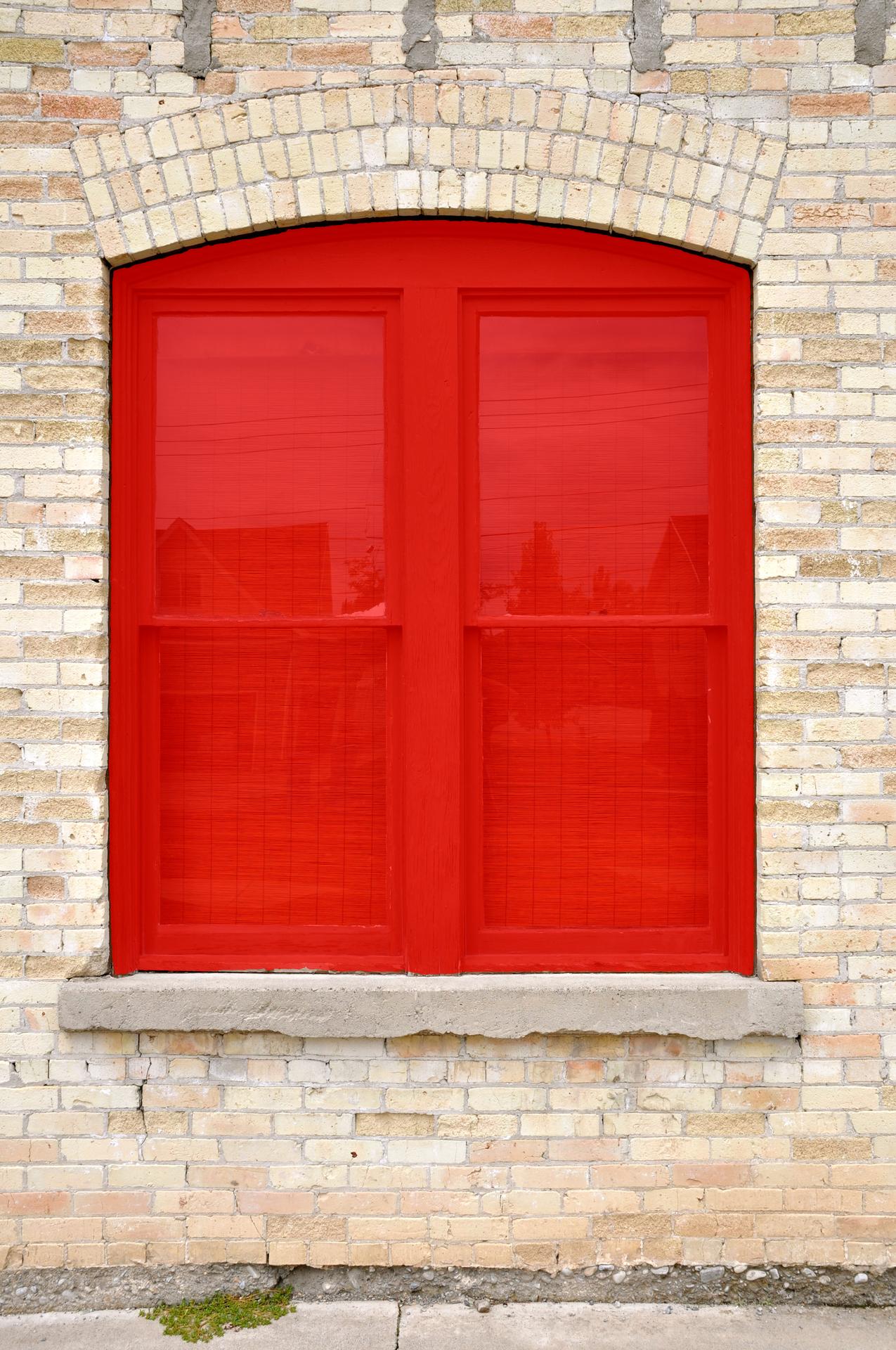} &
            \includegraphics[height=2.6cm]{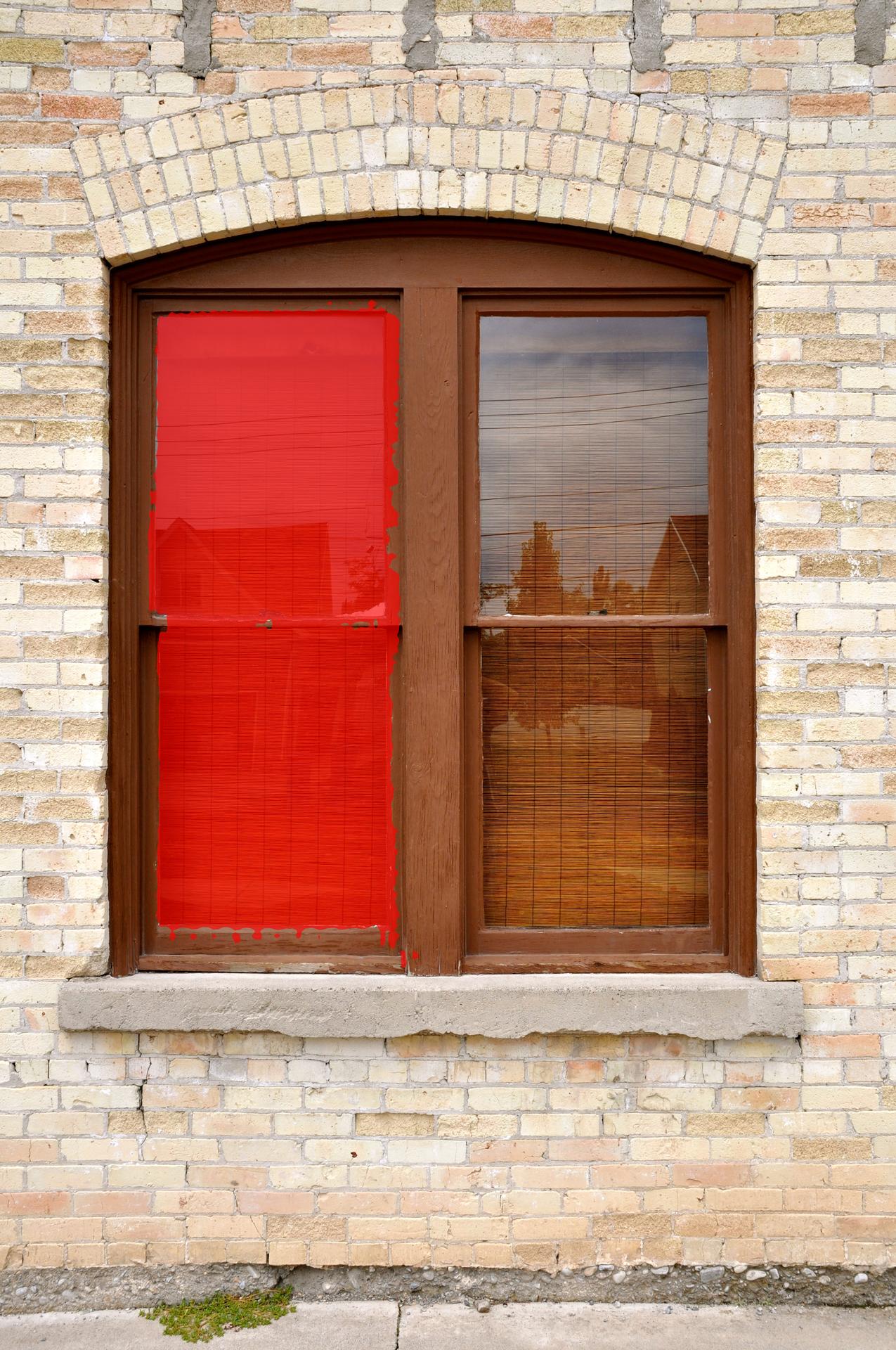} &
            \includegraphics[height=2.6cm]{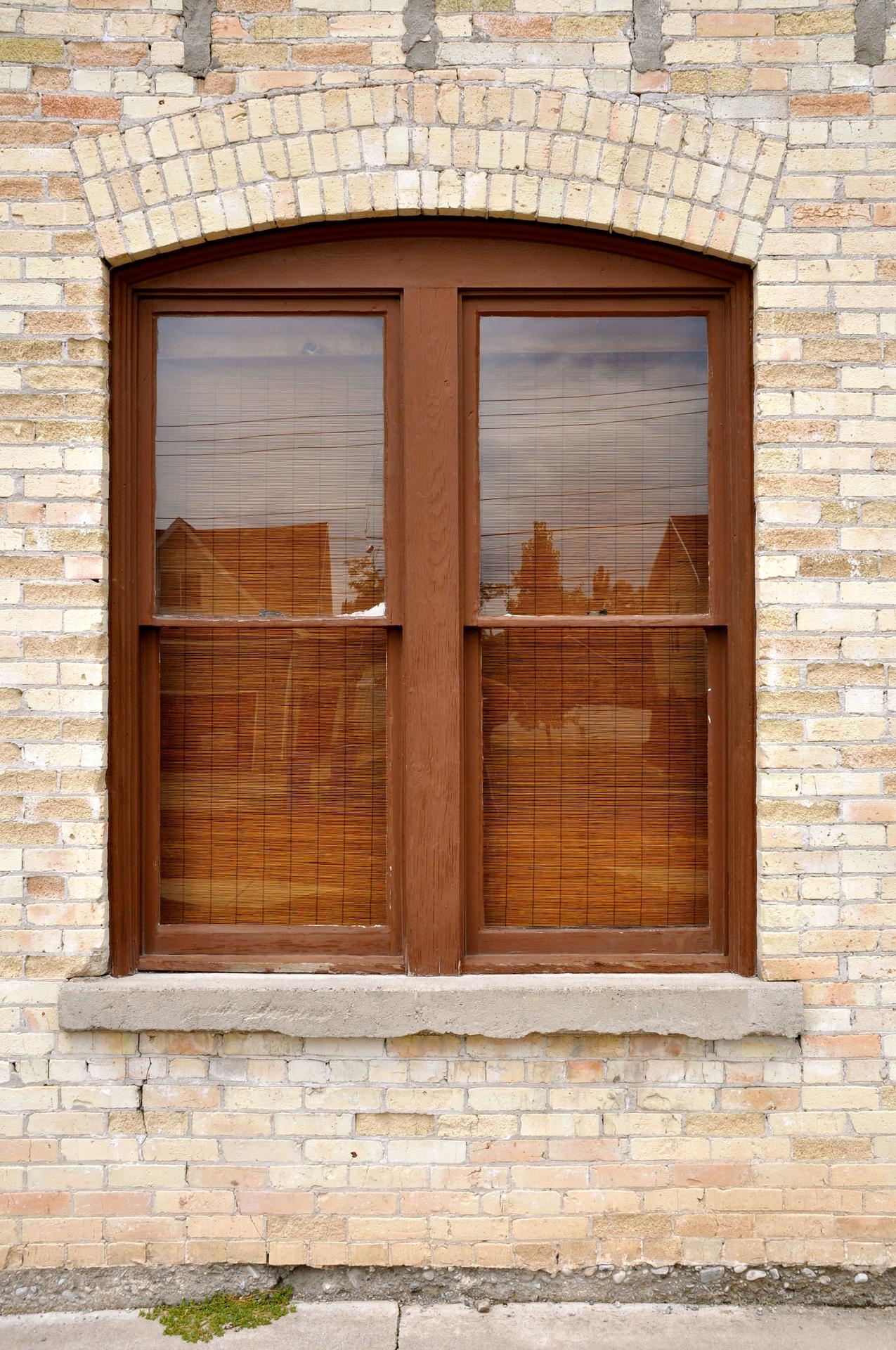} \\
        
            \raisebox{0.6cm}{\(\mathbf{I}', c\)} &
            \raisebox{0.6cm}{\(\boldsymbol{\varnothing}\)} &
            \includegraphics[height=2.6cm]{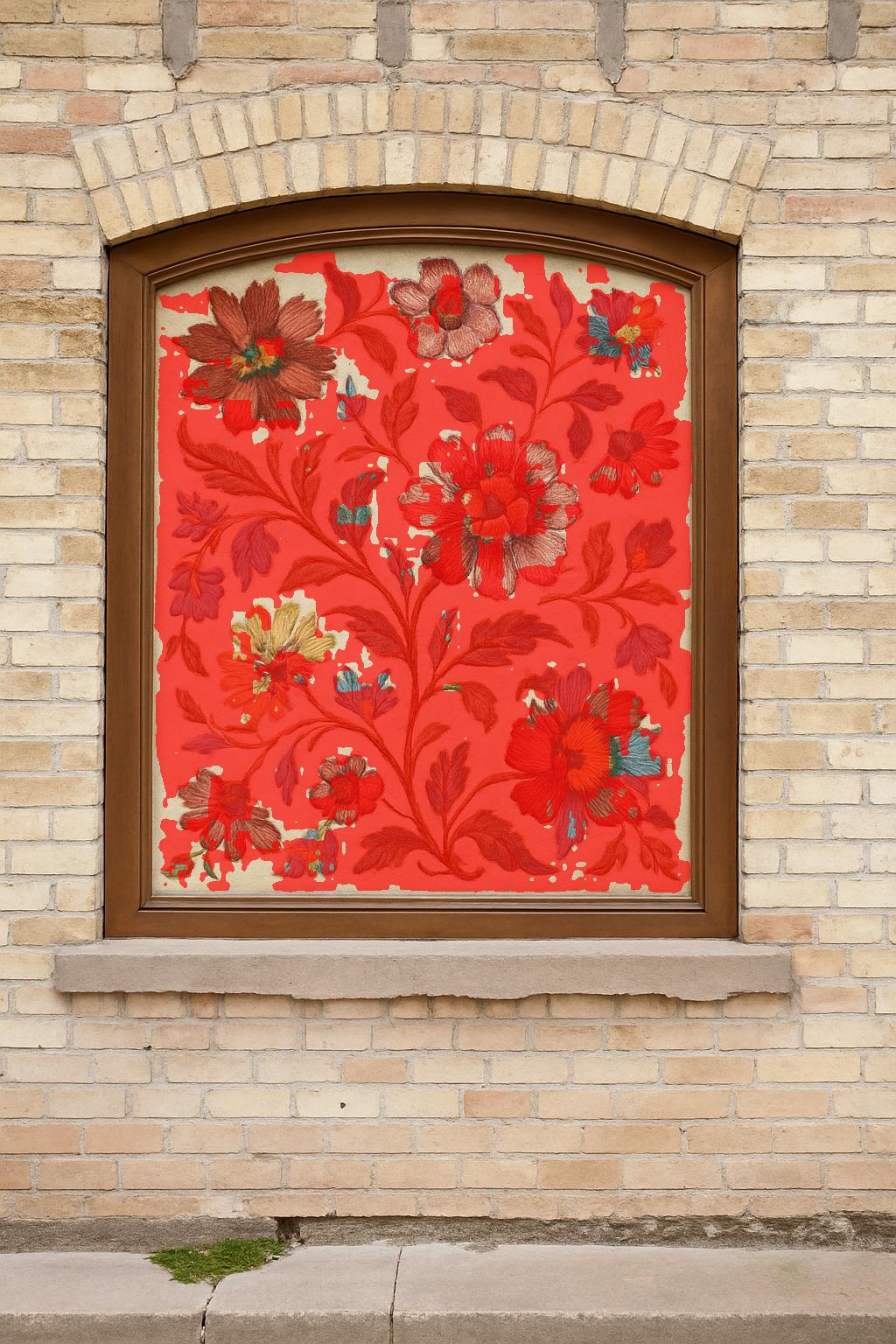} &
            \includegraphics[height=2.6cm]{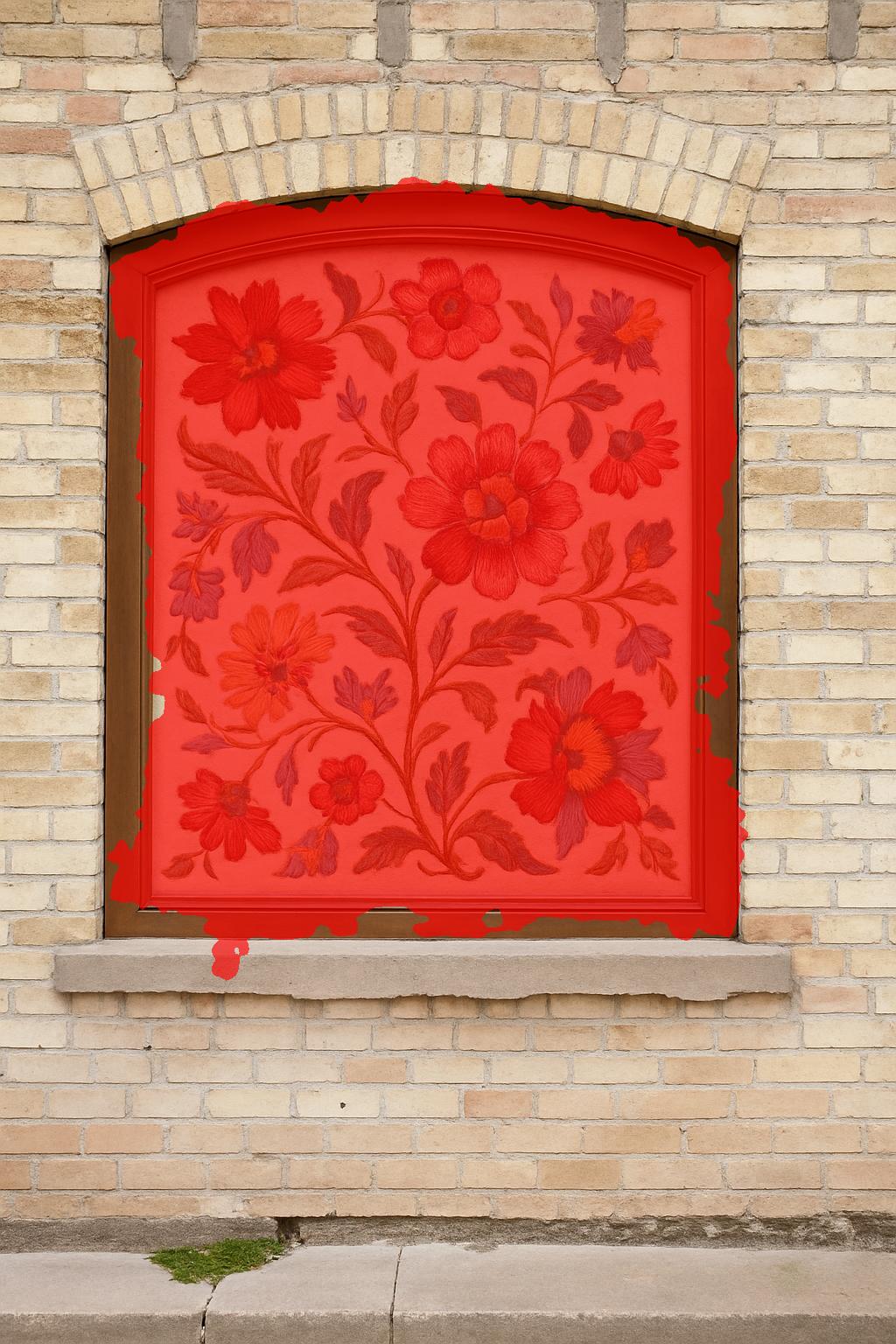} &
            \includegraphics[height=2.6cm]{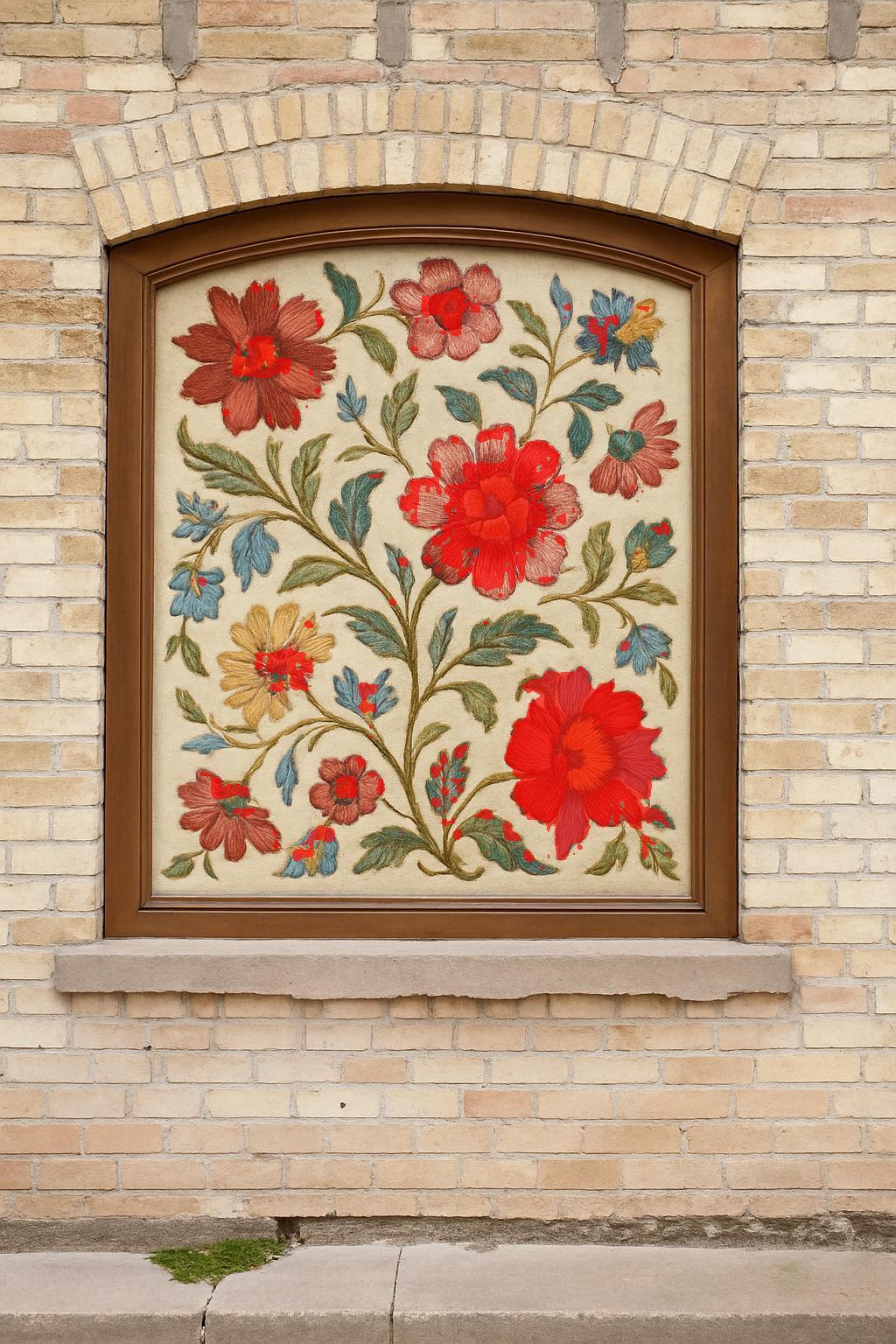} &
            \includegraphics[height=2.6cm]{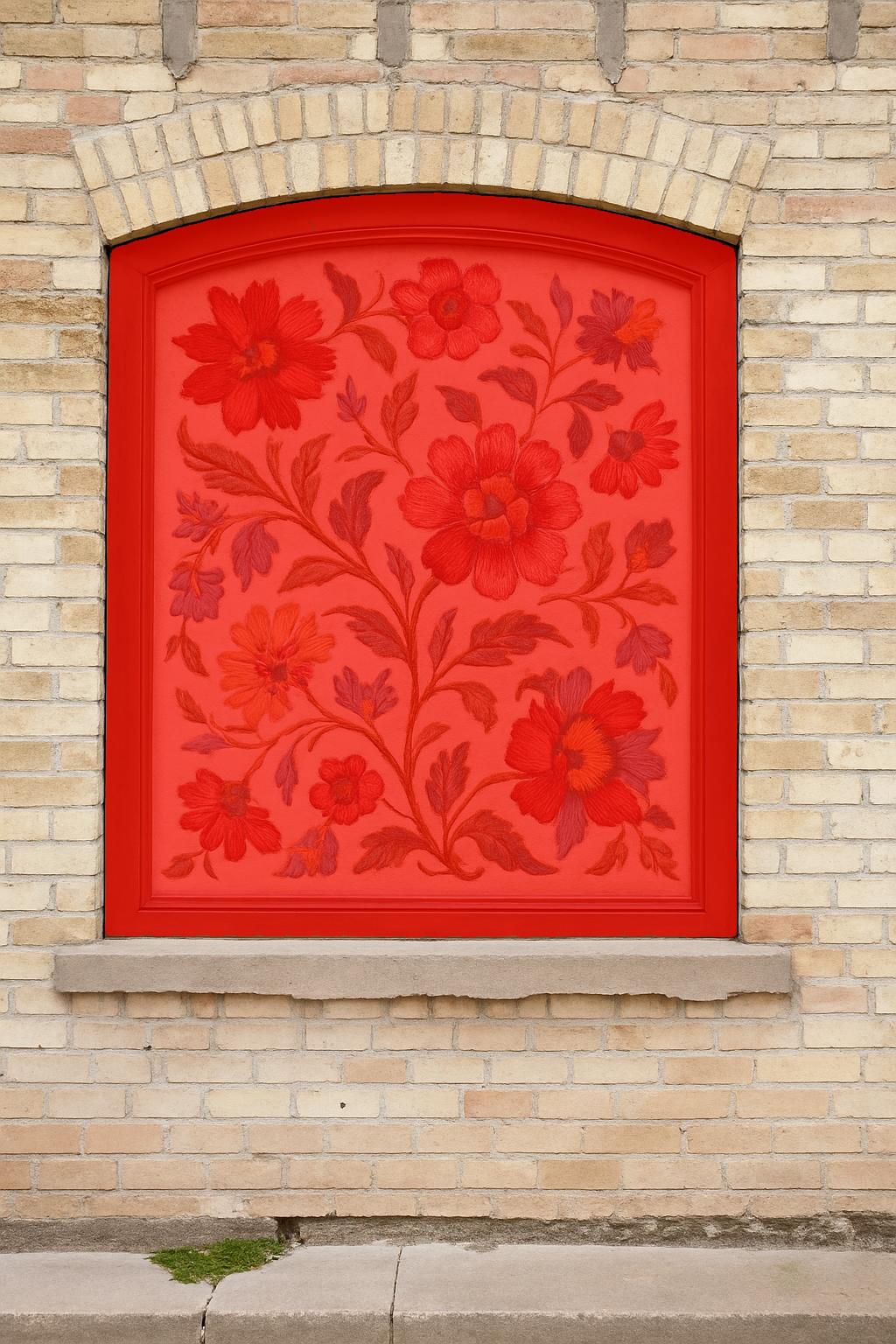} &
            \includegraphics[height=2.6cm]{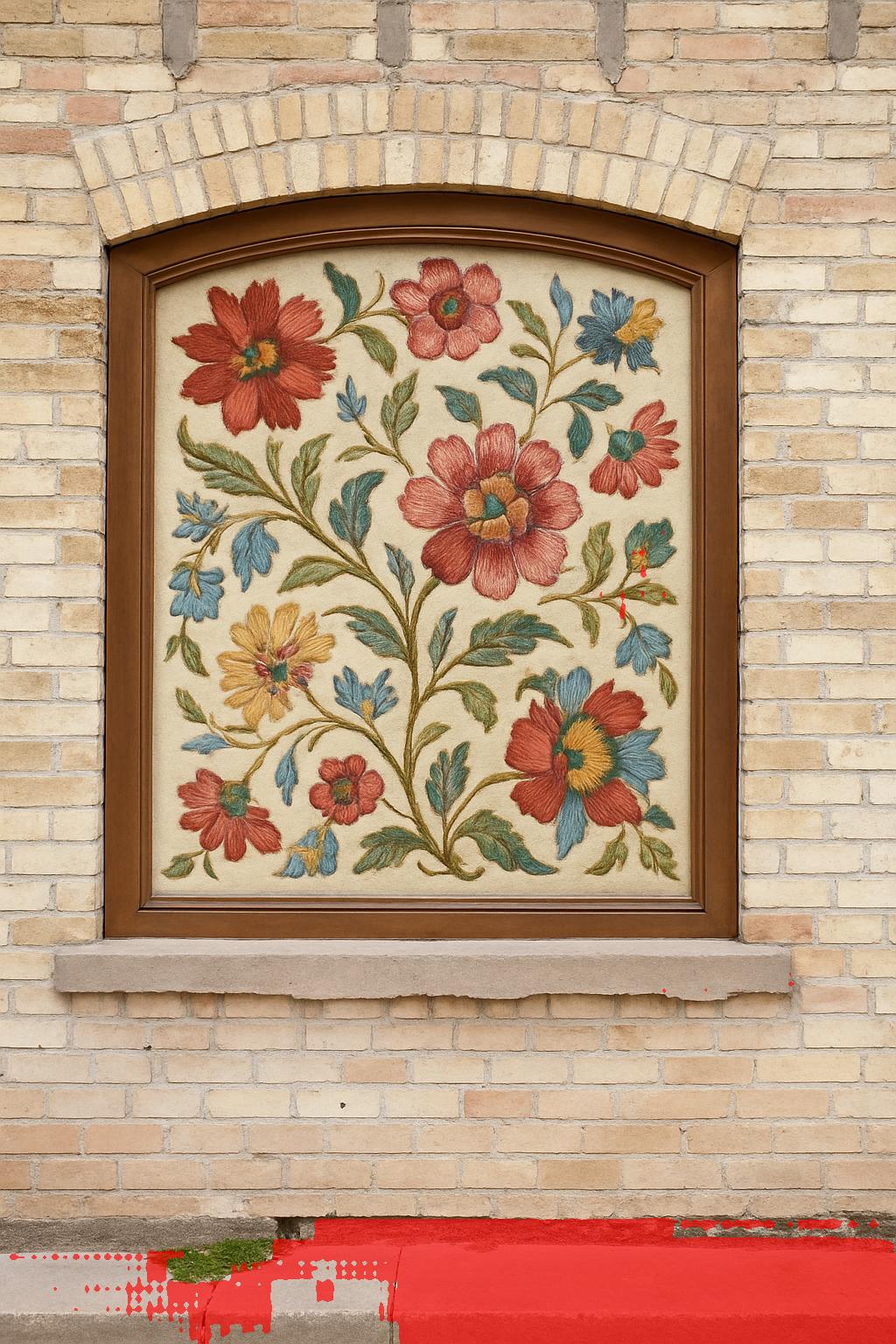} &
            \includegraphics[height=2.6cm]{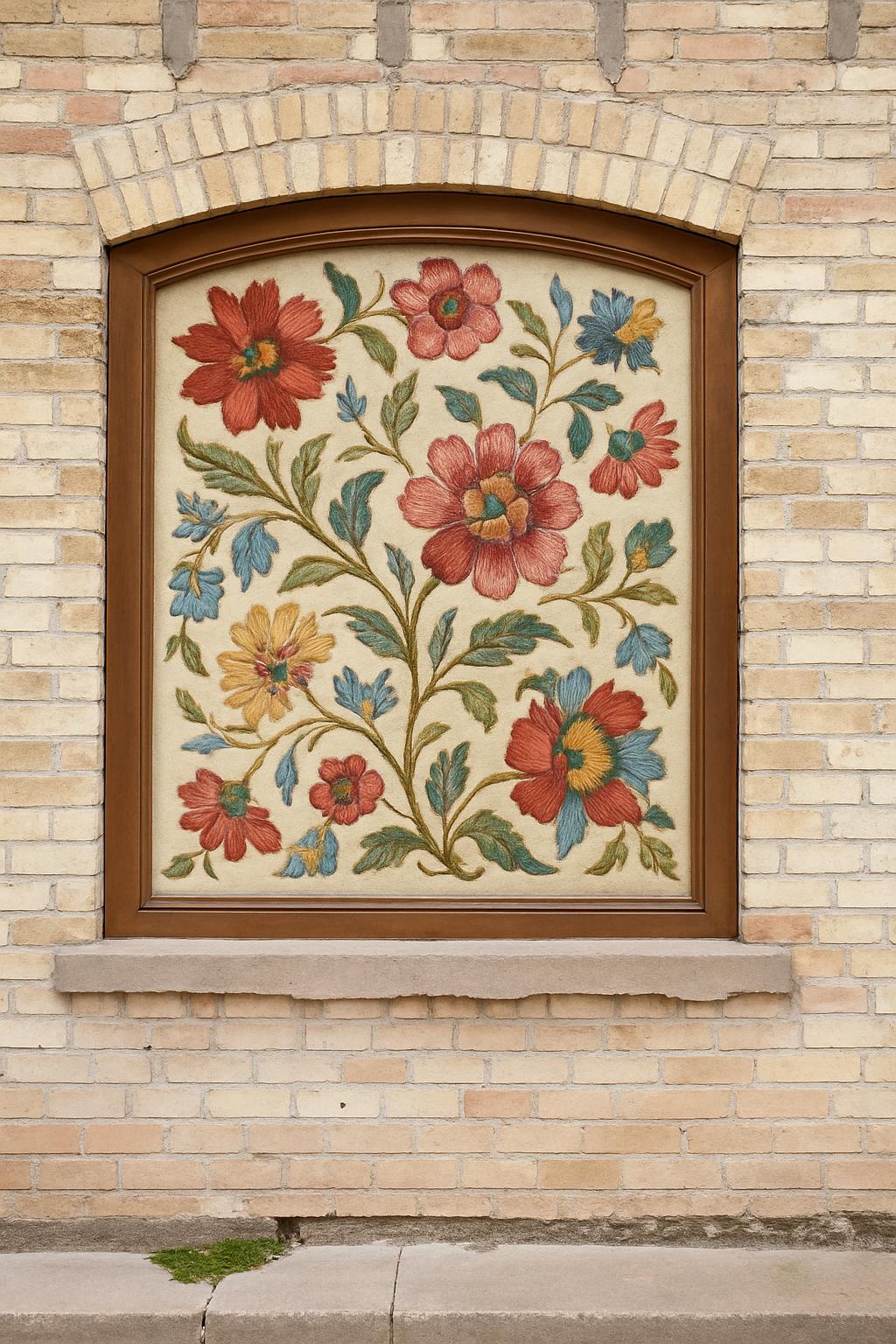} \\

            \raisebox{0.6cm}{\(\mathbf{I}', c'\)} &
            \includegraphics[height=2.6cm]{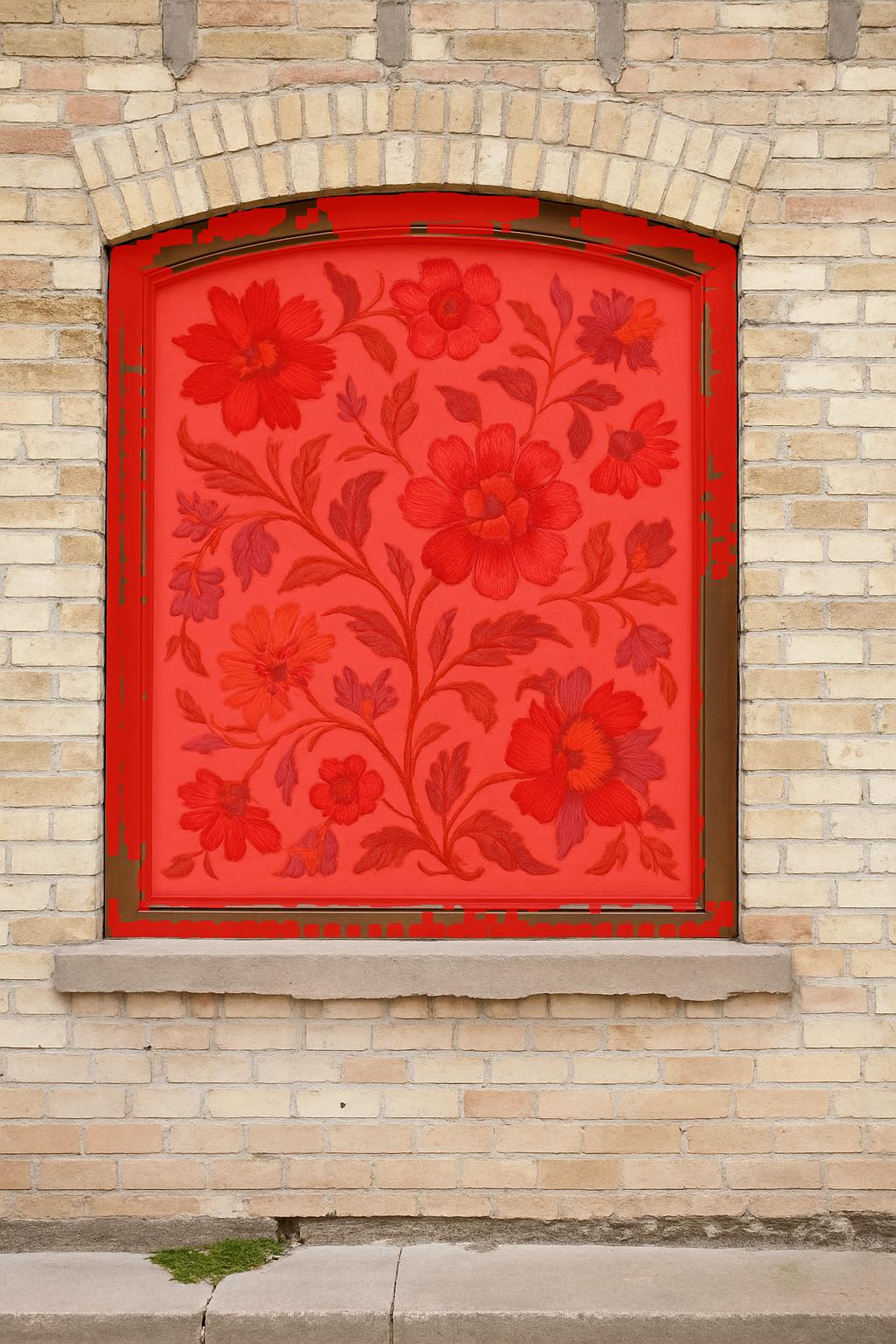} &
            \includegraphics[height=2.6cm]{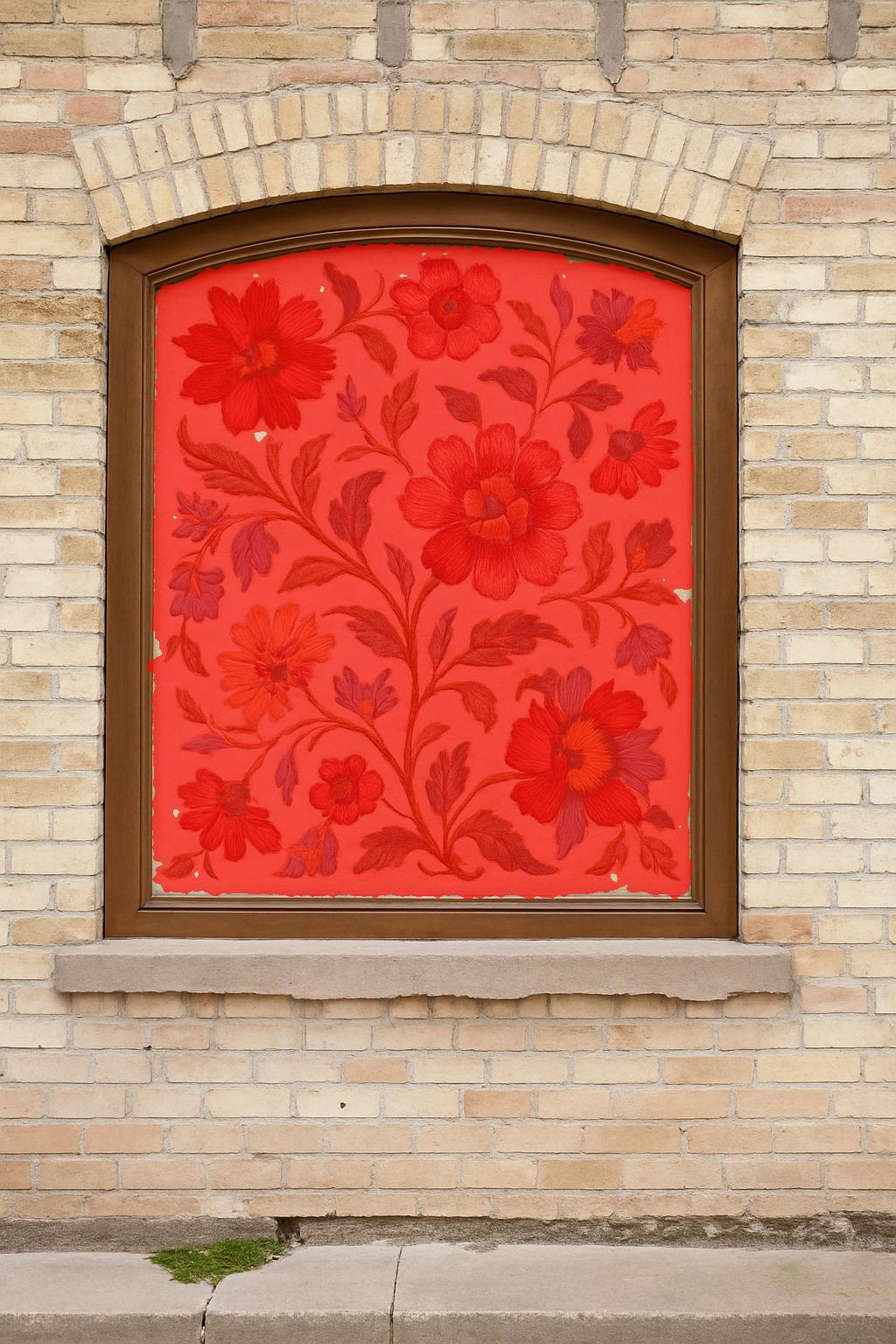} &
            \includegraphics[height=2.6cm]{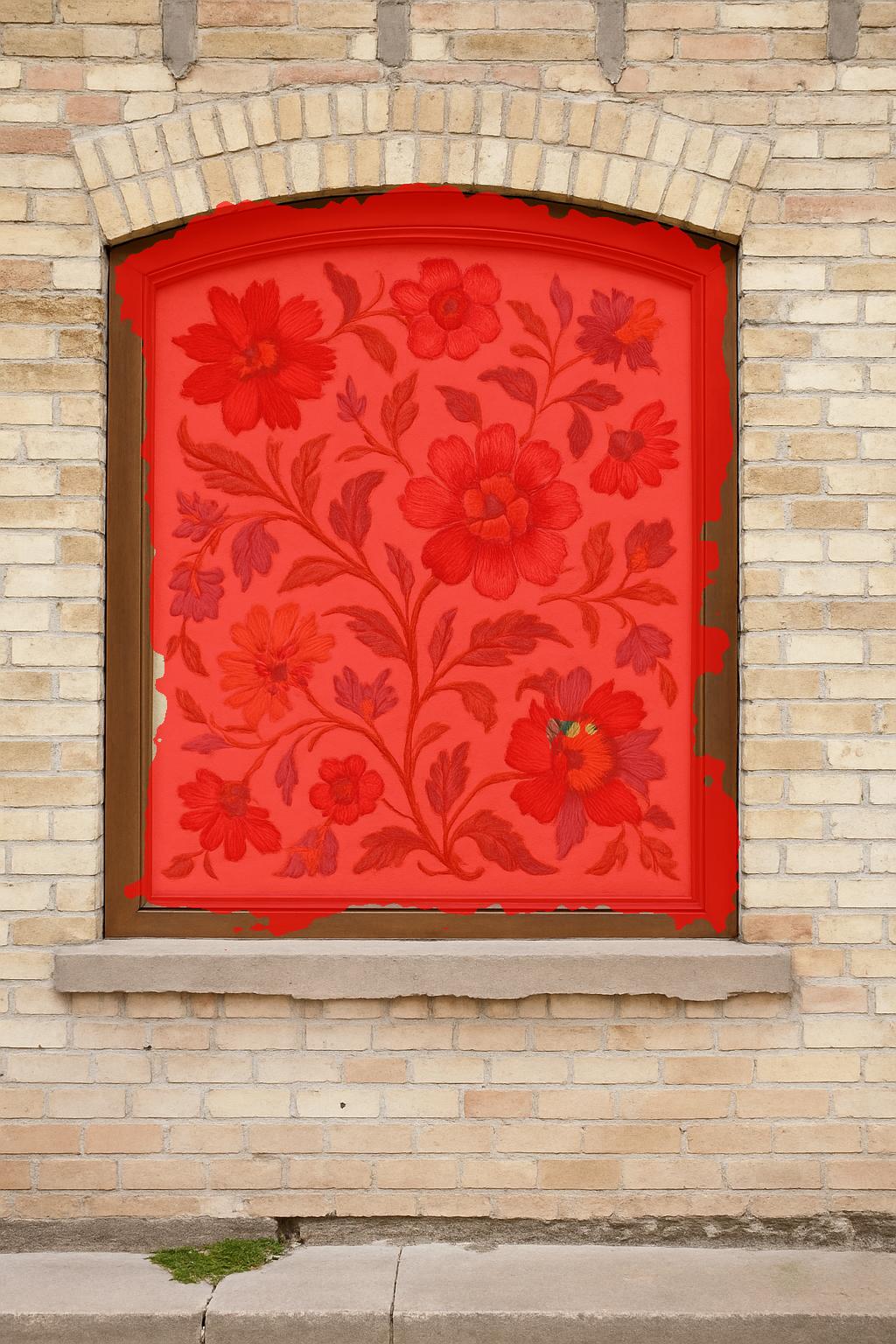} &
            \includegraphics[height=2.6cm]{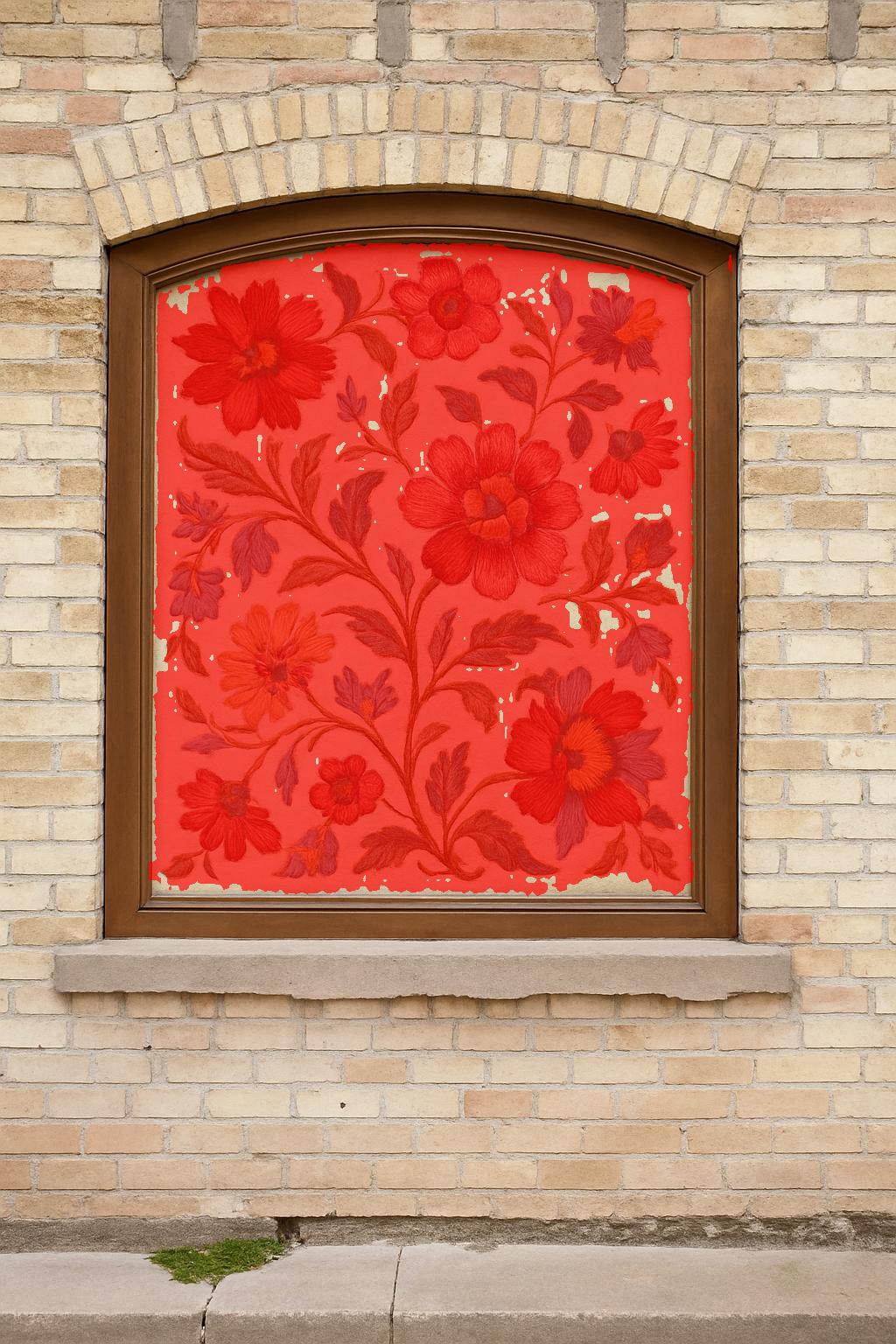} &
            \includegraphics[height=2.6cm]{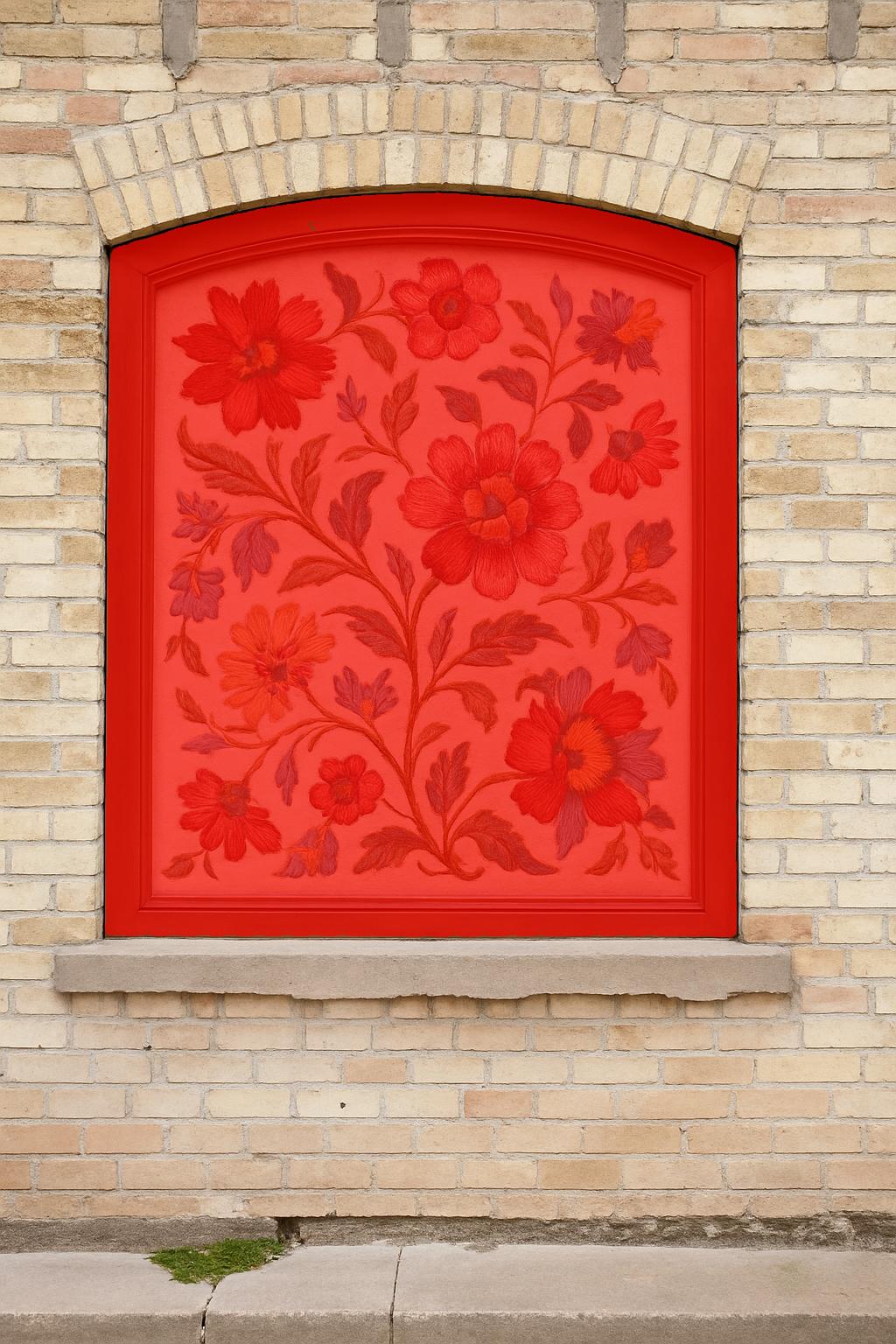} &
            \includegraphics[height=2.6cm]{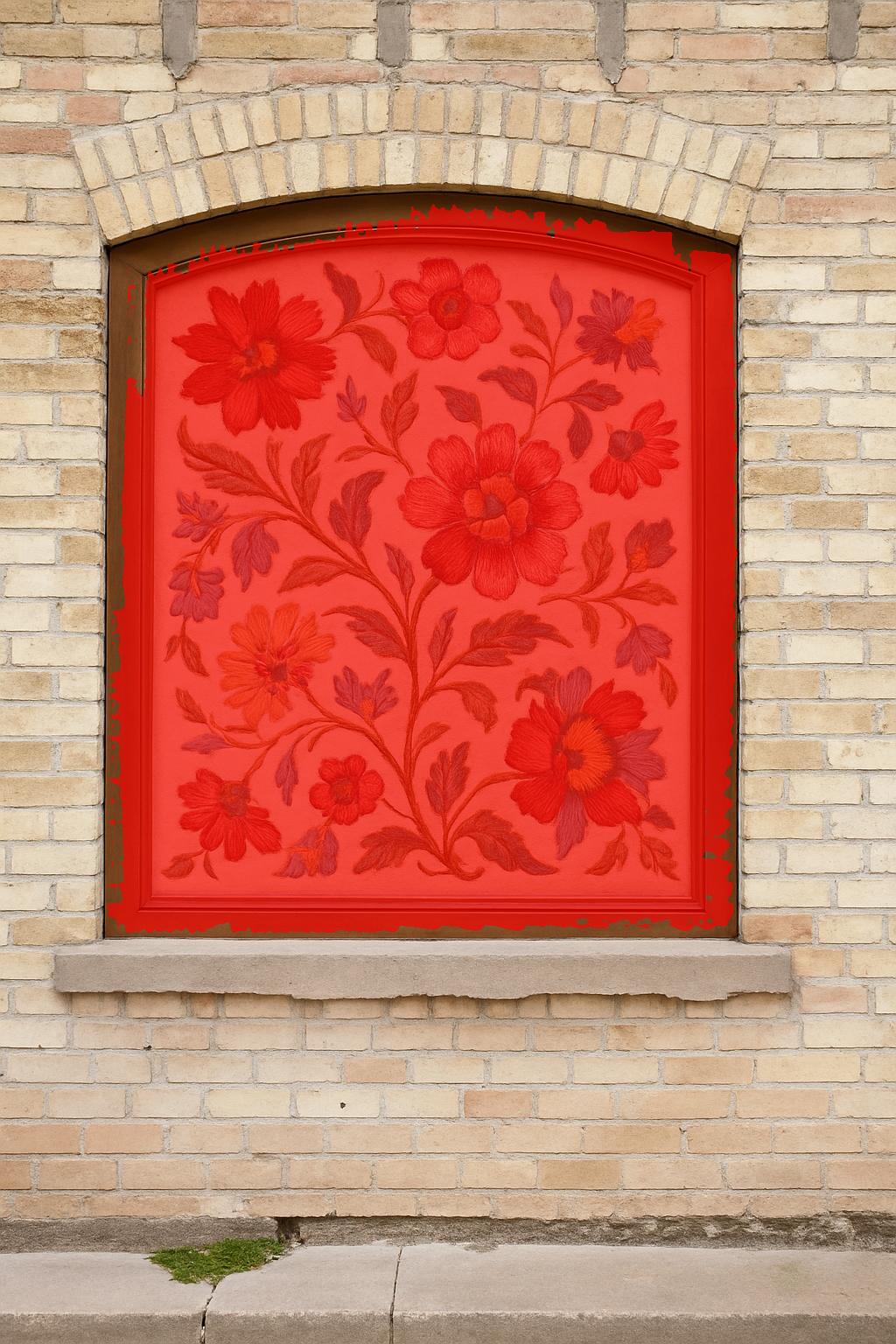} &
            \includegraphics[height=2.6cm]{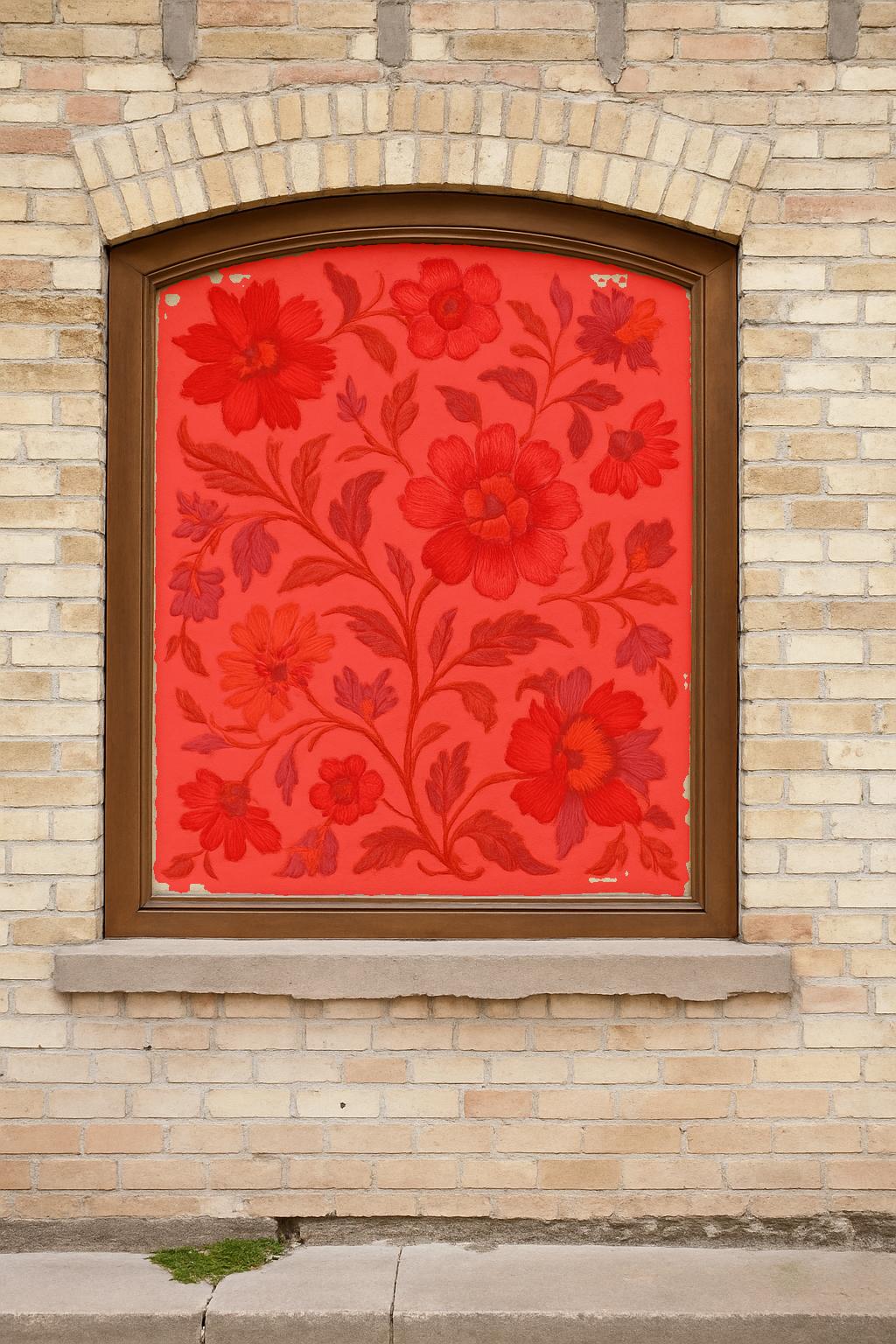} \\
        \end{tabular}}
        \caption*{\(c=\)``
        What specific part in the picture can provide us with this reflection?'' and \(c'=\)``
        What specific part in the picture can provide ornamental detail?''.}
    \end{minipage}
    \vspace{-0.3cm}
    \caption{\textbf{Qualitative Comparison of Reasoning Segmentation Models across Factual and Counterfactual Pairs.} The left panel presents results on \textbf{Referring} data, whereas the right panel presents results on \textbf{Reasoning} data. Here, \(c\) and \(c'\) denote the query prompts.}
    \label{fig:qualitative_app_2}
    \vspace{-0.1cm}
\end{figure*}

%% file: assets/tables/size_ablation.tex
\begin{table*}[t!]
    \centering
    \caption{\textbf{Comparison of Reasoning Segmentation Models on \benchmarkname{} Metrics: Referring segmentation} on small (S), medium (M), and large (L) object mask sizes. Arrows indicate (\(\uparrow\) higher is better, \(\downarrow\) lower is better, where better refers to fewer hallucinations). Best performance highlighted with \colorbox{LightBlue}{\phantom{\rule{1ex}{1ex}}}, second best performances underscored with \_.}
    \label{tab:size_ablation_ref}
    \vspace{-0.3cm}
    \resizebox{\textwidth}{!}{%
    \begin{tabular}{lcccccccccccccccc}
        \toprule
        \textbf{Model} 
        & \multicolumn{3}{c}{\(\dioutext \uparrow\)} 
        & \multicolumn{3}{c}{\(\diouvis \uparrow\)} 
        & \multicolumn{3}{c}{\(\cmsfact \downarrow\)} 
        & \multicolumn{3}{c}{\(\cmscf \downarrow\)} \\
        \cmidrule(lr){2-4} \cmidrule(lr){5-7} \cmidrule(lr){8-10} \cmidrule(lr){11-13}
        & S & M & L & S & M & L & S & M & L & S & M & L \\
        \midrule
        \textbf{LISA-7B}~\cite{lai2024lisa} & 0.3912 & \underline{0.4748} & 0.4646 & 0.2326 & 0.2788 & 0.3308 & 0.2869 & 0.3095 & 0.3271 & 0.9926 & 0.6819 & 0.6044 \\ 
        \textbf{PixelLM-7B}~\cite{ren2024pixellm} & 0.3825 & 0.4002 & 0.3833 & 0.4155 & 0.4025 & 0.4000 & 0.4214 & 0.4943 & 0.4916 & 1.0041 & 0.6693 & 0.6115 \\        
        \textbf{GLaMM-7B}~\cite{rasheed2024glamm} & 0.3124 & 0.3379 & 0.3177 & 0.2768 & 0.2915 & 0.3485 & 0.3857 & 0.4183 & 0.4567 & 0.8029 & 0.5724 & 0.4922 \\       
        \textbf{LISA-13B}~\cite{lai2024lisa} & 0.4079 & 0.4725 & \underline{0.4709} & 0.3574 & 0.3820 & 0.4304 & 0.2948 & 0.3311 & 0.3233 & 0.9262 & 0.6328 & 0.5093 \\
        \textbf{PixelLM-13B}~\cite{ren2024pixellm} & \underline{0.4099} & 0.4402 & 0.4170 & \colorbox{LightBlue}{0.4208} & 0.4247 & 0.4333 & 0.3615 & 0.4512 & 0.4520 & 1.0009 & 0.6708 & 0.5963 \\   
        \textbf{Seg-Zero}~\cite{liu2025seg} & 0.3687 & 0.4252 & 0.3687 & \underline{0.4144} & \colorbox{LightBlue}{0.5643} & \underline{0.6066} & 0.5233 & 0.4252 & 0.3687 & 0.6426 & 0.5637 & 0.4853 \\ 
        \textbf{VisionReasoner}~\cite{liu2025visionreasoner} & 0.3586 & 0.3862 & 0.3341 & 0.3890 & 0.5042 & 0.5571 & 0.9242 & 0.8314 & 0.6768 & 0.8900 & 0.7239 & 0.5265 \\ 
        \textbf{SESAME-7B}~\cite{wu2024see} & 0.3969 & 0.4239 & 0.4209 & 0.3358 & 0.3593 & 0.3922 & \underline{0.1964} & \underline{0.1969} & \underline{0.2130} & \underline{0.5532} & \colorbox{LightBlue}{0.4125} & \colorbox{LightBlue}{0.3573} \\
        \midrule
        \textbf{\modelnamegradient{} (Ours)} & \colorbox{LightBlue}{0.4971} & \colorbox{LightBlue}{0.6320} & \colorbox{LightBlue}{0.6610} & 0.4115 & \underline{0.5355} & \colorbox{LightBlue}{0.6080} & \colorbox{LightBlue}{0.1610} & \colorbox{LightBlue}{0.1145} & \colorbox{LightBlue}{0.1151} & \colorbox{LightBlue}{0.5416} & \underline{0.4630} & \underline{0.3692} \\ 
        \bottomrule
    \end{tabular}
    }
    \vspace{-0.2cm}
\end{table*}

%% file: assets/tables/size_ablation2.tex
\begin{table*}[t!]
    \centering
    \caption{\textbf{Comparison of Reasoning Segmentation Models on \benchmarkname{} Metrics: Reasoning segmentation} on small (S), medium (M), and large (L) object mask sizes. Arrows indicate (\(\uparrow\) higher is better, \(\downarrow\) lower is better, where better refers to fewer hallucinations). Best performance highlighted with \colorbox{LightBlue}{\phantom{\rule{1ex}{1ex}}}, second best performances underscored with \_.} 
    \label{tab:size_ablation_reason}
        \vspace{-0.3cm}
    \resizebox{\textwidth}{!}{%
    \begin{tabular}{lcccccccccccccccc}
        \toprule
        \textbf{Model} 
        & \multicolumn{3}{c}{\(\dioutext \uparrow\)} 
        & \multicolumn{3}{c}{\(\diouvis \uparrow\)} 
        & \multicolumn{3}{c}{\(\cmsfact \downarrow\)} 
        & \multicolumn{3}{c}{\(\cmscf \downarrow\)} \\
        \cmidrule(lr){2-4} \cmidrule(lr){5-7} \cmidrule(lr){8-10} \cmidrule(lr){11-13}
        & S & M & L & S & M & L & S & M & L & S & M & L \\
        \midrule      
        \textbf{LISA-7B}~\cite{lai2024lisa} & 0.2840 & {0.2809} & 0.3326 & 0.2810 & 0.2336 & 0.2242 & 0.5204 & 0.4331 & 0.4026 & 0.7082 & 0.6843 & 0.7019 \\ 
        \textbf{PixelLM-7B}~\cite{ren2024pixellm} & 0.2741 & 0.2639 & 0.3023 & 0.1989 & 0.2458 & 0.2213 & 0.4158 & 0.4978 & 0.3499 & 0.5463 & 0.6356 & 0.5412 \\        
        \textbf{GLaMM-7B}~\cite{rasheed2024glamm} & 0.2793 & 0.2391 & 0.2265 & \underline{0.3331} & \underline{0.3315} & 0.2856 & 0.4261 & 0.3633 & 0.2956 & \colorbox{LightBlue}{0.3954} & \underline{0.3999} & \underline{0.4099} \\       
        \textbf{LISA-13B}~\cite{lai2024lisa} & \underline{0.3814} & 0.2713 & \underline{0.4061} & \colorbox{LightBlue}{0.3564} & 0.2860 & 0.3260 & 0.5213 & 0.4940 & 0.4860 & 0.6710 & 0.6537 & 0.6741 \\
        \textbf{PixelLM-13B}~\cite{ren2024pixellm} & {0.3360} & 0.2909 & 0.3172 & {0.3149} & {0.3178} & {0.2904} & 0.4334 & 0.4690 & 0.3685 & \underline{0.5053} & 0.5448 & 0.5662 \\   
        \textbf{Seg-Zero}~\cite{liu2025seg} & 0.2887 & 0.2248 & 0.2218 & 0.2587 & {0.3253} & 0.3102 & 0.5948 & 0.6675 & 0.6836 & 0.7462 & 0.8185 & 0.7172 \\ 
        \textbf{VisionReasoner}~\cite{liu2025visionreasoner} & 0.2445 & 0.2437 & 0.2530 & 0.2896 & \colorbox{LightBlue}{0.3430} & 0.2789 & 0.7747 & 0.7481 & 0.6751 & 0.8398 & 0.9086 & 0.8125 \\ 
        \textbf{SESAME-7B}~\cite{wu2024see} & 0.3478 & \underline{0.3211} & 0.3356 & 0.2571 & 0.2999 & \underline{0.3307} & \underline{0.2954} & \underline{0.3062} & \underline{0.2949} & {0.5180} & {0.5237} & {0.4200} \\
        \midrule
        \textbf{\modelnamegradient{} (Ours)} & \colorbox{LightBlue}{0.4355} & \colorbox{LightBlue}{0.3828} & \colorbox{LightBlue}{0.4216} & 0.3171 & 0.3042 & \colorbox{LightBlue}{0.3370} & \colorbox{LightBlue}{0.2299} & \colorbox{LightBlue}{0.1410} & \colorbox{LightBlue}{0.1163} & 0.5151 & \colorbox{LightBlue}{0.3924} & \colorbox{LightBlue}{0.3235} \\ 
        \bottomrule
    \end{tabular}
    }
    \vspace{-0.2cm}
\end{table*}

%% file: assets/figures/loss_abl_iou.tex
\begin{figure}[t]
    \centering
    \includegraphics[width=\columnwidth]{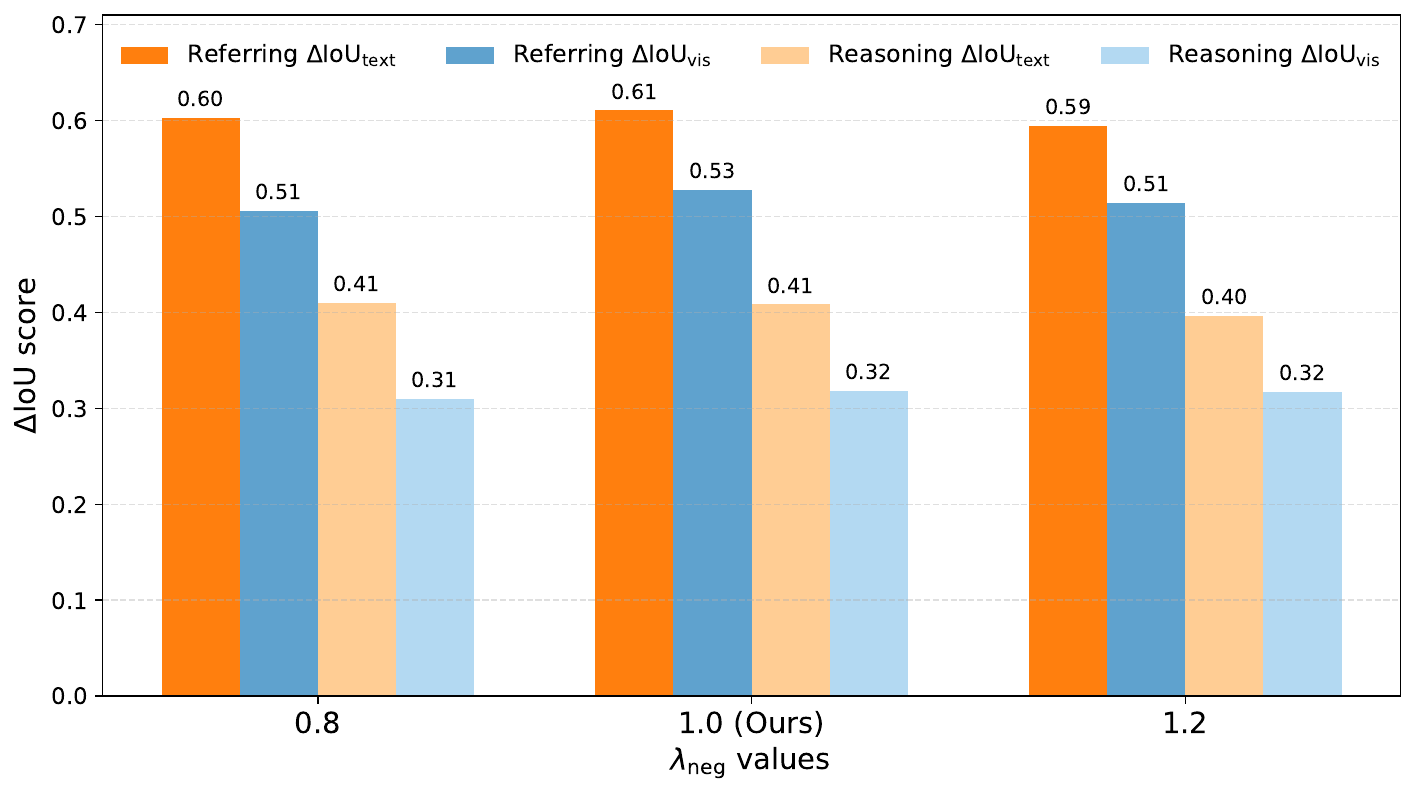}
    \vspace{-0.8cm}
    \caption{
    \textbf{Ablation over $\lambda_{\text{neg}}$} on $\Delta$IoU for textual (\textit{text}) and visual (\textit{vis}) on Referral and Reasoning. 
    Here, higher values \(\diou\) reflect better post-hallucination mitigation consistency.
    }
    \label{fig:delta_iou_lambda_ablation}
    \vspace{-0.3cm}
\end{figure}

%% file: assets/figures/loss_abl_cms.tex
\begin{figure}[t]
    \centering
    \includegraphics[width=0.99\columnwidth]{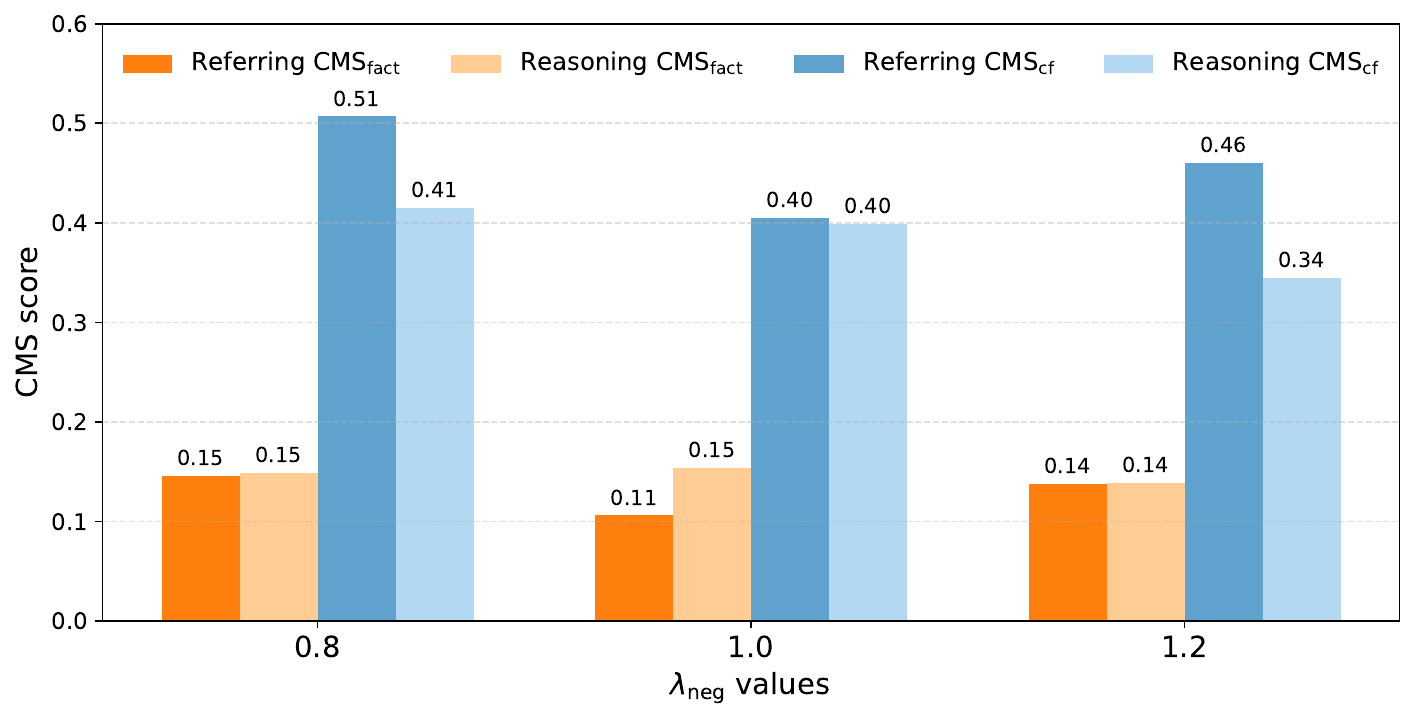}
    \vspace{-0.4cm}
    \caption{
    \textbf{Ablation over $\lambda_{\text{neg}}$} on (\(\cms\)) for both Referral and Reasoning settings. 
    We report factual (\textit{fact}) and counterfactual (\textit{cf}) \(\cms\) for three choices of \(\lambda_{\text{neg}}\). 
    Lower \(\cms\) indicates better suppression of hallucinated regions.
    }
    \label{fig:cms_lambda_ablation}
    \vspace{-0.4cm}
\end{figure}

%% file: main.bib
@String(CVPR= {IEEE Conf. Comput. Vis. Pattern Recog.})

@String(ICCV= {Int. Conf. Comput. Vis.})

@String(ECCV= {Eur. Conf. Comput. Vis.})

@String(NIPS= {Adv. Neural Inform. Process. Syst.})

@String(TOG= {ACM Trans. Graph.})

@String(TIP  = {IEEE Trans. Image Process.})

@String(ICLR = {Int. Conf. Learn. Represent.})

@String(CVPR  = {CVPR})

@String(ICCV  = {ICCV})

@String(ECCV  = {ECCV})

@String(NIPS  = {NeurIPS})

@String(TOG   = {ACM TOG})

@String(TIP   = {IEEE TIP})

@String(ICLR  = {ICLR})

@String(ICML  = {International Conference on Machine Learning (ICML)})

@String(EMNLP = {Conference on Empirical Methods in Natural Language Processing (EMNLP)})

@String(ACL = {Annual Meeting of the Association for Computational Linguistics (ACL)})

@inproceedings{boerner2023access,
    title={{ACCESS: Advancing Innovation: NSF's Advanced Cyberinfrastructure Coordination Ecosystem: Services \& Support}},
    author={Boerner, Timothy J and Deems, Stephen and Furlani, Thomas R and Knuth, Shelley L and Towns, John},
    booktitle={2023 Practice and Experience in Advanced Research Computing, PEARC 2023},
    year={2023},
    organization={Association for Computing Machinery}
}

@inproceedings{li2023evaluating,
    title={{Evaluating Object Hallucination in Large Vision-Language Models}},
    author={Li, Yifan and Du, Yifan and Zhou, Kun and Wang, Jinpeng and Zhao, Wayne Xin and Wen, Ji-Rong},
    booktitle=EMNLP,
    year={2023}
}

@inproceedings{lovenia2024negative,
    title={{Negative Object Presence Evaluation (NOPE) to Measure Object Hallucination in Vision-Language Models}},
    author={Lovenia, Holy and Dai, Wenliang and Cahyawijaya, Samuel and Ji, Ziwei and Fung, Pascale},
    booktitle={Proceedings of the 3rd Workshop on Advances in Language and Vision Research (ALVR)},
    year={2024}
}

@inproceedings{rohrbach2018object,
    title={{Object Hallucination in Image Captioning}},
    author={Rohrbach, Anna and Hendricks, Lisa Anne and Burns, Kaylee and Darrell, Trevor and Saenko, Kate},
    booktitle=EMNLP,
    year={2018}
}

@inproceedings{nguyen2025calico,
  title={{CALICO: Part-Focused Semantic Co-Segmentation with Large Vision-Language Models}},
  author={Nguyen, Kiet A and Juvekar, Adheesh and Yu, Tianjiao and Wahed, Muntasir and Lourentzou, Ismini},
  booktitle=CVPR,
  year={2025}
}

@inproceedings{liu2025phd,
  title={{PhD: A ChatGPT-Prompted Visual hallucination Evaluation Dataset}},
  author={Liu, Jiazhen and Fu, Yuhan and Xie, Ruobing and Xie, Runquan and Sun, Xingwu and Lian, Fengzong and Kang, Zhanhui and Li, Xirong},
  booktitle=CVPR,
  year={2025}
}

@inproceedings{wu2024see,
    title={{See, Say, and Segment: Teaching LMMs to Overcome False Premises}},
    author={Wu, Tsung-Han and Biamby, Giscard and Chan, David and Dunlap, Lisa and Gupta, Ritwik and Wang, Xudong and Gonzalez, Joseph E and Darrell, Trevor},
    booktitle=CVPR,
    year={2024}
}

@article{wu2024towards,
    title={{Towards Robust Referring Image Segmentation}},
    author={Wu, Jianzong and Li, Xiangtai and Li, Xia and Ding, Henghui and Tong, Yunhai and Tao, Dacheng},
    journal=TIP,
    year={2024},
}

@inproceedings{li2023blip,
    title={{BLIP-2: Bootstrapping Language-Image Pre-training with Frozen Image Encoders and Large Language Models}},
    author={Li, Junnan and Li, Dongxu and Savarese, Silvio and Hoi, Steven},
    booktitle=ICML,
    year={2023}
}

@inproceedings{alayrac2022flamingo,
    title={{Flamingo: a Visual Language Model for Few-Shot Learning}},
    author={{Alayrac, Jean-Baptiste and Donahue, Jeff and Luc, Pauline and Miech, Antoine and Barr, Iain and Hasson, Yana and Lenc, Karel and Mensch, Arthur and Millican, Katherine and Reynolds, Malcolm and others}},
    booktitle=NIPS,
    year={2022}
}

@inproceedings{sinha2025guiding,
    title={{Guiding Vision-Language Model Selection for Visual Question-Answering Across Tasks, Domains, and Knowledge Types}},
    author={Sinha, Neelabh and Jain, Vinija and Chadha, Aman},
    booktitle={Proceedings of the First Workshop of Evaluation of Multi-Modal Generation},
    year={2025}
}

@inproceedings{liu2023gres,
    title={{GRES: Generalized Referring Expression Segmentation}},
    author={Liu, Chang and Ding, Henghui and Jiang, Xudong},
    booktitle=CVPR,
    year={2023}
}

@inproceedings{geigle2024does,
    title={{Does Object Grounding Really Reduce Hallucination of Large Vision-Language Models?}},
    author={Geigle, Gregor and Timofte, Radu and Glava{\v{s}}, Goran},
    booktitle=EMNLP,
    year={2024}
}

@inproceedings{ma2024groma,
    title={{Groma: Localized Visual Tokenization for Grounding Multimodal Large Language Models}},
    author={Ma, Chuofan and Jiang, Yi and Wu, Jiannan and Yuan, Zehuan and Qi, Xiaojuan},
    booktitle=ECCV,
    year={2024},
}

@inproceedings{wang2022ofa,
    title={{OFA: Unifying Architectures, Tasks, and Modalities Through a Simple Sequence-to-Sequence Learning Framework}},
    author={Wang, Peng and Yang, An and Men, Rui and Lin, Junyang and Bai, Shuai and Li, Zhikang and Ma, Jianxin and Zhou, Chang and Zhou, Jingren and Yang, Hongxia},
    booktitle=ICML,
    year={2022}
}

@inproceedings{chen2020uniter,
    title={{UNITER: UNiversal Image-TExt Representation Learning}},
    author={Chen, Yen-Chun and Li, Linjie and Yu, Licheng and El Kholy, Ahmed and Ahmed, Faisal and Gan, Zhe and Cheng, Yu and Liu, Jingjing},
    booktitle=ECCV,
    year={2020},
}

@inproceedings{chen2024spatialvlm,
    title={{SpatialVLM: Endowing Vision-Language Models with Spatial Reasoning Capabilities}},
    author={{Chen, Boyuan and Xu, Zhuo and Kirmani, Sean and Ichter, Brian and Sadigh, Dorsa and Guibas, Leonidas and Xia, Fei}},
    booktitle=CVPR,
    year={2024}
}

@inproceedings{liu2024visual,
    title={{Visual Instruction Tuning}},
    author={{Liu, Haotian and Li, Chunyuan and Wu, Qingyang and Lee, Yong Jae}},
    booktitle=NIPS,
    year={2024}
}

@inproceedings{huang2024visual,
    title={{Visual Hallucinations of Multi-modal Large Language Models}},
    author={Huang, Wen and Liu, Hongbin and Guo, Minxin and Gong, Neil},
    booktitle=ACL,
    year={2024}
}

@article{zhai2023halle,
    title={{HallE-Control: Controlling Object Hallucination in Large Multimodal Models}},
    author={Zhai, Bohan and Yang, Shijia and Xu, Chenfeng and Shen, Sheng and Keutzer, Kurt and Li, Chunyuan and Li, Manling},
    journal={arXiv preprint arXiv:2310.01779},
    year={2023}
}

@inproceedings{wang2022counterfactual,
    title={{Counterfactual Cycle-Consistent Learning for Instruction Following and Generation in Vision-Language Navigation}},
    author={Wang, Hanqing and Liang, Wei and Shen, Jianbing and Van Gool, Luc and Wang, Wenguan},
    booktitle=CVPR,
    year={2022}
}

@inproceedings{parvaneh2020counterfactual,
    title={{Counterfactual Vision-and-Language Navigation: Unravelling the Unseen}},
    author={Parvaneh, Amin and Abbasnejad, Ehsan and Teney, Damien and Shi, Javen Qinfeng and Van den Hengel, Anton},
    booktitle=NIPS,
    year={2020}
}

@inproceedings{liang2020learning,
    title={{Learning to Contrast the Counterfactual Samples for Robust Visual Question Answering}},
    author={Liang, Zujie and Jiang, Weitao and Hu, Haifeng and Zhu, Jiaying},
    booktitle=EMNLP,
    year={2020}
}

@inproceedings{niu2021counterfactual,
    title={{Counterfactual VQA: A Cause-Effect Look at Language Bias}},
    author={Niu, Yulei and Tang, Kaihua and Zhang, Hanwang and Lu, Zhiwu and Hua, Xian-Sheng and Wen, Ji-Rong},
    booktitle=CVPR,
    year={2021}
}

@inproceedings{lai2024lisa,
    title={{LISA: Reasoning Segmentation via Large Language Model}},
    author={Lai, Xin and Tian, Zhuotao and Chen, Yukang and Li, Yanwei and Yuan, Yuhui and Liu, Shu and Jia, Jiaya},
    booktitle=CVPR,
    year={2024}
}

@inproceedings{zhu2017unpaired,
    title={{Unpaired Image-to-Image Translation using Cycle-Consistent Adversarial Networks}},
    author={Zhu, Jun-Yan and Park, Taesung and Isola, Phillip and Efros, Alexei A},
    booktitle=ICCV,
    year={2017}
}

@inproceedings{karras2019style,
    title={{A Style-Based Generator Architecture for Generative Adversarial Networks}},
    author={Karras, Tero and Laine, Samuli and Aila, Timo},
    booktitle=CVPR,
    year={2019}
}

@article{avrahami2023blended,
    title={{Blended Latent Diffusion}},
    author={Avrahami, Omri and Fried, Ohad and Lischinski, Dani},
    journal=TOG,
    year={2023},
}

@inproceedings{nichol2022glide,
    title={{GLIDE: Towards Photorealistic Image Generation and Editing with Text-Guided Diffusion Models}},
    author={Nichol, Alexander Quinn and Dhariwal, Prafulla and Ramesh, Aditya and Shyam, Pranav and Mishkin, Pamela and Mcgrew, Bob and Sutskever, Ilya and Chen, Mark},
    booktitle=ICML,
    year={2022}
}

@article{wang2023instructedit,
    title={{InstructEdit: Improving Automatic Masks for Diffusion-based Image Editing With User Instructions}}, 
    author={Qian Wang and Biao Zhang and Michael Birsak and Peter Wonka},
    year={2023},
    journal={arXiv preprint arXiv:2305.18047},
}

@inproceedings{li2024zone,
    title={{ZONE: Zero-Shot Instruction-Guided Local Editing}},
    author={Li, Shanglin and Zeng, Bohan and Feng, Yutang and Gao, Sicheng and Liu, Xiuhui and Liu, Jiaming and Li, Lin and Tang, Xu and Hu, Yao and Liu, Jianzhuang and others},
    booktitle=CVPR,
    year={2024}
}

@inproceedings{brooks2023instructpix2pix,
    title={{InstructPix2Pix: Learning to Follow Image Editing Instructions}},
    author={Brooks, Tim and Holynski, Aleksander and Efros, Alexei A},
    booktitle=CVPR,
    year={2023}
}

@inproceedings{kirillov2023segment,
    title={{Segment Anything}},
    author={Kirillov, Alexander and Mintun, Eric and Ravi, Nikhila and Mao, Hanzi and Rolland, Chloe and Gustafson, Laura and Xiao, Tete and Whitehead, Spencer and Berg, Alexander C and Lo, Wan-Yen and others},
    booktitle=ICCV,
    year={2023}
}

@inproceedings{ren2024pixellm,
    title={{PixelLM: Pixel Reasoning with Large Multimodal Model}},
    author={Ren, Zhongwei and Huang, Zhicheng and Wei, Yunchao and Zhao, Yao and Fu, Dongmei and Feng, Jiashi and Jin, Xiaojie},
    booktitle=CVPR,
    year={2024}
}

@inproceedings{zhang2024omg,
    title={{OMG-LLaVA: Bridging Image-level, Object-level, Pixel-level Reasoning and Understanding}},
    author={Zhang, Tao and Li, Xiangtai and Fei, Hao and Yuan, Haobo and Wu, Shengqiong and Ji, Shunping and Loy, Chen Change and Yan, Shuicheng},
    booktitle=NIPS,
    year={2024}
}

@inproceedings{rasheed2024glamm,
    title={{GLaMM: Pixel Grounding Large Multimodal Model}},
    author={Rasheed, Hanoona and Maaz, Muhammad and Shaji, Sahal and Shaker, Abdelrahman and Khan, Salman and Cholakkal, Hisham and Anwer, Rao M and Xing, Eric and Yang, Ming-Hsuan and Khan, Fahad S},
    booktitle=CVPR,
    year={2024}
}

@inproceedings{ben2024mitigating,
  title={{Mitigating Open-Vocabulary Caption Hallucinations}},
  author={Ben-Kish, Assaf and Yanuka, Moran and Alper, Morris and Giryes, Raja and Averbuch-Elor, Hadar},
  booktitle=EMNLP,
  year={2024}
}

@article{zhang2023siren,
    title={{Siren's Song in the AI Ocean: A Survey on Hallucination in Large Language Models}},
    author={Zhang, Yue and Li, Yafu and Cui, Leyang and Cai, Deng and Liu, Lemao and Fu, Tingchen and Huang, Xinting and Zhao, Enbo and Zhang, Yu and Chen, Yulong and others},
    journal={arXiv preprint arXiv:2309.01219},
    year={2023}
}

@inproceedings{yu2016modeling,
  title={{Modeling Context in Referring Expressions}},
  author={Yu, Licheng and Poirson, Patrick and Yang, Shan and Berg, Alexander C and Berg, Tamara L},
  booktitle=ECCV,
  year={2016},
}

@article{ren2024grounded,
    title={{Grounded SAM: Assembling Open-World Models for Diverse Visual Tasks}},
    author={Ren, Tianhe and Liu, Shilong and Zeng, Ailing and Lin, Jing and Li, Kunchang and Cao, He and Chen, Jiayu and Huang, Xinyu and Chen, Yukang and Yan, Feng and others},
    journal={arXiv preprint arXiv:2401.14159},
    year={2024}
}

@inproceedings{yin2024benchmarking,
    title={{Benchmarking Segmentation Models with Mask-Preserved Attribute Editing}},
    author={Yin, Zijin and Liang, Kongming and Li, Bing and Ma, Zhanyu and Guo, Jun},
    booktitle=CVPR,
    year={2024}
}

@inproceedings{zhang2023magicbrush,
  title={{MagicBrush: A Manually Annotated Dataset for Instruction-Guided Image Editing}},
  author={Zhang, Kai and Mo, Lingbo and Chen, Wenhu and Sun, Huan and Su, Yu},
  booktitle=NIPS,
  year={2023}
}

@inproceedings{huang2024smartedit,
    title={{SmartEdit: Exploring Complex Instruction-based Image Editing with Multimodal Large Language Models}},
    author={Huang, Yuzhou and Xie, Liangbin and Wang, Xintao and Yuan, Ziyang and Cun, Xiaodong and Ge, Yixiao and Zhou, Jiantao and Dong, Chao and Huang, Rui and Zhang, Ruimao and others},
    booktitle=CVPR,
    year={2024}
}

@inproceedings{xia2024gsva,
    title={{GSVA: Generalized Segmentation via Multimodal Large Language Models}},
    author={Xia, Zhuofan and Han, Dongchen and Han, Yizeng and Pan, Xuran and Song, Shiji and Huang, Gao},
    booktitle=CVPR,
    year={2024}
}

@inproceedings{song2024doubly,
    title={{Doubly Abductive Counterfactual Inference for Text-based Image Editing}},
    author={Song, Xue and Cui, Jiequan and Zhang, Hanwang and Chen, Jingjing and Hong, Richang and Jiang, Yu-Gang},
    booktitle=CVPR,
    year={2024}
}

@inproceedings{sobieski2024rethinking,
    title={{Rethinking Visual Counterfactual Explanations Through Region Constraint}},
    author={Sobieski, Bartlomiej and Grzywaczewski, Jakub and Sadlej, Bart{\l}omiej and Tivnan, Matthew and Biecek, Przemyslaw},
    booktitle=ICLR,
    year={2024}
}

@article{wahed2024prima,
    title={{PRIMA: Multi-Image Vision-Language Models for Reasoning Segmentation}},
    author={Wahed, Muntasir and Nguyen, Kiet A and Juvekar, Adheesh Sunil and Li, Xinzhuo and Zhou, Xiaona and Shah, Vedant and Yu, Tianjiao and Yanardag, Pinar and Lourentzou, Ismini},
    journal={arXiv preprint arXiv:2412.15209},
    year={2024}
}

@article{liu2025seg,
    title={{Seg-Zero: Reasoning-Chain Guided Segmentation via Cognitive Reinforcement}},
    author={Liu, Yuqi and Peng, Bohao and Zhong, Zhisheng and Yue, Zihao and Lu, Fanbin and Yu, Bei and Jia, Jiaya},
    journal={arXiv preprint arXiv:2503.06520},
    year={2025}
}

@article{wei2024hyperseg,
    title={{HyperSeg: Towards Universal Visual Segmentation with Large Language Model}},
    author={Wei, Cong and Zhong, Yujie and Tan, Haoxian and Liu, Yong and Zhao, Zheng and Hu, Jie and Yang, Yujiu},
    journal={arXiv preprint arXiv:2411.17606},
    year={2024}
}

@article{wei2024instructseg,
    title={{InstructSeg: Unifying Instructed Visual Segmentation with Multi-modal Large Language Models}},
    author={Wei, Cong and Zhong, Yujie and Tan, Haoxian and Zeng, Yingsen and Liu, Yong and Zhao, Zheng and Yang, Yujiu},
    journal={arXiv preprint arXiv:2412.14006},
    year={2024}
}

@inproceedings{dai2023plausible,
    title={{Plausible May Not Be Faithful: Probing Object Hallucination in Vision-Language Pre-training}},
    author={Dai, Wenliang and Liu, Zihan and Ji, Ziwei and Su, Dan and Fung, Pascale},
    booktitle=ACL,
    year={2023}
}

@article{bhattad2025visual,
  title={{Visual Jenga: Discovering Object Dependencies via Counterfactual Inpainting}},
  author={Bhattad, Anand and Preechakul, Konpat and Efros, Alexei A},
  journal={arXiv preprint arXiv:2503.21770},
  year={2025}
}

@article{tversky1977features,
    title={{Features of Similarity}},
    author={Tversky, Amos},
    journal={Psychological review},
    year={1977},
    publisher = {American Psychological Association}
}

@inproceedings{yu2024revisiting,
  title={Revisiting counterfactual problems in referring expression comprehension},
  author={Yu, Zhihan and Li, Ruifan},
  booktitle=CVPR,
  pages={13438--13448},
  year={2024}
}

@inproceedings{hu2022lora,
title={Lo{RA}: Low-Rank Adaptation of Large Language Models},
author={Edward J Hu and Yelong Shen and Phillip Wallis and Zeyuan Allen-Zhu and Yuanzhi Li and Shean Wang and Lu Wang and Weizhu Chen},
booktitle=ICLR,
year=2022
}

@article{liu2025visionreasoner,
  title        = {VisionReasoner: Unified Visual Perception and Reasoning via Reinforcement Learning},
  author       = {Liu, Yuqi and Qu, Tianyuan and Zhong, Zhisheng and Peng, Bohao and Liu, Shu and Yu, Bei and Jia, Jiaya},
  journal      = {arXiv preprint arXiv:2505.12081},
  year         = {2025}
}

@article{zhang2024countercurate,
  title={Countercurate: Enhancing physical and semantic visio-linguistic compositional reasoning via counterfactual examples},
  author={Zhang, Jianrui and Cai, Mu and Xie, Tengyang and Lee, Yong Jae},
  journal={Findings of the Association for Computational Linguistics ACL 2024},
  year={2024}
}

@inproceedings{liu2024finecops,
  title={FineCops-Ref: A new Dataset and Task for Fine-Grained Compositional Referring Expression Comprehension},
  author={Liu, Junzhuo and Yang, Xuzheng and Li, Weiwei and Wang, Peng},
  booktitle=EMNLP,
  year={2024}
}

@article{liu2025medsam3,
  title={MedSAM3: Delving into Segment Anything with Medical Concepts},
  author={Liu, Anglin and Xue, Rundong and Cao, Xu R and Shen, Yifan and Lu, Yi and Li, Xiang and Chen, Qianqian and Chen, Jintai},
  journal={arXiv preprint arXiv:2511.19046},
  year={2025}
}
